\documentclass{article} %

\usepackage[svgnames,table]{xcolor}
\usepackage[colorlinks,allcolors=blue]{hyperref}

\usepackage{preprint}

\usepackage{etoolbox}
\AtBeginEnvironment{tabularx}{\footnotesize\centering}
\AtBeginEnvironment{figure}{\centering}

\usepackage{graphicx}
\definecolor{lightgray}{gray}{0.9}
\definecolor{LightBlue}{rgb}{0.68, 0.85, 0.9}
\definecolor{DarkSeaGreen}{cmyk}{0.69,0,0.50,0.5}
\definecolor{DarkGreen}{rgb}{0,.5,0}
\definecolor{MyGray}{gray}{.5}
\definecolor{ForestGreen}{rgb}{0,0.71,0.04001}
\usepackage[english]{babel} 
\usepackage{xspace}
\usepackage[authoryear]{natbib}
\usepackage{amsmath}
\usepackage{paralist}
\usepackage[shortlabels]{enumitem}
\usepackage{array}
\usepackage{subfig}
\usepackage{amssymb,amsthm}
\usepackage{booktabs}
\usepackage{tabularx}
\usepackage{multirow}
\usepackage{multicol}
\usepackage{mdframed}
\usepackage{algorithm}
\usepackage{algpseudocodex}

\usepackage{acro}
\acsetup{format-include-endings=true, group-citation=true, list-style=tabular} 
\DeclareAcronym{CDT}{short = DTF, long = {digital twin framework}}
\DeclareAcronym{DTMC}{short = DTMC, long = {discrete{\Hyphdash}time Markov chain}}
\DeclareAcronym{LTS}{short = LTS, long = {labelled transition system}}
\DeclareAcronym{MDP}{short = MDP, long = {Markov decision process}, long-plural=es}
\DeclareAcronym{PCTL}{short = PCTL, long = {probabilistic computation
    tree logic}}
\DeclareAcronym{pDTMC}{short = pDTMC, long = {parametric \ac{DTMC}}}
\DeclareAcronym{pGCL}{short = pGCL, long = {probabilistic guarded command language}}

\usepackage{cleveref}
\Crefname{section}{Section}{Sections}
\Crefname{table}{Table}{Tables}
\Crefname{figure}{Figure}{Figures}
\Crefname{equation}{Equation}{Equations}
\Crefname{lstlisting}{Listing}{Listings}

\usepackage{listings}
\lstdefinelanguage{prism}{
  morekeywords={A, bool, clock, const,
    ctmc, C, double, dtmc, E, endinit, endinvariant, endmodule, endrewards, endsystem,
    false, formula, filter, func, F, global, G, init, invariant, I, int, label, max, mdp,
    min, module, X, nondeterministic, Pmax, Pmin, P, probabilistic, prob, pta, rate,
    rewards, Rmax, Rmin, R, S, stochastic, system, true, U, W},
  morecomment=[l]{//}}
\lstdefinelanguage{yapscript}{
  morekeywords={Settings, outputDepth, endangermentDepth, mitigationDepth,
    simulationLength, suppressMishaps, suppressEndangerments,
    suppressMitigations, suppressLoops, suppressStateLabel,
    suppressMishapLabel, suppressEndangermentsLabel,
    suppressMitigationsLabel, OperationalSituation,OpSit, Activity,
    include, successor, initialState, ControlLoop,Ctr,Loop,
    mode,event,update,cf,target, for, HazardModel, HazMod, alias, desc,
    description, partOf, poweredBy, offRepair, direct, activatedBy,
    detectedBy, mitigatedBy, resumedBy, embodiedBy, guard, constraint,
    endMitigatedBy, requires, requiresOneOf, requiresNOf, requiresOcc,
    causes, permits, excludes, prevents, preventsMit, mitPreventsMit, final,
    mishap, triggeredBy},
  morecomment=[l]{//},
  morecomment=[l]{\#},
  morecomment=[l]{desc},
  morecomment=[s]{/*}{*/},
  sensitive=false}
\lstdefinestyle{arxiv}{
  mathescape=true, breaklines=true,
  basicstyle=\sffamily\small, keywordstyle=\bfseries, identifierstyle=\slshape,
  commentstyle=\color{gray}, emphstyle=\bfseries, xleftmargin=13pt, 
  tabsize=4, numbers=left, numberstyle=\footnotesize, numbersep=5pt, columns=flexible}
\newcommand{\yapid}[1]{\textsf{\upshape #1}}
\newcommand{\yapkeyw}[1]{\textcolor{black}{\textsf{\upshape #1}}}
\newcommand{\rgsym}[1]{\textrm{\textit{#1}}}

\usepackage{ctable}

\newcommand{\sem}[1]{[\![#1]\!]}
\newcommand{\semRS}[1]{\sem{#1}_{R}}

\usepackage{cancel}

\newcommand{\rfct}[1][f]{\mathsf{#1}}
\newcommand{\Phases}[1][\rfct]{\mathit{Ph}_{#1}}
\newcommand{\event}[1][]{\mathit{#1}}
\newcommand{\rapac}[2][]{\event[#2]_{\mathit{#1}}}
\newcommand{\acend}[2][]{\rapac[#1]{e}^{\rfct[#2]}}
\newcommand{\acmis}[2][]{\rapac[#1]{\underline{e}}^{\rfct[#2]}}
\newcommand{\acmit}[2][]{\rapac[#1]{m}^{\rfct[#2]}}
\newcommand{\acres}[2][]{\rapac[#1]{r}^{\rfct[#2]}}
\newcommand{\inact}[1][f]{\xcancel{\mathit{#1}}} %
\newcommand{\activ}[1][f]{\mathit{#1}}
\newcommand{\mitig}[1][f]{\overline{\mathit{#1}}}
\newcommand{\mishp}[1][f]{\underline{\mathit{#1}}}

\newtheorem{definition}{Definition}
\usepackage[font={small}]{caption}
\usepackage[shortcuts]{extdash}
\usepackage{mathabx}
\usepackage{natbib}
\newmdtheoremenv[hidealllines=true,leftmargin=-4pt,rightmargin=-4pt,%
backgroundcolor=gray!20,%
innerleftmargin=4pt,innerrightmargin=4pt,%
ntheorem]{example}{Example}

\newenvironment{wrapfigure}[3][]{%
  \begin{figure}}{\end{figure}}
\newenvironment{wraptable}[3][]{%
  \begin{table}}{\end{table}}

\lstdefinestyle{arxiv}{
  basicstyle=\sffamily\scriptsize}
\makeatletter%
\begin{document}
\newcommand{\shorttitle}{Verified Synthesis of Safety Controllers}
\title{Verified Synthesis of Optimal Safety Controllers
  for Human-Robot Collaboration\thanks{
    This research has received funding from the Assuring Autonomy
    International Programme~(AAIP grant CSI: Cobot), a partnership between
    Lloyd’s Register Foundation and the University of York, and from the
    UKRI project EP/V026747/1 `Trustworthy Autonomous Systems Node in
    Resilience'.
  }}
\author{
  Mario Gleirscher\\ %
  University of Bremen, Germany\\ \texttt{mario.gleirscher@uni-bremen.de}
  \And
  Radu Calinescu\\ University of York, UK\\ \texttt{radu.calinescu@york.ac.uk}
  \And
  James Douthwaite\\ University of Sheffield, UK\\ \texttt{j.douthwaite@sheffield.ac.uk}
  \And
  Benjamin Lesage\\ University of York, UK\\ \texttt{benjamin.lesage@york.ac.uk}
  \And
  Colin Paterson\\ University of York, UK\\ \texttt{colin.paterson@york.ac.uk}
  \And
  Jonathan Aitken\\ University of Sheffield, UK\\ \texttt{jonathan.aitken@sheffield.ac.uk}
  \And
  Rob Alexander\\ University of York, UK\\ \texttt{rob.alexander@york.ac.uk}
  \And
  James Law\\ University of Sheffield, UK\\ \texttt{j.law@sheffield.ac.uk}}

\maketitle
\begin{abstract}
  We present a tool-supported approach for the synthesis,
  verification and validation of the control software responsible for
  the safety of the human-robot interaction in manufacturing processes
  that use collaborative robots. In human-robot collaboration,
  software-based safety controllers are used to improve operational
  safety, e.g., by triggering shutdown mechanisms or emergency stops
  to avoid accidents. Complex robotic tasks and increasingly close
  human-robot interaction pose new challenges to controller developers
  and certification authorities. Key among these challenges is the
  need to assure the correctness of safety controllers under explicit
  (and preferably weak) assumptions. Our controller synthesis,
  verification and validation approach is informed by the
  process, risk analysis, and relevant safety regulations for the
  target application. Controllers are selected from a design space of
  feasible controllers according to a set of optimality criteria,
  are formally verified against correctness criteria, and are translated into
  executable code and validated in a digital twin. The resulting
  controller can detect the occurrence of hazards, move the process
  into a safe state, and, in certain circumstances, return the process
  to an operational state from which it can resume its original task.
  We show the effectiveness of our software engineering
  approach through a case study involving the development of a safety
  controller for a manufacturing work cell equipped with a
  collaborative robot.
\end{abstract}

  \vspace{1em}

\section{Introduction}
\label{sec:intro}

Effective collaboration between humans and
robots~\citep{Nicolaisen1985-OccupationalSafetyIndustrial,
  Jones1986-StudySafetyProduction} %
can leverage their complementary skills.  But such collaboration is
difficult to achieve because of uncontrolled hazards and because
sensing, tracking, and safety measures are either still unexploited in
practice~\citep{Santis2008-atlasphysicalhumanrobot} or they are
difficult to validate following state-of-the-art safety
regulations~\citep{Chemweno2020-Orientingsafetyassurance}.
Since the 1980s, remote programming (also called
tele{\Hyphdash}programming) and simulation have led to some reduction
of hazard exposure.  However, the effectiveness of human-robot
collaboration is still limited because of frequent conservative
shutdowns, simplistic emergency stops, and unspecific error handling
procedures.  Extensive guarding arrangements interfere
with manufacturing processes and mobile robot applications.
But effective work processes and complex tasks require continuous 
close human-robot interaction~(e.g.\xspace mutual take-over of tasks), mutual
clarification of intent, and trading off
risk~\citep{Hayes2013-ChallengesSharedEnvironment,
  Villani2018-Surveyhumanrobotcollaboration}.  From an operator's
perspective, robot movements need to be predictable, and potential
impacts on the human body need to be attenuated.  From a control
perspective, the confident monitoring and control of the robot speed
and the separation between machines and humans require high-quality
stereo vision and laser scanners to distinguish several safety zones.
A decade after these issues were discussed in
\citet{Alami2006-Safedependablephysical} and
\citet{Haddadin2009-RequirementsSafeRobots},
\citet{Ajoudani2017-Progressprospectshumanrobot} %
emphasise that the complex safety challenges of collaborative
robots~(cobots for short,
\citealp{Gillespie2001-generalframeworkcobot}) remain largely
unresolved.
Increasingly complex robotic systems reduce the ability to understand
and mitigate risks.  Safety is a major barrier to the more widespread
adoption of cobots, with organisations such as manufacturers having to
resort to sub-optimal processes due to safety concerns.  Methods of
ensuring safe human-robot interaction will deliver \$7.3bn of savings,
reducing costs of US-manufactured goods by 1\%
\citep{Anderson2016-EconomicImpactTechnology}.  However, little such
method and tool support is available for engineers that have to
implement and confidently assure cobot safety requirements, that is,
the results of cobot hazard analyses and risk
assessments~\citep{Chemweno2020-Orientingsafetyassurance}.

\paragraph{Problem.}
\label{sec:safety-req}

Among the measures for improving cobot safety, the monitoring of
application processes, the handling of critical events, and the
mitigation of operational risk are the responsibility of
software{\Hyphdash}based \emph{safety controllers}.  To facilitate
smooth human-robot collaboration with minimal interruption, the
software engineers responsible for developing these controllers need to
closely consider the process with its variety and complexity of
adverse events, such as unusual operator behaviour and equipment
failure modes.  This leads to complex requirements and design spaces
for the safety controllers, so these engineers must address questions
in several areas:
\begin{enumerate}
\item \emph{Risk assessment.} Which controller minimises the
  probability of incidents in the presence of human and sensor errors?
\item \emph{Controller synthesis.} Which design minimises nuisance to the
  human, maximises productivity, etc.\xspace while maintaining safety?
\item \emph{Controller verification.} Does a controller handle
  hazards when detected and return the system to a useful safe state?
\item \emph{Controller validation.} Does an implementation of the
  synthesised controller exhibit the intended behaviour?
\end{enumerate}

\paragraph{Preliminary solution.}
\label{sec:preliminary-solution}

In \cite{Gleirscher2020-SafetyControllerSynthesis}, we provide initial
answers to the first three questions, by introducing a preliminary
software engineering approach for the synthesis of
discrete{\Hyphdash}event safety controllers that meet safety
requirements and optimise process performance in human-robot
collaboration.  We model the application~(e.g.\xspace a manufacturing
process) as a %
\ac{MDP}, and select correct{\Hyphdash}by{\Hyphdash}construction
controllers from an associated design space.  The process model
describes the behaviour of all involved actors including the
controller.  To describe critical events~(e.g.\xspace hazards) and controller
actions~(e.g.\xspace safety mode changes), we employ the notion of \emph{risk
  structures} \citep{Gleirscher2017-FVAV,
  Gleirscher2021-RiskStructuresDesign} implemented in the
risk-informed controller designer
\textsc{Yap}\xspace~\citep{Gleirscher-YapManual,Gleirscher2020-YAPToolSupport} used
for risk structuring, qualitative risk analysis, and synthesis of
risk-informed safety controllers.  In particular, \textsc{Yap}\xspace supports risk
modelling and controller design with a domain-specific language and
automates the transformation of risk structures into guarded command
language and, in turn, \acp{MDP}.  The preliminary approach from
\citet{Gleirscher2020-SafetyControllerSynthesis} facilitates the
verification of the safety of the \ac{MDP} and of probabilistic
reach-avoid properties of selected \ac{MDP} policies including the
controller.  A verified controller extracted from such a policy
detects hazards and controls their mitigation by the execution of a
safety function, a transition to a particular safety mode, or a safer
process task or activity.  Furthermore, in certain circumstances, the
controller returns the process to an operational state from which it
can resume its original task.

\paragraph{Contributions.}
\label{sec:contrib}

In this paper, we extend our preliminary software engineering approach for safety controller synthesis
\citep{Gleirscher2020-SafetyControllerSynthesis,
  Gleirscher2020-YAPToolSupport} with multiple new capabilities, and
we overcome several of its limitations, as explained below.

We refine the process and risk modelling to enable the synthesis of
finer-grained controllers.  For this, we provide a refined factor
notion in \Cref{sec:refin-fact-modell}.  We improve and extend the
verification stage by checking additional properties, increasing the
confidence in the synthesised controllers.  We employ the stochastic model synthesis tool
\textsc{evoChecker}\xspace~\citep{Gerasimou2018-Synthesisprobabilisticmodels} in
addition to the probabilistic model checker
\textsc{Prism}\xspace~\citep{Kwiatkowska2011-PRISM4Verification}, extending the
verification capabilities of the optimal synthesis procedure to
significantly more complex combinations of controller requirements.

We translate the controllers into executable code for a \ac{CDT}
to foster controller validation in a realistic environment.  We
assure the correctness of this translation by reusing properties
from model checking in run-time verification.  The \ac{CDT}
validation of the safety controller is based on randomised use-case
tests that satisfy well-defined state-based coverage criteria.

We demonstrate in the \ac{CDT} how the synthesised safety controller
can automatically resume the nominal procedure after the mitigation of
hazards.  We provide experimental results that confirm our safety
controller's ability to achieve an increased process utility in
addition to ensuring high levels of freedom from accidents.

Overall, we increase the level of technical detail, to improve the
understanding and reproducibility of our results.
\Cref{sec:running-example} introduces our case study as a running
example, and \Cref{sec:prelim} provides the theoretical background.
We describe our process modelling and controller design method in
\Cref{sec:approach}, our approach to verified optimal synthesis in
\Cref{sec:optim-contr-synth}, and explain in \Cref{sec:ctr-depl} how
we implement, deploy, and validate the synthesised controllers in a
realistic purpose built digital twin.  We evaluate and discuss our
approach in \Cref{sec:eval}.  \Cref{sec:related-work} highlights
related work.  We conclude with a short summary in
\Cref{sec:conclusion}.

 \section{Running Example: Manufacturing Cobots}
\label{sec:running-example}

\begin{wrapfigure}{r}{.5\columnwidth}
  \subfloat[Work cell with cobot (conceptual, top view)]{%
    \includegraphics[width=5cm]{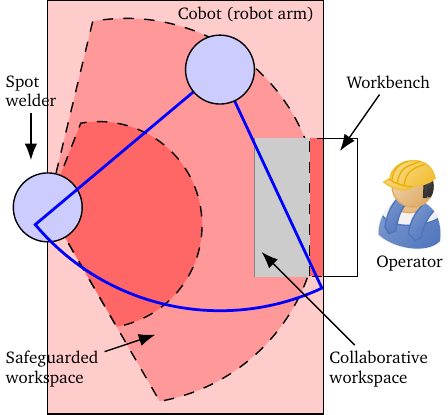}
    \label{fig:setting-conceptual}}
  \hspace{.5em}
  \subfloat[Process activities]{%
    $\qquad$\includegraphics[width=.14\textwidth]{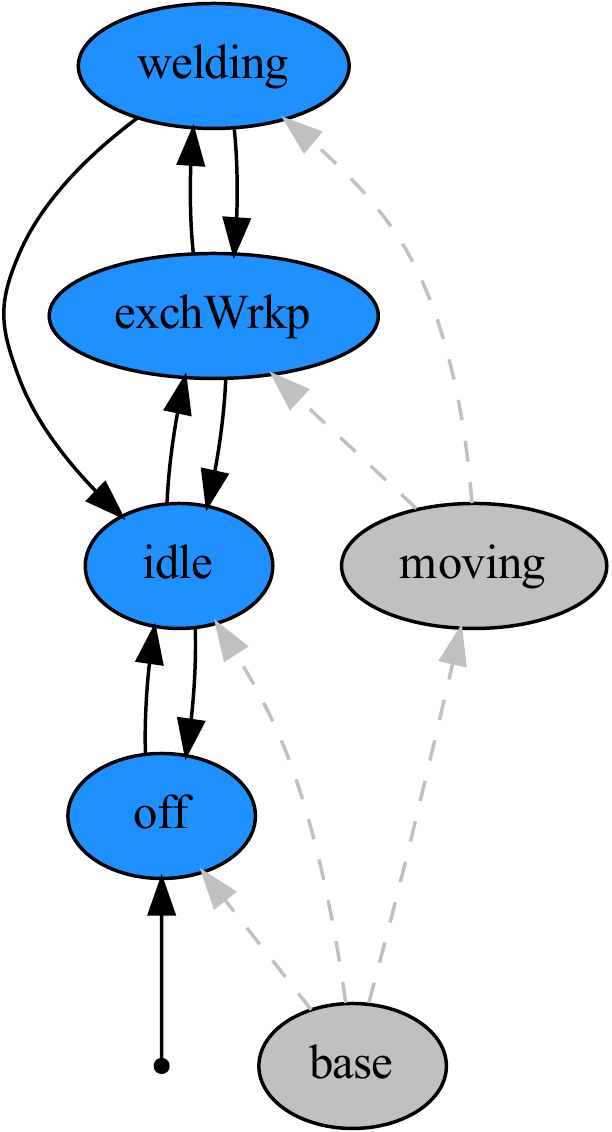}
    \label{fig:setting-activities}}
  \caption{Work cell concept (a) and activities in the manufacturing
    process (b) performed by the operator, the robot, and the spot welder,
    classified by the activity groups \emph{moving} and
    \emph{base}~(in light gray)
    \label{fig:setting}}
\end{wrapfigure}

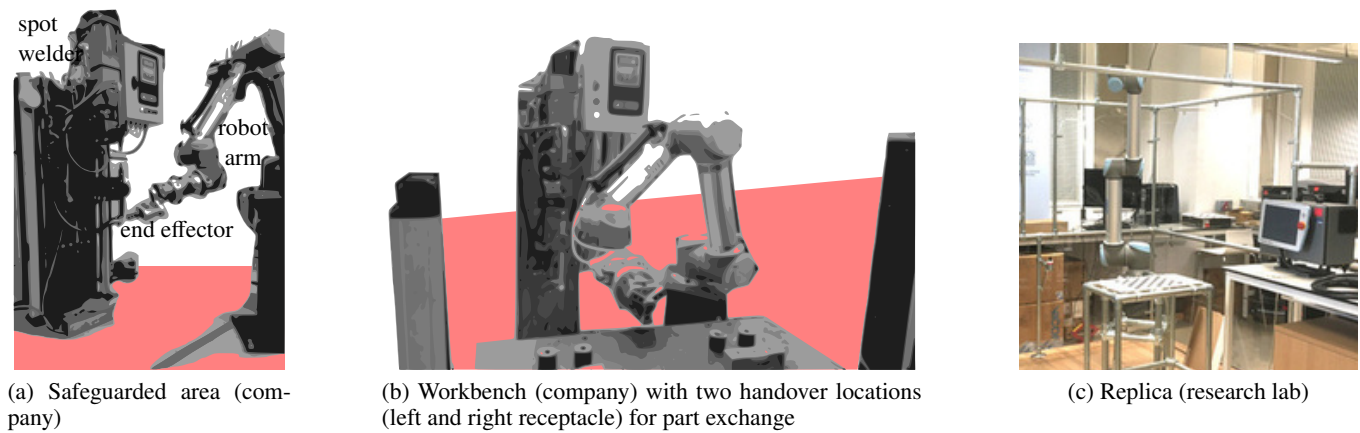
\begin{figure}
  \centering
  \makeatletter
  \subfloat[Safeguarded area (company)]{{\footnotesize
      \def\svgwidth{3.6cm}
\begingroup%
  \makeatletter%
  \providecommand\color[2][]{%
    \renewcommand\color[2][]{}%
  }%
  \providecommand\transparent[1]{%
    \renewcommand\transparent[1]{}%
  }%
  \providecommand\rotatebox[2]{#2}%
  \newcommand*\fsize{\dimexpr\f@size pt\relax}%
  \newcommand*\lineheight[1]{\fontsize{\fsize}{#1\fsize}\selectfont}%
  \ifx\svgwidth\undefined%
    \setlength{\unitlength}{226.87545776bp}%
    \ifx\svgscale\undefined%
      \relax%
    \else%
      \setlength{\unitlength}{\unitlength * \real{\svgscale}}%
    \fi%
  \else%
    \setlength{\unitlength}{\svgwidth}%
  \fi%
  \global\let\svgwidth\undefined%
  \global\let\svgscale\undefined%
  \makeatother%
  \begin{picture}(1,1.33155289)%
    \lineheight{1}%
    \setlength\tabcolsep{0pt}%
    \put(0,0){\includegraphics[width=\unitlength,page=1]{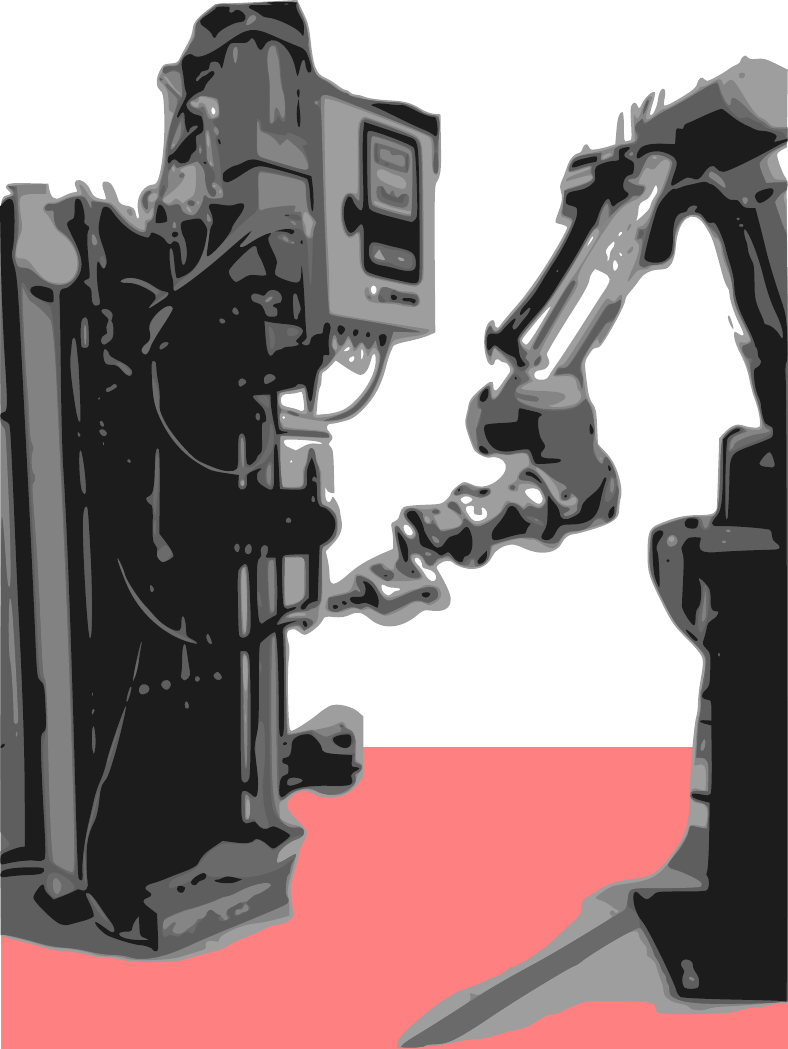}}%
    \put(0.01642296,1.25038127){\color[rgb]{0,0,0}\makebox(0,0)[lt]{\lineheight{1.25}\smash{\begin{tabular}[t]{l}spot\\welder\end{tabular}}}}%
    \put(0.8474964,0.86574726){\color[rgb]{0,0,0}\makebox(0,0)[t]{\lineheight{1.25}\smash{\begin{tabular}[t]{c}robot\\arm\end{tabular}}}}%
    \put(0.39497232,0.48872829){\color[rgb]{0,0,0}\makebox(0,0)[lt]{\lineheight{1.25}\smash{\begin{tabular}[t]{l}end effector\end{tabular}}}}%
  \end{picture}%
\endgroup%
 }
    \label{fig:setting-actual2}}
  \makeatother
  \hfill
  \subfloat[Workbench (company) with two handover locations (left and
  right receptacle) for part exchange]{
    \includegraphics[width=7cm]{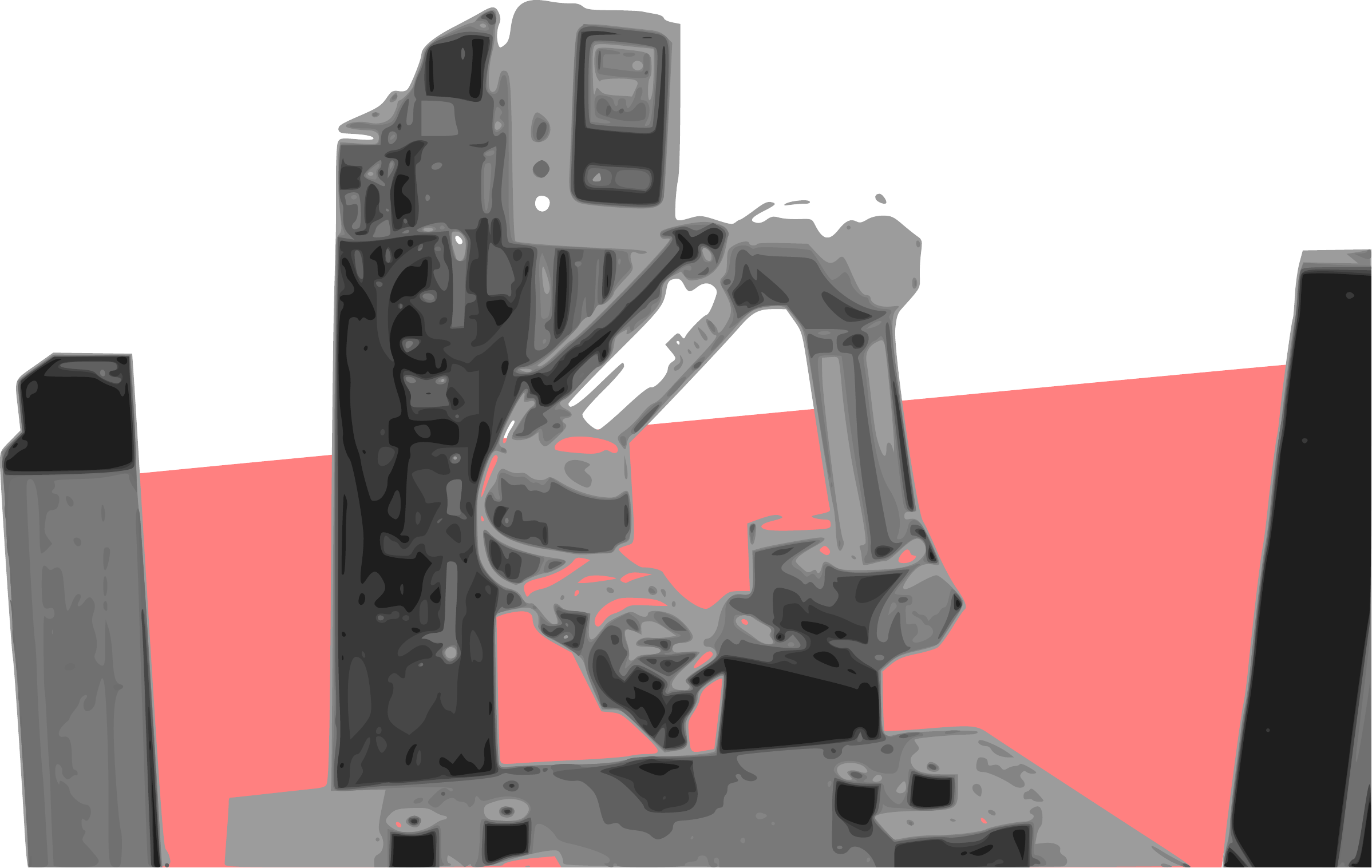}
    \label{fig:setting-actual1}}
  \hfill
  \subfloat[Replica (research lab)]{
    \includegraphics[width=4.5cm,trim=100 120 10 100,clip]{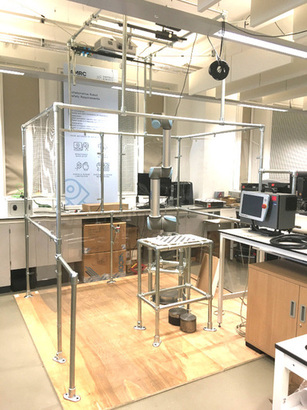}
    \label{fig:setting-replica}}
  \caption{Actual (a, b) and replicated (c) work cell with cobot
    \label{fig:cobot-all}}
\end{figure}

\Cref{fig:cobot-all} shows a cobot-equipped manufacturing \emph{work
  cell} at a UK company (with the pictures anonymised for
confidentiality reasons) and replicated in a testbed at the University
of Sheffield~(\Cref{fig:setting-replica}).  In the corresponding
process, $\mathcal{P}$, an operator, a stationary collaborative
robotic manipulator (robot arm for short), and a spot welder
(\Cref{fig:setting-conceptual}) collaborate repetitively on several
activities~(\Cref{fig:setting-activities}).\footnote{For the sake of
  simplicity, we use the notion of an \emph{activity} as a hypernym
  describing a task, a situation, a use case, or a scenario.}
Previous safety analysis~(i.e.,\xspace hazard identification, risk assessment,
and requirements derivation) resulted in two sensors.  The first one
is a range finder using a rotating laser beam (indicated with the
highlighted area at the bottom of \Cref{fig:setting-actual2}) to
determine the distance between the spot welder and a person or an
object intruding into the highlighted area and triggering a slow down
or an emergency stop if the intruder approaches the spot welder.  The
second one is a light barrier (the highlighted curtain indicated in
\Cref{fig:setting-actual1}) triggering such a stop if something like a
person's arm reaches across the workbench while the robot or the spot
welder are active.
\Cref{tab:riskana} shows our partial safety analysis of the cell
following the guidance in~\Cref{sec:intro}.  The right column
specifies safety goals against each accident and controller
requirement candidates~(e.g.\xspace mode-switch requirements) handling each
latent cause in the left column, and indicating how the hazard is to
be removed.
The running example is part of a case study organised around the AAIP
project CSI:~Cobot.\footnote{See
  \url{https://www.sheffield.ac.uk/sheffieldrobotics/about/csi-cobot}.}
Due to Covid-19 restrictions limiting access to physical facilities
during a critical phase of the project, we use a digital twin of the
work cell as a target platform to deploy and validate synthesised
safety controllers.

\begin{table*}%
  \caption{Our partial safety analysis of the manufacturing cell
    referring to the measures recommended in \citet{ISO15066}}
  \label{tab:riskana}
  \begin{tabularx}{\textwidth}{l>{\hsize=.35\hsize}X>{\hsize=.65\hsize}X}
    \toprule
    \textbf{Id}
    & \textbf{Critical Event} (risk factor)
    & \textbf{Safety Requirement}
    \\\specialrule{\lightrulewidth}{\aboverulesep}{0em}
    \rowcolor{lightgray}{\large$\phantom{\mid}$\hspace{-.25em}}
    & \textbf{Accident} (to be prevented or alleviated)
    & \textbf{Safety Goal}
    \\\specialrule{\lightrulewidth}{0em}{\belowrulesep}
    $\rfct[RC]$
    & The \underline{R}obot arm harshly \underline{C}ollides with an
      operator. 
    & The robot shall \emph{avoid} harsh active collisions with the
      operator.
    \\
    $\rfct[WS]$
    & \underline{W}elding \underline{S}parks cause operator injuries
      (skin burns).
    & The welding process shall \emph{reduce} sparks injuring the operator.
    \\
    $\rfct[RT]$
    & The \underline{R}obot arm \underline{T}ouches the operator.
    & The robot shall \emph{avoid} active contact with the operator.
    \\\specialrule{\lightrulewidth}{\aboverulesep}{0em}
    \rowcolor{lightgray}{\large$\phantom{\mid}$\hspace{-.25em}}
    & \textbf{Latent Cause}~(to be mitigated timely)$^{\dagger}$
    & \textbf{Controller Requirement}$^\ddagger$
    \\\specialrule{\lightrulewidth}{0em}{\belowrulesep}
    $\rfct[HRW]$
    & The \underline{H}uman operator and the \underline{R}obot use the
      \underline{W}orkbench at the same time.
    & (m) The robot shall perform an appropriate mitigation (e.g.\xspace a
      safety-rated monitored stop) and (r) resume \emph{normal
      operation} after the \emph{operator} has left the \emph{shared
      workbench}.  
    \\
    $\rfct[HW]$
    & The \underline{H}uman operator is entering the \underline{W}orkbench 
      while the robot is away from the workbench.
    & (m) If the robot moves a workpiece to the workbench then
      it shall switch to \emph{power \& force limiting} mode and (r) resume
      \emph{normal operation} after the \emph{operator} has left the
      \emph{workbench}.
    \\
    $\rfct[HS]$
    & The \underline{H}uman operator has entered the
      \underline{S}afeguarded area while the robot or the
      spot welder are active.
    & (m) The \emph{spot welder} shall be \emph{switched off}, the
      \emph{robot} to \emph{speed \& separation monitoring}, and 
      the operator be notified to leave. (r) Robot and spot welder shall
      resume normal mode after the operator has left. 
    \\
    $\rfct[HC]$
    & The \underline{H}uman operator is \underline{C}lose to the
      welding spot while the robot is working and the spot welder is active.
    & (m) The \emph{spot welder} shall be \emph{switched off}, the
      \emph{robot} to \emph{safety-rated monitored stop}. (r) Both shall
      resume \emph{normal or idle mode with a reset procedure} after the
      operator has left.  
    \\\bottomrule
    \multicolumn{3}{l}{$^{\dagger}$m: mitigation requirement, r:
    resumption requirement, $^\ddagger$subjected to generalisation to
    define a controller design space}
  \end{tabularx}
\end{table*}

 \section{Background}
\label{sec:prelim}

This section summarises the background of the presented approach,
particularly, cobot safety, probabilistic model checking, system
safety analysis, and risk-informed controller modelling.

\subsection{Robot Safety: From Industrial to Collaborative}
\label{sec:safe-human-robot}

\begin{wraptable}{r}{.5\columnwidth}
  \caption{Cobot safety measures associated to stages in the
    causal chain of events}%
  \label{tab:measures}
  \begin{tabularx}{.5\columnwidth}{%
      >{\bfseries\hsize=.15\hsize}X
      >{\hsize=.33\hsize}X
      >{\hsize=.52\hsize}X}
    \specialrule{\heavyrulewidth}{\aboverulesep}{0em}
    \rowcolor{lightgray}{\large$\phantom{\mid}$\hspace{-.25em}}
    \textbf{Stage}
    & \textbf{Type of Measure}
    & \textbf{Examples}
    \\\specialrule{\lightrulewidth}{0em}{\belowrulesep}
    \multirow{2}{1.5cm}{Hazard prevention}
    & 1. Safeguard/barrier
    & Fence, cage, interlock
    \\
    & 2. IT safety 
    & \emph{Verified$^\dagger$ safety controller}
    \\
    & 3. IT security 
    & Security-verified$^\dagger$ (safety) controller
    \\\midrule
    \multirow{2}{1.5cm}{Hazard mitigation \&
      accident prevention}
    & 4. Reliability 
    & Fault-tolerant scene interpretation
    \\
    & 5. Workspace intrusion detection
    & Speed \& separation monitoring,
    safety-rated monitored stop
    \\
    & 6. Shift of control
    & Hand-guided operation
    \\\midrule
    \multirow{2}{1.5cm}{Accident mitigation (alleviation)}
    & 7. Power \& force limitation
    & Low weight parts, flexible surfaces; variable impedance,
    touch-sensitive, \& force-feedback control
    \\
    & 8. System halt
    & Emergency stop, dead-man's switch
    \\\bottomrule
    \multicolumn{3}{l}{$^\dagger$avoidance of development or programming
      mistakes} 
  \end{tabularx}
\end{wraptable}

Hazards from robots have been studied since the advent of industrial
robotics in the 1970s, resulting in risk taxonomies based on
workspaces, tasks, and human body
regions~\citep{Sugimoto1977-SafetyEngineeringIndustrial,
  Jones1986-StudySafetyProduction, Alami2006-Safedependablephysical,
  Haddadin2009-RequirementsSafeRobots,
  Wang2017-Humanrobotcollaborativeassembly,
  Kaiser2018-SafetyRelatedRisks,
  Matthias2011-Safetycollaborativeindustrial,
  Marvel2015-CharacterizingTaskBased}.  The majority of hazards are
\emph{impact hazards}~(e.g.\xspace unexpected movement, reach beyond area,
dangerous workpieces, hazardous manipulation), \emph{trapping
  hazards}~(e.g.\xspace operator locked in cage), and \emph{failing
  equipment} (e.g.\xspace valve, cable, sensor, controller).  In the 1980s,
robots were programmed interactively by operators being in the cage
while powered, which had caused frequent accidents from trapping and
collision.  However, from the late 1990s on, the increased use of
tele-programming contributed to the reduction of accidents related to
such hazards.

Addressing hazards outside the programming stage involves the
examination of each \emph{mode of operation}~(e.g.\xspace normal,
maintenance) for its hazardous behaviour, and the use of safety
controllers to trigger mode-specific \emph{safety
  measures}~\citep{Jones1986-StudySafetyProduction}. %
Malfunction diagnostics~(e.g.\xspace fault detection, wear-out monitoring)
can further inform these controllers.  \Cref{tab:measures} shows a
variety of measures~\citep{Santis2008-atlasphysicalhumanrobot} to
prevent or mitigate hazards and accidents by reducing the probability
of the occurrence and the severity of the consequences of these
hazards.  If these measures use electronic or mechatronic equipment,
we speak of \emph{functional}\footnote{ \emph{Functional safety}~(see
  IEC~61508, ISO~26262) deals with the dependability, particularly,
  correctness and reliability, of \emph{critical} programmable
  electronic systems.  Safety functions or ``functional measures'' are
  the archetype of such systems.} measures~(e.g.\xspace safety modes as
exemplified below) and of intrinsic measures otherwise~(e.g.\xspace a fence
around a robot, flexible robot surfaces).  Functional measures
focusing on the correctness and reliability of a controller (a
programmable electronic or software system) are called dependability
measures~\citep{Alami2006-Safedependablephysical,
  Avizienis2004-Basicconceptstaxonomy}.  Functional measures are said
to be passive if they focus on severity reduction~(e.g.\xspace force-feedback
control), \emph{active} otherwise.  In this work, we focus on the
verified synthesis~\citep{kress-gazit_synthesis_2018} of safety
controllers that realise active functional safety measures.

Standardisation of safety requirements for industrial robots
\citep{Sugimoto1977-SafetyEngineeringIndustrial} culminated in
ANSI/RIA R15.06, \citet{ISO10218}, 13482, and 15066.  Following
ISO~10218, such robot systems comprise a robot arm, a robot
controller, an end{\Hyphdash}effector, and a work piece~(see,
e.g.\xspace\Cref{fig:setting-conceptual}).
According to \citet{Helms2002-robatworkRobotassistant} and
\citet{Kaiser2018-SafetyRelatedRisks}, one can distinguish four
scenarios of human-robot interaction:
\begin{inparaenum}[(i)]
\item Encapsulation in a fenced robot work space,
\item co{\Hyphdash}existence without fencing but separation of human
  and robot work space,
\item cooperation with alternative exclusive use of shared work space,
  and
\item collaboration with simultaneous use of shared work space and
  close interaction.
\end{inparaenum} 
Cooperation and collaboration are the two most interactive of these
scenarios and motivate our work.
In collaborative operation, the operator and the
cobot~\citep{Gillespie2001-generalframeworkcobot} can occupy the
collaborative workspace simultaneously while the cobot is performing
tasks~\citep[Pt.~3.1]{ISO15066}.  The \emph{collaborative workspace}
has to be a subset of the \emph{safeguarded workspace}.

Based on these definitions, ISO~15066 recommends four \emph{safety
  modes}.  First, a \emph{safety-rated monitored stop} is an active
functional measure realised as a mode where the robot is still powered
but there is no simultaneous activity of the robot and the operator in
the shared workspace.
Second, \emph{hand-guided operation} refers to a mode with
zero-gravity control, that is, control without actuation beyond the
compensation of gravity, solely guided by an operator.  Hand guidance
requires the robot to be in a compliant state, with control exerted by
the operator through physical manipulation.
Third, \emph{speed \& separation monitoring} as an active functional
measure refers to a mode where speed is continuously adapted to the
distance of the robot and an operator.
Forth, \emph{power \& force limiting} is a mode with reduced impact of
the robot on the human body and the robot's power and applied forces
are limited.  In this mode, a robot should not impact a human with
more than a defined force, with acceptable forces mapped out for
different impact points on the body.
\citet{Heinzmann2003-QuantitativeSafetyGuarantees} propose such a mode
always active during a collaborative activity described as a
discrete-event controller.  \citet{Long2018-industrialsecuritysystem}
propose a distance-triggered scheme to switch between
nominal~(max.~velocity), reduced~(speed limiting), and
passive~(hand-guided) operating modes.
\cite{Kaiser2018-SafetyRelatedRisks} and
\cite{Villani2018-Surveyhumanrobotcollaboration} describe and combine
these modes with work layouts.

In addition to these safety modes,
\cite{Alami2006-Safedependablephysical} highlight the necessity of a
general shift from robots whose motion is controlled only by following
a pre-specified route of positions (also known as position control) to
robots whose motion control minimises contact forces and energy (also
known as interaction control).  Overall,
interaction{\Hyphdash}controlled robots, with less pre-planning and
fewer assumptions on workspace structure and robot actions, can
exploit the mentioned safety modes more effectively than
position-controlled robots with extensive pre-planning and stronger
assumptions.

\subsection{Probabilistic Model Checking and Trace Checking}
\label{sec:probmc}

The proposed approach employs \acp{MDP}, and corresponding 
sets of \acp{DTMC}, as a formal model of the
process $\mathcal{P}$, and uses the policy space of an \ac{MDP},
describing the degrees of freedom for decision making, to model the
design space for verified controller synthesis.

\begin{definition}{\acf{MDP}.}
  \label{def:mdp}
  Given all distributions $Dist(\cdot)$ over a sort~(e.g.\xspace an action
  alphabet $A_{\mathcal{P}}$ of a process $\mathcal{P}$), an
  \ac{MDP} is a tuple
  $\mathcal{M} = (S,s_0,A_{\mathcal{P}},\delta_{\mathcal{P}},L)$
  with a set $S$ of states, an initial state $s_0\in S$, a
  probabilistic transition function
  $\delta_{\mathcal{P}}\colon S\times A_{\mathcal{P}} \to
  Dist(S)$, and a map
  $L\colon S \to 2^{AP}$ labelling $S$ with atomic propositions
  $AP$~\citep{Kwiatkowska2007-StochasticModelChecking}.
\end{definition}

Given a map $A\colon S \to 2^{A_{\mathcal{P}}}$, $|A(s)| > 1$
signifies non-deterministic choice in $s$.  Choice resolution 
for $S$ forms a policy.  
\begin{definition}{Policy.}
  \label{def:policy}
  A \emph{policy} is a map
  $\pi\colon S \to Dist(A_{\mathcal{P}})$ s.t.\xspace\
  $\pi(s)(a)>0 \Rightarrow a\in A(s)$.  $\pi$ is
  \emph{deterministic} if $\forall s\in S$ $\exists a\in A(s)\colon$
  $\pi(s)(a) = 1 \land \forall
  a'\in A_{\mathcal{P}}\setminus\{a\}\colon \pi(s)(a') = 0$.
\end{definition}
In this paper, we restrict our self to the consideration of
deterministic policies.\footnote{More precisely, we only consider
  memoryless policies, with some restrictions on \ac{MDP} policy
  synthesis, however, not relevant for our purposes.}  Let
$\Pi_{\mathcal{M}}$ be the space of all such policies for $\mathcal{M}$.
Then, \emph{action rewards} defined by a map
$r^q_{action}\colon S \times A_{\mathcal{P}} \to \mathbb{R}_{\geq 0}$ allow the
comparison of policies in $\Pi_{\mathcal{M}}$ based on a quantity $q$.

Verification of $\mathcal{M}$ can be done using \ac{PCTL} whose
properties over $AP$ are formed by
\begin{align*}
  \phi
  \textstyle
    ::=
    \mathit{true}
    \mid ap
    \mid \neg \phi
    \mid \phi \land \phi
    \mid \mathop{\textbf{E}} \varphi
    \mid \mathop{\textbf{A}} \varphi
    \mid \mathop{\textbf{P}}_{\sim b\mid [\min\mid\max]=?}\phi
    \qquad\text{and}\qquad
  \varphi
    \textstyle
    ::= \mathop{\textbf{X}} \phi
    \mid \phi \mathop{\textbf{U}}^{[\sim b]} \phi
\end{align*}
with $ap\in AP$; an optional bound $b\in\mathbb{N}_+$ for
$\mathop{\textbf{U}}^{[\sim b]}$ with $\sim\;\in\{<,\leq,=,\geq\}$; the
quantification operators $\mathop{\textbf{P}}_{\sim b\mid [\min\mid\max]=?}\phi$
to verify (or with $=?$, to quantify) probabilities, and
$\mathop{\textbf{S}}_{\sim b\mid =?}\mathit{ap}$ to determine long-run
probabilities of an atomic proposition $\mathit{ap}$.\footnote{With
  $[\cdot]$ and $\cdot\mid\cdot$, we denote optional and alternative
  syntactic choice in sub- or superscript language elements.
  $[\cdot]$ is also a mandatory notational element to encapsulate a
  formula following a temporal operator.}  States correspond to
valuations of state variables of type $\mathbb{B}$, $\mathbb{N}$, or
$\mathbb{R}$.  Hence, propositions in $AP$ are of the form $x\sim f$
for a variable $x$ and a function (or constant)
$f\colon S\to \mathbb{R}\cup\mathbb{B}$.  We use the abbreviations
$\mathit{false}\equiv\neg\mathit{true}$,
$\mathop{\textbf{F}}\phi \equiv \mathit{true}\mathop{\textbf{U}}\phi$,
$\mathop{\textbf{G}}\phi \equiv \neg\mathop{\textbf{F}}\neg\phi$, and
$\phi\mathop{\textbf{W}}\psi\equiv \phi\mathop{\textbf{U}}\psi \lor
\mathop{\textbf{G}}\phi$.  The \ac{PCTL} extension
$\mathop{\textbf{R}}^q_{\sim b\mid[\min\mid
  \max]=?}[\mathop{\textbf{F}}\phi\mid\mathop{\textbf{C}}^{[\sim b]}]$ calculates
reachability rewards~($\mathop{\textbf{F}}\phi$) and (optionally bounded)
accumulative action rewards~($\mathop{\textbf{C}}^{[\sim b]}$).  With
$\mathcal{M}\models\phi$ ($s\models\phi$), we state that the \ac{MDP}
$\mathcal{M}$ from state $s_0$ (the state $s\in S$) \emph{satisfies}
the property~$\phi$.  Let $\sem{\phi}_S$ denote the largest subset
$S'\subseteq S$ with $\forall s\in S'\colon s\models\phi$ for a state
predicate $\phi$.

For the checking of a recorded trace $t$ of a system, we use fragments
of linear and metric temporal logic whose properties over $AP$ are
formed by the propositional fragment as defined above and by
$\mathop{\textbf{F}}$, $\mathop{\textbf{G}}$, and $\mathop{\textbf{U}}_I$, with a finite
interval $I\subset\mathbb{R}_{\geq 0}\times\mathbb{R}_{\geq 0}$.  Comprehensive treatments
of \ac{PCTL}, linear and metric temporal logic can be found in, for
example, \citet{Kwiatkowska2007-StochasticModelChecking,
  Basin2015-MonitoringMetricFirst} and
\citet{Baier2008-PrinciplesModelChecking}.

The concise construction of $\delta_{\mathcal{P}}$, the behaviour of
$\mathcal{P}$, can be facilitated by using a flavour of \ac{pGCL}, for
example, as implemented in the \textsc{Prism}\xspace tool
\citep{Kwiatkowska2007-StochasticModelChecking,
  Kwiatkowska2011-PRISM4Verification}.  Guarded commands have the form
$[\alpha]\; \gamma \longrightarrow \upsilon$ with an event (or action)
label $\alpha$ and a probabilistic update $\upsilon$ applicable to
$s\in S$ only if $s\models\gamma$, where $\gamma$ is an expression in
the propositional fragment of \ac{PCTL}.\footnote{We use
  $\longrightarrow$ to separate guard and update expressions and
  $\rightarrow$ both for logical implication and the definition of
  mappings.}  With $+$ for probabilistic choice and $\&$ to compose
assignments, general updates are defined as
$\upsilon ::= \pi_1\colon \upsilon_1 + \dots + \pi_n\colon \upsilon_n$
with $\Sigma_{i\in 1..n}\pi_i = 1$ and $\upsilon_i$ being a multiple
assignment
$\upsilon_i ::= x_{i_1}'=f_{i_1} \mathbin{\&}\dots\mathbin{\&}
x_{i_n}'=f_{i_n}$ to state variables $x_k$ based on functions $f_j$.

\subsection{Risk Modelling for Controller Design}
\label{sec:safan}

We view an application~(e.g.\xspace\Cref{sec:running-example}) as a process,
$\mathcal{P}$, monitored and influenced by a safety controller to
mitigate hazards and prevent accidents.  Critical events, such as
accidents (or mishaps), their causes, and causal factors~(e.g.\xspace
hazards), are state properties.  We express these properties as
subsets of $S$.  In particular, \emph{mishaps} $\mishp[F]\subset S$
are undesired states~(e.g.\xspace all states where a person is injured by
welding sparks).  For a mishap $\mishp\subset\mishp[F]$, we further
define a subset $\Xi_{\mishp}\subset S$ from which $\mishp$ is
reachable, for example, all states in $S$ where the operator is near
the spot welder while the latter is active.  We call $\Xi_{\mishp}$
the \emph{causes} of $\mishp$.  Causes are intersections of
\emph{causal factors}, in particular, factors related to the subject
of protection~(e.g.\xspace the operator to be protected by the safety
controller when being near the spot welder) and a \emph{hazard}
$\activ$ as a causal factor related to the system~(e.g.\xspace the spot
welder being active; \citealt{Leveson1995-SafewareSystemSafety,
  Leveson2012-EngineeringSaferWorld}).  We call causes latent or
\emph{controllable}\footnote{As opposed to immediate causes with
  limited or no risk handling controls.} if there are sufficient
resources to prevent the accident~(e.g.\xspace time for removing $\activ$ by
transition to $S\setminus\Xi_{\mishp}$).
Controllability can be justified, for example, by assuming that if the
spark flow is low there is a small time span left for the spot welder to be
stopped or the operator to leave without leading to an accident.
$\activ$ can also refer to states in $S\setminus \Xi_{\mishp}$ being
critical because certain events~(e.g.\xspace an operator approaches the
spot welder) cause a transition to $\Xi_{\mishp}$, and possibly $\mishp$,
if $\activ$ stays active, further conditions hold, and no safety
measures are put in place promptly.  Below, in
\Cref{sec:refin-fact-modell}, we will use state propositions to
identify the discussed critical events.

\label{sec:yap}
Based on these notions, risk modelling can be facilitated by
specifying risk factors and combining them into \emph{risk
  structures}~\citep{Gleirscher2021-RiskStructuresDesign}.  A
\emph{risk factor} $\rfct$ is a \ac{LTS} modelling the life cycle of a
critical event in terms of phases.  In its basic form, $\rfct$ has the
phases inactive~($\inact$), active~($\activ$), mitigated~($\mitig$),
and mishap~($\mishp$).  Transitions between these phases signify
endangerment events~($\acend{}$) as well as mitigation~($\acmit{}$)
and resumption~($\acres{}$) actions.  A risk structure from a factor
set $F$~(e.g.\xspace column \textbf{Id} in \Cref{tab:riskana}) operates over
a \emph{risk (state) space}~$R(F) = \bigtimes_{\rfct\in F}\Phases$
with $\Phases = \{\inact,\activ,\mitig,\mishp\}$.
Furthermore, let $\Xi\subset S$ be the set of states labelled with at
least one cause, describing the abstract state where any critical
event has at least been sensed by the controller~(e.g.\xspace $\rfct[HC]$
with its handling about to start, i.e.,\xspace the controller transitioning
from $\inact[HC]$ to $\activ[HC]$).  We call $\Xi$ the
\emph{non-accident $F$-unsafe region}.
One can hypothesise relationships between critical events using factor
dependencies.  Such relationships can be identified and justified by,
for example, a hazard operability study, a failure mode effects
analysis, or a fault tree analysis.  Further details about risk
structures will be introduced along with our approach explained below
in \Cref{sec:approach}.

\subsection{Digital Twins} 
\label{subsec:what-is-a-digital-twin}

We focus on digital twins as a platform for controller deployment and
validation.  Digital twinning is a key Industry 4.0 technology for
enabling industrial automation, smart processes, and process
autonomy~\citep{Negri2017,Bolton2018}.  Together with the Internet of
Things and machine-to-machine communication, digital twinning enables
both the creation of a better data infrastructure and the adoption of
smart manufacturing technologies.
A digital twin provides greater access to data relating to, and
control over, a physical system, and has particular value in the
design, implementation, and evaluation of processes and safety
controllers.

\cite{Kritzinger2018} and \cite{Tao2018digital} define
a digital twin as:
\begin{quote}
  ``A digital twin is an integrated multi-physics, multi-scale,
  probabilistic simulation of a complex product and uses the best
  available physical models, sensor updates, etc., to mirror the life
  of its corresponding twin."
\end{quote}

\begin{wrapfigure}[12]{r}{.5\columnwidth}
  \includegraphics[width=.5\columnwidth]{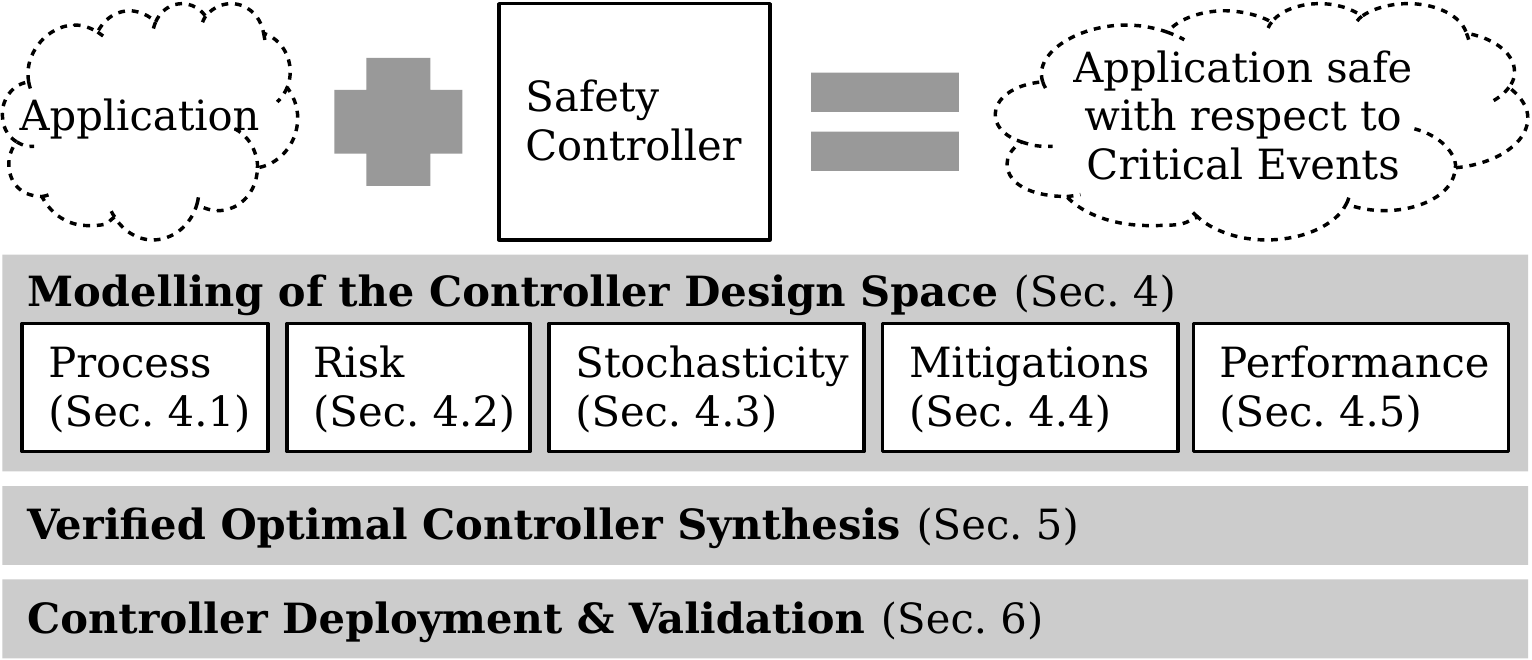}
  \caption{Stages of the proposed approach to safety controller
    synthesis}
  \label{fig:procedure}
\end{wrapfigure}

More specifically, a digital twin is a digital representation of a
physical system that operates in parallel\footnote{A digital twin must
  be able to operate synchronously with the physical system, but
  asynchronous operation is also permissible.} with the real system.
This concurrency can persist throughout the life-cycle of the physical
system.  Communication between the physical twin and its digital
representation is bilateral, and as a result, both can be mutually
informed by real-world or simulated sensor data, requested actions and
decisions.  As a means for safety verification and analysis, a digital
twin presents:
\begin{inparaenum}[(i)]
\item A faithful representation of the process domain and state space,
\item a means to interrogate and collect data that may not be readily
  available from the physical system (independent of hardware
  limitations), and
\item an interface to the physical twin through which real-world
  responses to new safety procedures can be demonstrated.
\end{inparaenum}

 \section{Modelling for the Synthesis of Safety Controllers}
\label{sec:approach}

\Cref{fig:procedure} provides an overview of the proposed approach to
the synthesis of safety controllers.  The main idea is that the
control engineer designs a safety controller on top of an application,
in this instance, comprising activities where humans and robots
collaborate.  The intention of the deployed controller is to increase
safety with respect to the critical events under consideration.

In the \emph{modelling} stage, the control engineer creates several
models to obtain a design space including controller candidates to
select from during synthesis.  When modelling the \emph{process} of
the application, the engineer as a domain expert describes the actions
performed by any of the actors in the application.  The engineer then
performs a risk analysis resulting in a \emph{risk model} that informs
the process model with a notion of operational risk.  The abstraction
chosen for the model and a usual lack of knowledge about process
details require the integration of \emph{stochasticity} and
\emph{performance} estimates into the model.  With the risk- and
performance-informed stochastic process model, the engineer can
specify controller behaviour in form of a \emph{mitigation} model.  In
the \emph{controller synthesis} stage, an abstract controller is
automatically synthesised from the design space according to risk- and
performance-based optimality criteria.  Finally, in the
\emph{deployment and validation} stage, this controller is translated
automatically into an executable form.  The three stages are supported
by the tools \textsc{Yap}\xspace, \textsc{Prism}\xspace, \textsc{evoChecker}\xspace, and the \acl{CDT}.  These
stages are detailed in the following sub-sections and illustrated with
several examples from our case study introduced in
\Cref{sec:running-example}.

\subsection{Modelling Processes}
\label{sec:process-modelling}

\begin{wrapfigure}{r}{.5\columnwidth}
  \includegraphics[width=.5\columnwidth]{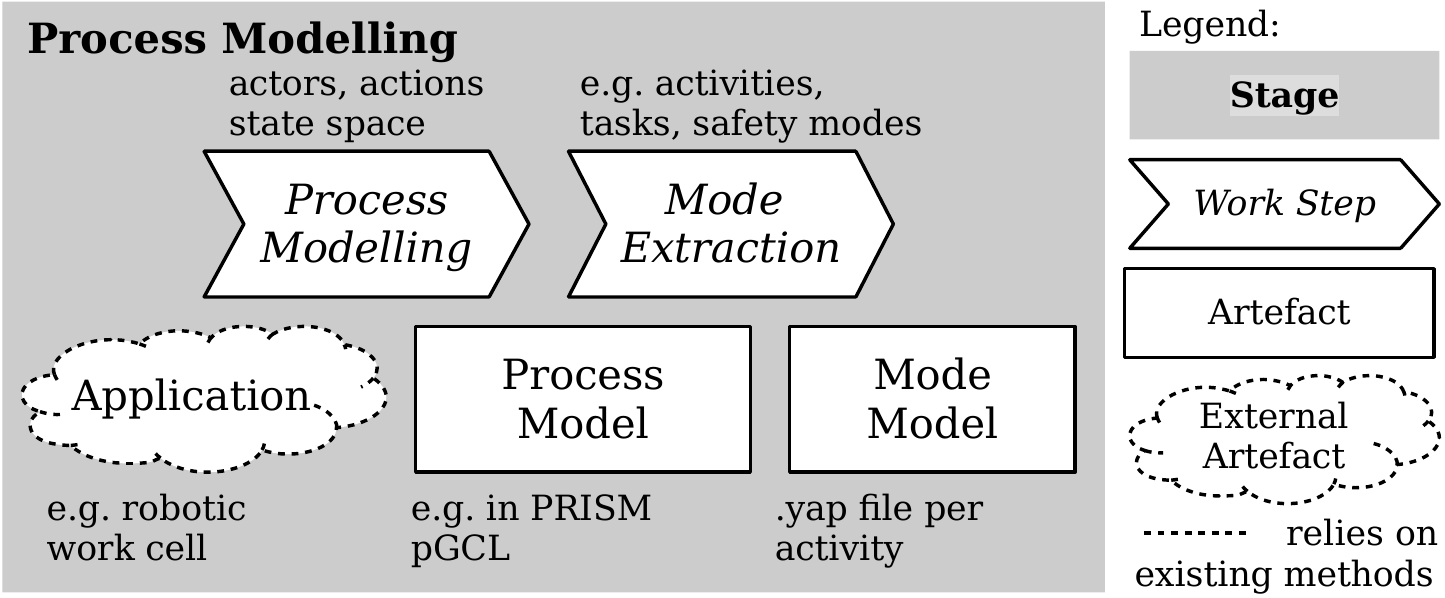}
  \caption{Overview of the work steps and artifacts of the process modelling stage}
  \label{fig:procedure-process}
\end{wrapfigure}

For process modelling (\Cref{fig:procedure-process}), we use \ac{pGCL}
to specify the actions of the process.  The process model defines the
state space to be manipulated by these actions and includes the
actions of the cobot and the environment including human operators.
We group actions into \emph{activities} as abstractions of the
process.  Activities describe certain tasks and facilitate the
transition to risk modelling by structuring hazard identification and
the creation of a hazard list.  The \emph{mode model} resulting from
this abstraction step is an abstract \ac{LTS}.

We describe the process, $\mathcal{P}$, as a set of guarded commands,
distinguishing \emph{actions} of relevant actors~(e.g.\xspace a robot arm, a
spot welder, an operator) and the safety controller from \emph{events}
of a sensor module and shared ``manipulables''~(e.g.\xspace workpiece
support).  Following \Cref{sec:probmc}, the structure of the guarded
commands describing the behaviour in $\mathcal{P}$ follows the pattern
\[
  [\alpha]\; \dots\land \gamma \longrightarrow \upsilon \;+\; \dots
\]
with an action label $\alpha$, an action-specific condition $\gamma$
and a generic update $\upsilon$ (see \Cref{sec:probmc}) being part of
the overall guard and update expressions.
The state space $S$ is built from discrete
variables~(cf.\xspace\Cref{lst:statespace}) capturing the world state~(e.g.\xspace
robot location, workbench status), sensory inputs~(e.g.\xspace range finder),
control outputs~(e.g.\xspace robot behaviour, notifications), user
inputs~(e.g.\xspace start button), and modes~(e.g.\xspace activities, safety modes).

\subsubsection{Process Execution from a Controller's Perspective}
\label{sec:execution-semantics}

\begin{wrapfigure}{r}{.4\columnwidth}
  \includegraphics[width=.4\columnwidth]{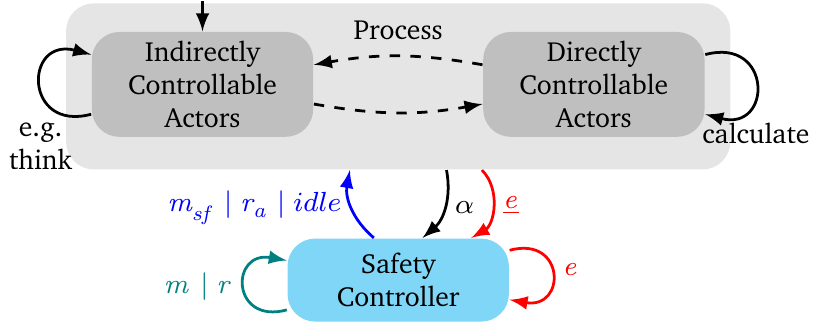}
  \caption{Execution from the viewpoint of the safety controller.  The
    execution is coordinated by passing a single token between two
    groups of actors (dashed arcs) performing actions in the process.
    The controller can perform an update or issue control inputs to the
    process after every atomic event occurring in the process.
    \label{fig:int-sem}}
\end{wrapfigure}

To account for the interaction of the safety controller with the
process, we include a fair cyclic execution scheme into the process
model.  This scheme emulates the simultaneity of the controller and
the process.  Execution steps alternate between the controller and a
set $\mathcal{A}$ of actors, ensuring in each cycle that each actor
can take its turn~(\Cref{fig:int-sem}).  Regarding the way the
controller can influence the process, we distinguish between directly
controllable actors~(e.g.\xspace cobot, spot welder) and indirectly or not
controllable actors~(e.g.\xspace human operator).  Both kinds of actors can
perform logical and physical actions following two corresponding
command patterns:
\begin{align}
  \label{eq:log-act}
  [\alpha]\; &\mathit{ok}_p \land \dots \land 
  \gamma \longrightarrow \upsilon \;+\; \dots
  \tag{logical action}\\
  \label{eq:phys-act}
  [\alpha]\; &\mathit{ok}_p \land \dots \land \gamma \longrightarrow \upsilon
  \mathbin{\&} t'=sc \;+\; \dots
  \tag{physical action}
\end{align}
where $\mathit{ok}_p$ guards the turn of an actor $p$ and $t$ stores the token
passed between the safety controller and the other actors.  $t'=sc$
denotes the safety controller's turn.  $\mathit{ok}_p$ can include a condition
for terminating execution when reaching a final~(or goal) state,
resulting in $\mathit{ok}_p \equiv t=p \land \neg\mathit{final}$.  The
controller can stay idle or, most eagerly, intervene whenever one of
the actors has completed an action.  This scheme is more restrictive
than the CSP-style\footnote{Following the synchronous interleaving
  semantics of Hoare's Communicating Sequential Processes.}
generalised parallel composition of concurrent processes available by
default in tools such as
\textsc{Prism}\xspace~\citep{Kwiatkowska2011-PRISM4Verification}.

\subsubsection{Modelling Activities and Safety Modes}
\label{sec:activities}

To facilitate the design of a powerful class of safety controllers, we
organise actions~(e.g.\xspace grab work piece, move robot arm to spot welder)
of $\mathcal{P}$'s controllable actors using mode variables, here, one
variable for the engagement of each actor in an
activity~(e.g.\xspace\yapid{ract}) and one for the safety mode of the whole
process~(\yapid{safmod}).  Mode variables group actions and, from a
controller perspective, allow high-level process control by filtering
enabled actions.  Thus, the structure of the guarded commands for
$\mathcal{P}$ is refined according to the pattern
\[
  [\alpha]\; \mathit{ok}_p \land \gamma_{\mathit{sm}} \land \gamma_a \land
  \gamma \longrightarrow \upsilon\, [\mathbin{\&}\, t'=sc] \;+\; \dots
\]
where $\gamma_a$ guards the enabling of actions in certain
activities~(e.g.\xspace exchange work piece, \Cref{fig:setting-activities})
and $\gamma_{\mathit{sm}}$ in certain safety modes~(e.g.\xspace speed \&
separation monitoring).  We obtain safety- and task-aware actions by
conjoining $\gamma_{sm}\land\gamma_{a}$.  The update of $t$ is
optional to combine certain actor-internal updates into atomic
actions.

\begin{example}
  \label{exa:robotactions}
  The following listing describes part of the enumerations used to
  define the discrete state space $S$ of $\mathcal{M}$.  

  \begin{lstlisting}[language={prism},label=lst:statespace]
// spatial locations
const int atTable = 0; const int sharedTbl = 1; const int inCell = 2; const int atWeldSpot = 3;
// range finder signals
const int far = 0; const int near = 1; const int close = 2;
// notification signals
const int ok = 0; const int leaveArea = 1; const int resetCtr = 2;
  \end{lstlisting}
  The following commands specify two actions,
  \yapid{r\_moveTo\-Table} and \yapid{r\_grab\-LeftWorkpiece}, for the
  actor \yapid{robotArm} in the activity \yapid{exchWrkp}.
  \vspace{-2em}
  \begin{lstlisting}[aboveskip=-2em,language={prism},
  emph={ract,exchWrkp,r\_moveToTable,r\_grabLeftWorkpiece},
  label=lst:robotarm,multicols=2]
[r_moveToTable] OK_wc 
  & (safmod=normal|safmod=ssmon|safmod=pflim) 
  & ract=exchWrkp & (rloc != sharedTbl) 
  & (((wps!=right) & reffocc=1) | (wps=left & (reffocc=0))) 
  -> (rloc'=sharedTbl)&(turn'=sc);
[r_grabLeftWorkpiece] OK_wc 
  & (safmod=normal|safmod=ssmon|safmod=pflim|safmod=hguid) 
  & ract=exchWrkp & rloc=sharedTbl & reffocc=0 & wps=left 
  -> (reffocc'=1)&(wpfin'=0)&(wps'=empty)&(turn'=sc);
  \end{lstlisting}
\end{example}

\subsection{Modelling Risk}
\label{sec:risk-modelling}

\begin{wrapfigure}{r}{.5\columnwidth}
  \includegraphics[width=.5\columnwidth]{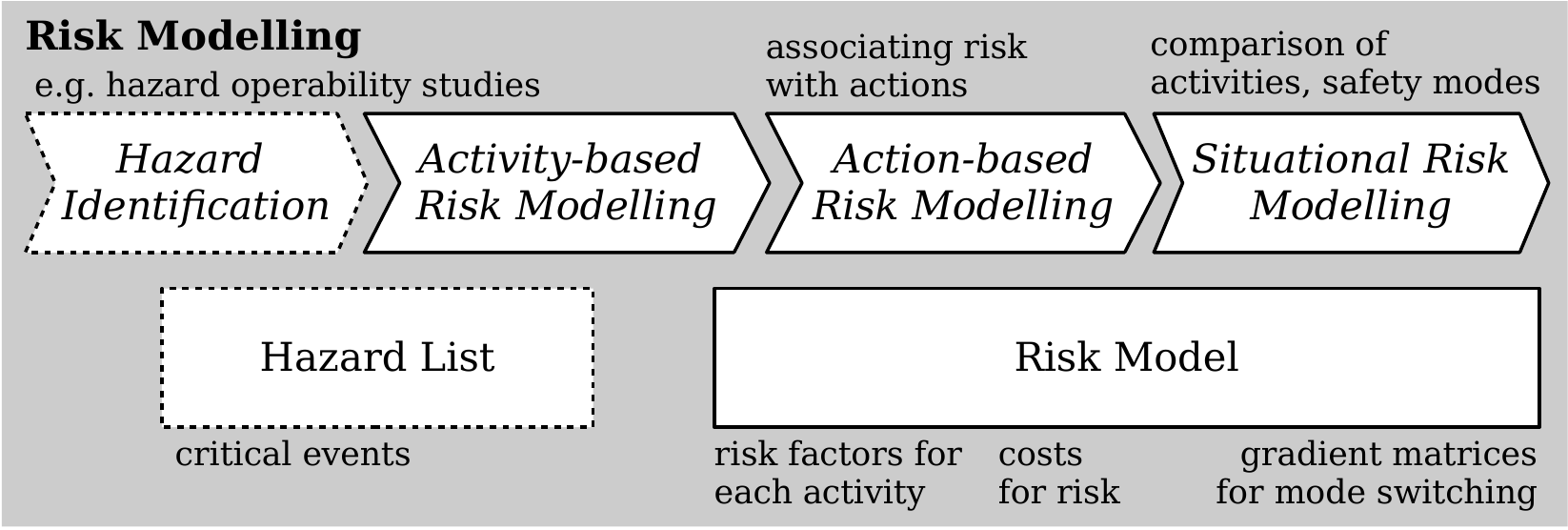}
  \caption{Overview of the work steps and artifacts of the risk modelling stage}
  \label{fig:procedure-risk}
\end{wrapfigure}

For our synthesis approach to lead to correct and effective
controllers, we need an expressive risk
model~(\Cref{fig:procedure-risk}).  To obtain such a model, we
translate the hazard list into a set of risk
factors~(\Cref{sec:activity-based-risk}).  For this, we transcribe
safety analysis results into factor \acp{LTS}~(\Cref{sec:yap}).  Then,
we define a risk profile for each of the process
actions~(\Cref{sec:action-based-risk}).  For each hazard in the hazard
list, the \emph{risk profile of an action} describes the risk that a
performance of the action results in an accident related to this
hazard.  The final step of risk modelling consists of capturing
risk-related situational change~(i.e.,\xspace a change of activity or safety
mode) in the process with a \emph{risk
  gradient}~(\Cref{sec:situ-risk-modell}).  For this, we associate a
numerical measure with each activity transition~(i.e.,\xspace each situational
change), that describes the change in overall risk level.  We proceed
analogously with the definition of safety modes and the corresponding
risk assessment.

\subsubsection{Refined Risk Factors}
\label{sec:refin-fact-modell}

\begin{wrapfigure}{r}{.5\columnwidth}
  \footnotesize
  \includegraphics[width=.5\columnwidth]{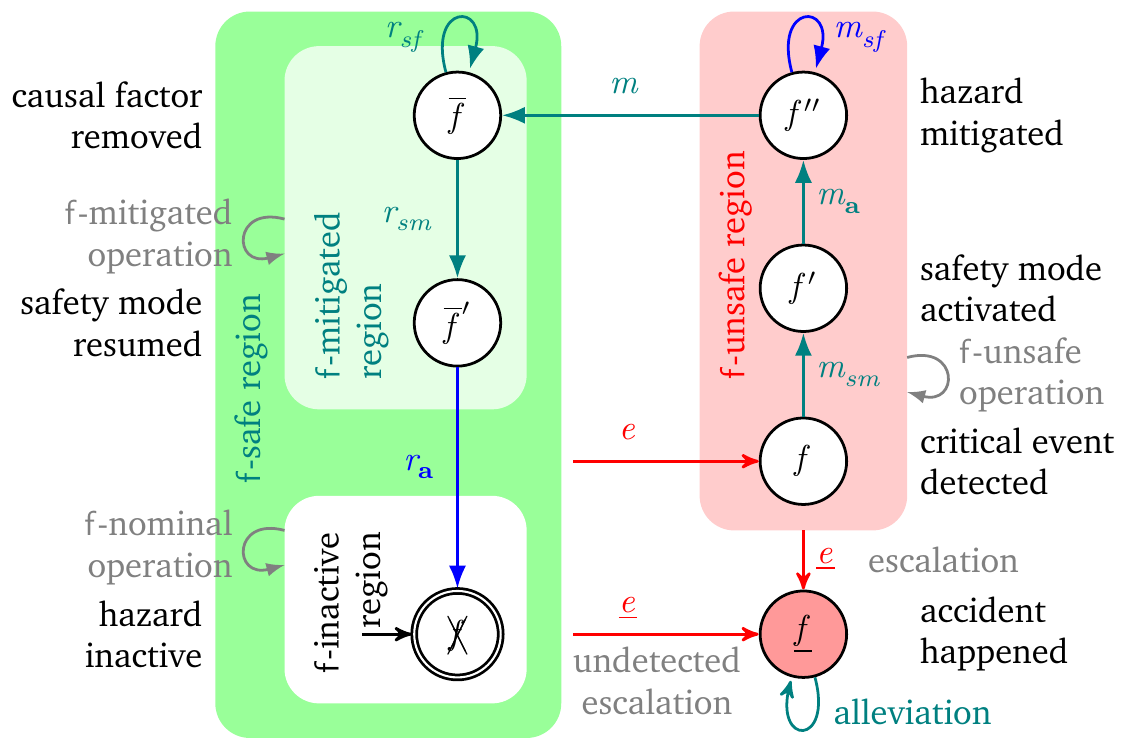}
  \caption{Refined notion of a risk factor $\rfct$.  The states of
    this \ac{LTS} are called phases, the transitions are labelled with
    events and actions.  Endangerment $\acend{}$ (monitored) and
    accident $\acmis{}$ events; actions $\acmit[*]{}$ and $\acres[*]{}$ of the
    safety controller with the actions $\acmit[sf]{}$ and $\acres[a]{}$ where the
    controller interacts with the process waiting for a response; in
    \textcolor{gray}{gray}, arbitrary process actions not controllable
    or observable by the safety controller.}
  \label{fig:riskfactor}
\end{wrapfigure}

Shown in \Cref{fig:riskfactor}, we develop and use a refinement
of the notion of a risk factor introduced in \Cref{sec:yap}.
As before, for a factor $\rfct$, we refer to any state of
$\mathcal{P}$ where $\rfct$ is inactive as the \emph{$\rfct$-inactive
  region} or, equivalently, as $\inact$.
Any state of $\mathcal{P}$ where $\rfct$ has occurred and any causal
factor is still active is subsumed by the \emph{$\rfct$-unsafe
  region}, which includes three phases: hazard detected
($\activ$), safety mode activated ($\activ'$), and hazard
  mitigated ($\activ''$).
The $\rfct$-\emph{mitigated region} refers to any state deviating from
$\mathcal{P}$'s nominal state but with any causal factor of $\rfct$
removed.  The $\rfct$-mitigated region includes the phase
$\mitig$, reached when causal factors of $\rfct$ have been removed,
and $\mitig'$, reached after deactivating a safety function and
resuming from the current safety mode.
The phases $\mitig$, $\mitig'$, and $\inact$ together constitute the
$\rfct$-\emph{safe region} of $\mathcal{P}$.

When a critical event~$\acend{}$ is detected in the $\rfct$-safe
region, $\rfct$ is switched to $\activ$.  For a critical event, we
distinguish its ground truth predicate $\chi$~(i.e.,\xspace the cause) from
the detector~(or monitoring) predicate $\zeta$~(i.e.,\xspace the sensor) where
$\chi\Leftrightarrow\zeta$ in case of perfect sensing.  \emph{Critical
  event detection} follows the pattern
\[
  [\acend{}]\; \mathit{ok}_S
  \land \zeta
  \land \mathop{rel}(a,\rfct)
  \land \neg(\activ\lor\mishp)
  \longrightarrow \activ
\]
where $\mathop{rel}(a,\rfct)$ is a relevance indicator determining
whether to react on $\zeta$ in an activity $a$ or, more generally, in
a particular situation.  With $\neg(\activ\lor\mishp)$, we ignore
re-occurrences while $\rfct$ is active or after an accident.
\emph{Idling} of the safety controller~(\Cref{fig:int-sem}) is
captured by
\[
  [\mathit{idle}]\; \mathit{ok}_S \land \neg\chi
  \land (\inact\lor\mitig) \longrightarrow t'=\mathit{next}(p)
\]
where $\mathit{next}(p)$ passes the token from the actor $p$ to the
next actor in $\mathcal{P}$~(\Cref{fig:int-sem}).  The other actions
will be explained in \Cref{sec:mitigation-modelling}.

\paragraph{Activity-based Risk.}
\label{sec:activity-based-risk}

Factor \acp{LTS} guide the formalisation of hazards, their causes, and
mishaps and the events in the causal chain~(e.g.\xspace a mishap event leads
to a mishap state).  This way, factors support the design of
mitigations to reduce accidents, and alleviations to reduce
consequences.  Hence, all critical events related to an activity
should be translated into a factor set.

\paragraph{Action-based Risk.}
\label{sec:action-based-risk}

We define an action multi-reward structure
$ r^{\mathit{risk}}_{\mathit{action}}\colon S \times
A_{\mathcal{P}}\times F \to\mathbb{R}_{\geq 0} $ over $\mathcal{M}$ to measure
overall and factor-specific risk in $\mathcal{P}$.  Action rewards
are guarded, requiring $\rfct$ being active.  For example, if
$r^{\mathit{risk}}_{\mathit{action}}(s,a,\rfct) > 0$ then $\rfct$ is
active in state~$s$.

\subsubsection{Situational Risk}
\label{sec:situ-risk-modell}

The decision space of the safety controller for mitigations and
resumptions includes choices from sets of safety modes
($S_{\mathit{sm}}$) and activities ($S_a$) to switch to from a
particular mode and activity.  To simplify the design space in
$\mathcal{M}$, we resolve this choice during generation
(\Cref{sec:design-space-gener}) using a \emph{categorical risk
  gradient}
$\nabla r = [\partial r/(a\to a'), \partial r/(sm\to sm')]^T$.  For
categorical variables $x$, we allow the convention
$\partial y/(x\to x')$ denoting the change of $y$ when $x$ changes its
value from $x$ to $x'$.

We implement $\nabla r$ with two skew-diagonal matrices
$\mathbb{R}^{|S_{sm}|\times|S_{sm}|}$ and
$\mathbb{R}^{|S_a|\times|S_a|}$, assuming that they can be (manually)
derived from safety analysis based on the following justification.
Assume that, in the activities $a, a'\in S_a$, actors vary in physical
movement, force, and speed.  If $a$ means more or wider movement,
higher force application, or higher speed than in $a'$, then a change
from $a$ to $a'$ will likely reduce risk.  Hence,
$\partial r/(a\to a') \geq 0$.
Similarly, assume that the safety modes $sm, sm' \in S_{\mathit{sm}}$
vary $\mathcal{P}$'s capabilities by relaxing or restricting the range
and behavioural shape of the permitted actions.  If $sm$ relaxes
capabilities more than $sm'$, then a change from $sm$ to $sm'$ will
likely reduce risk.  Again, $\partial r /(sm\to sm') \geq 0$.
Skew-diagonality of the matrices provides that
$\partial r/(sm'\to sm) \leq 0$.  The matrices for $\nabla r$ can be
specified as part of a \textsc{Yap}\xspace model.

\begin{figure*}
  \centering
  \includegraphics[width=\textwidth]{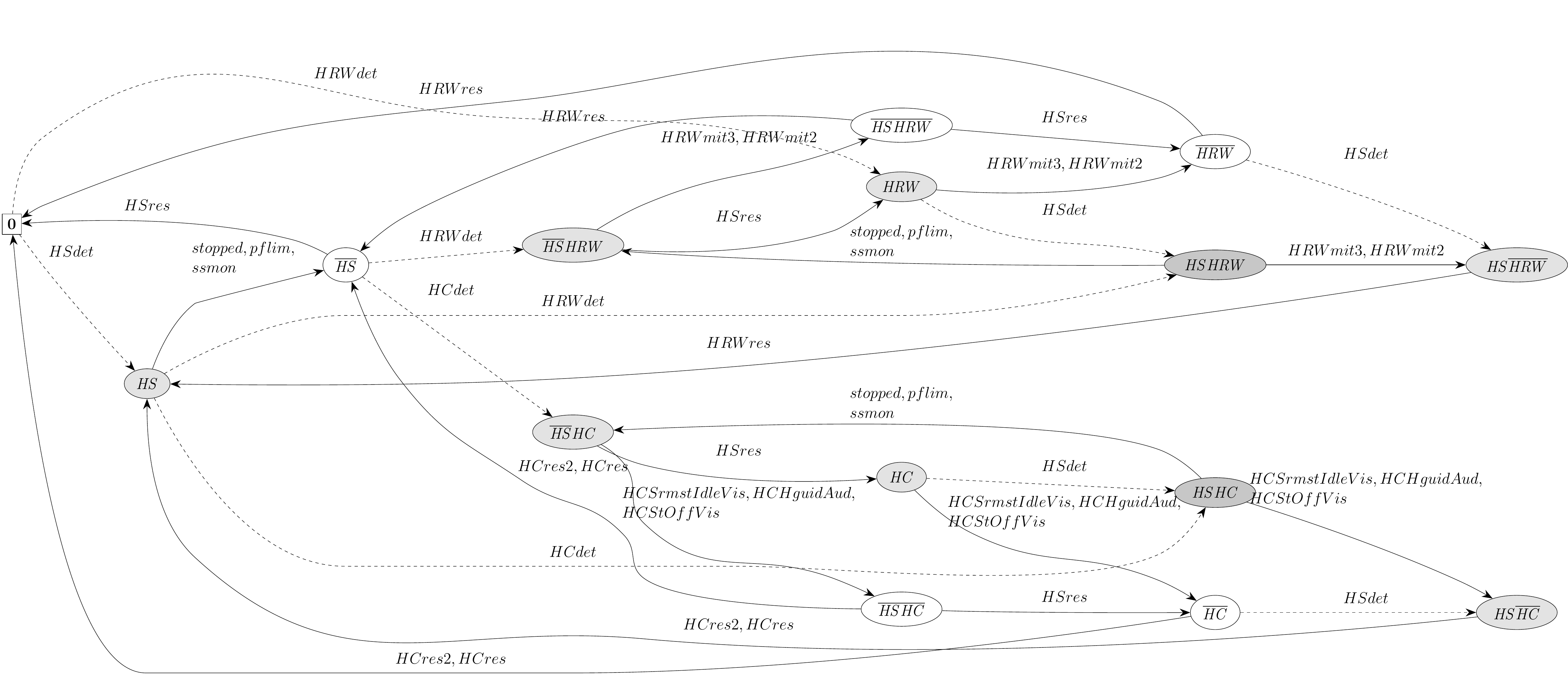}
  \caption{Risk graph for \Cref{exa:riskfactor:yap} from factors
    $\rfct[HC]$, $\rfct[HRW]$, and $\rfct[HS]$, 
    with 15 risk states, serving
    as a specification to be refined by a synthesised controller.  Dependencies
    ``$\rfct[HC]$~\yapkeyw{prevents}~$\rfct[HRW]$'' and
    ``$\rfct[HRW]$~\yapkeyw{prevents}~$\rfct[HC]$'' encode the
    assumption of one operator in the work cell.  Detectors \rgsym{HSdet} or
    \rgsym{HRWdet} (dottet arcs) instantiate endangerments, and
    \rgsym{stopped}, \rgsym{pflim}, \rgsym{ssmon}, and \rgsym{HSres} exemplify mitigations and resumptions
    (solid arcs).  The darker shaded a risk state, the more dangerous
    it is, qualitatively.
    \label{fig:riskgraph}}
\end{figure*}

\begin{example}
  \label{exa:riskfactor}
  We instantiate $\rfct$ according to \Cref{fig:riskfactor} for the
  hazard~$\rfct[HC]$ from \Cref{tab:riskana}.
  When an operator approaches an active spot welder, an event $\acend{HC}$
  is detected, activating $\rfct[HC]$ by a transition to a state where
  the predicate $\activ[HC]$ holds, a state in
  $\semRS{\activ[HC]} \subset R(F)$.
  The safety controller will then start with performing mitigations to
  reach phase $\mitig[HC]$.  The handling of $\rfct[HC]$ includes the
  switching to a speed \& separation monitoring mode, issuing an
  operator notification, and waiting for the operator's response.
  From $\mitig[HC]$, the controller can continue with
  resumptions~(e.g.\xspace switching from speed \& separation monitoring back
  to normal) to finally return to phase $\inact[HC]$ where both
  $\activ[HC]$ and $\mitig[HC]$ are false.
  Further endangerments~(e.g.\xspace erroneous robot movement) may
  reactivate $\rfct[HC]$ from phase $\mitig[HC]$ or $\mitig[HC]'$.
  An accident $\acmis{HC}$ with the spot welder or robot arm, leading to
  phase $\mishp[HC]$, can occur, for example, because of a faulty
  range finder responsible for the detection of $\acend{HC}$ or too
  slow mitigations $\acmit[sm]{HC}$ and $\acmit[a]{HC}$.  In
  $\mishp[HC]$, alleviations~(e.g.\xspace flexible robot surfaces, protective
  goggles) can reduce certain consequences.
  We start encoding $\rfct[HC]$ in a \textsc{Yap}\xspace model below.
  \label{exa:riskfactor:yap}
\begin{lstlisting}[language={yapscript}, emph={HCdet,HCmit,HCres}, label=lst:moving]
HC desc "(H)uman (C)lose to active spot welder and cobot working"
  requiresOcc (HS) // imposes relationship between cause/guard conditions of HS and HC, e.g. not HS => not HC
  mitPreventsMit (HS)
  guard "hSM_PERM & hACT_WELDING & hloc=atWeldSpot"
\end{lstlisting}
  As a factor dependency, we identify $\rfct[HC]$
  \yapkeyw{requiresOcc} $(\rfct[HS])$, expressing the assumption that
  the factor $\rfct[HS]$ must have occurred prior to the activation of
  $\rfct[HC]$.  Using the \yapkeyw{guard} directive, we specify the cause
  $\chi$, the ground truth predicate for the activation of
  $\rfct[HC]$.  \Cref{fig:riskgraph} shows the resulting risk
  structure for our case study as a risk graph.
  
  \vspace{.4em}
  \noindent
  \begin{minipage}[t]{.48\columnwidth}
  The risk profile for the robot arm actions and the factors
  $\rfct[HC]$ and $\rfct[HS]$ is encoded in the \textsc{Yap}\xspace model in an
  intuitive and compact manner.
  \begin{lstlisting}[language={yapscript}]
                        guard   risk_HC   risk_HS;
// actor: robotArm
r_moveToTable:          ""      "5"       "9";
r_grabLeftWorkpiece: 	""      "0"       "3";
r_placeWorkpieceRight:  ""      "0"       "3";
r_moveToWelder:         ""      "9"       "5";
  \end{lstlisting}
  \end{minipage}
  \hfill
  \begin{minipage}[t]{.47\columnwidth}
  Based on the activity automaton in \Cref{fig:setting-activities}, we
  encode $\nabla r$ in our \textsc{Yap}\xspace model with the two distance matrices
  \yapkeyw{act} and \yapkeyw{safmod}.
  \vspace{-2em}
  \begin{lstlisting}[aboveskip=-2em,language={yapscript},multicols=2]
distances act {
 off:       0;
 idle:      1 0; 
 exchWrkp:  3 2 0;
 welding:   5 4 2 0;
}

   
distances safmod {
 normal:   0;
 hguid:   -2  0;
 ssmon:   -1  1    0;
 pflim:   -2  0   -1  0;
 srmst:   -3 -1   -2 -1  0;
 stopped: -4 -2   -3 -2 -1  0;
}
  \end{lstlisting}
\end{minipage}
\end{example}

\subsection{Modelling Stochastic Adversarial Phenomena}
\label{sec:stochastic-modelling}

\begin{wrapfigure}[10]{r}{.35\columnwidth}
  \includegraphics[width=.35\columnwidth]{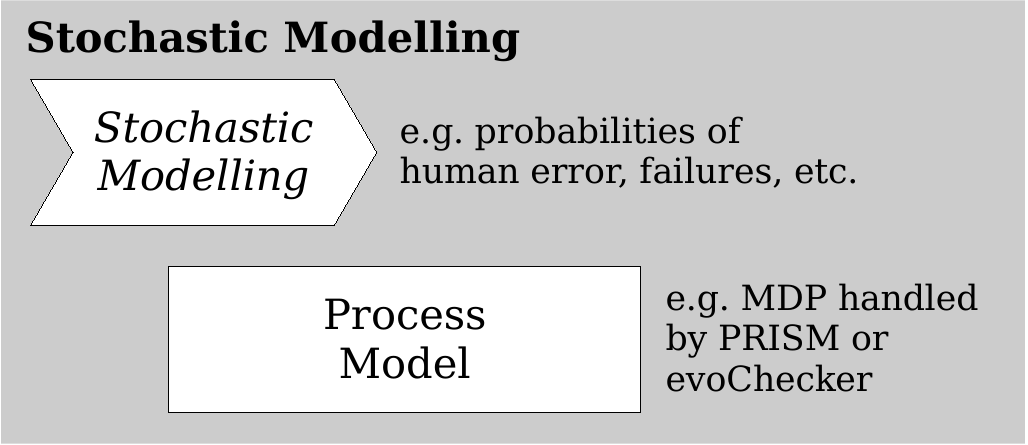}
  \caption{Overview of the work steps and artifacts of the stochastic modelling stage}
  \label{fig:procedure-stoch}
\end{wrapfigure}

In this stage~(\Cref{fig:procedure-stoch}), we integrate adversarial
stochastic phenomena into the process model.  Probabilistic choice in
$\mathcal{M}$ can be used to model various phenomena, such as
accidents, human error, and sensor failure.

\paragraph{Mishaps.}
\label{sec:acci-mod}

In the refined factor model (\Cref{fig:riskfactor}), a mishap
$\acmis{}$ leading to phase $\mishp$ is always possible, assumed to
happen more likely in the $\rfct$-unsafe region than in the
$\rfct$-safe region.  \emph{Accident-prone physical actions} follow
the two command patterns:
\begin{align*}
  [\alpha]\; \mathit{ok}_p \land \gamma \land \neg\chi
  \longrightarrow& \upsilon \mathbin{\&} t'=sc
  \\
  [\acmis[\alpha]{}]\; \mathit{ok}_p \land \gamma \land \chi \land \neg\mishp
  \longrightarrow& pr_{\mishp}\colon \mishp \mathbin{\&} \upsilon_a \mathbin{\&} t'=sc
  \;+\; (1-pr_{\mishp})\colon \upsilon_n \mathbin{\&} t'=sc
\end{align*}
with the probability $pr_{\mishp}$ of a mishap if the cause $\chi$ of
the critical event has occurred, independent of whether or not
$\rfct$'s activation~($\acend{\rfct}$) was detected.  $\upsilon_a$ and
$\upsilon_n$ can include specific updates if an accident occurs
respectively if it does not.  $pr_{\mishp}$ can be inferred from
observations, experiments, or accident statistics.  We keep using
$\gamma$ and $\upsilon$ for action-specific preconditions respectively
updates.

\paragraph{Human Error.}
\label{sec:human-error}

A human error model can be informed by hierarchical task
analysis~\citep{Stanton2006-Hierarchicaltaskanalysis}.  We introduce a
particular class of human errors into $\mathcal{M}$ using the pair of
probabilistic commands
\begin{align*}
  [\alpha_l]\; \mathit{ok}_{op} \land \neg\delta \land \gamma \longrightarrow\;&
  (sp?1:pr_{he})\colon \delta'=true
  \;+\; (sp?0:1-pr_{he})\colon t'=sc
  \\
  [\alpha_p]\; \mathit{ok}_{op} \land \delta \land \gamma
  \longrightarrow\;& \upsilon \mathbin{\&} \delta'=false \mathbin{\&} t'=sc
\end{align*}
with a predicate $sp$ specifying when action $\alpha_p$ is safe or
permitted to be performed, the probability $pr_{he}$ of an operator to
commit a specific error when this action is not safe or not permitted,
and a deontic flag $\delta$ controlling whether the logical action
$\alpha_l$ is to be reified into the physical action $\alpha_p$ with a
potentially dangerous update $\upsilon$.

\paragraph{Sensor Failure.}
\label{sec:sensor-failure}

Informed by a fault tree or failure mode effects analysis, one can
consider sensor and actuator faults in a way similar to human error.
To model \emph{fault behaviour}, we employ the pattern
\[
  [\alpha]\; \mathit{ok}_p \land \gamma \longrightarrow
  (1-pr_s)\colon \upsilon_{corr} \mathbin{\&} t'=sc
  \;+\; pr_s\colon \upsilon_{fail} \mathbin{\&} t'=sc
\]
with a probability $pr_s$ of a sensor failing to detect a specific
event implied by $\gamma$, an update $\upsilon_{corr}$ modelling the
correct behaviour of the sensor, and an update $\upsilon_{fail}$
modelling its failure behaviour.

\begin{example}
  \label{exa:stoch-phen}
  We model the accident that, with a 20\% chance, $\mishp[HC]$ follows
  $\activ[HC']$~(i.e.,\xspace $\rfct[HC]$ remains undetected because of a
   sensor fault) or $\activ[HC]$~(i.e.,\xspace the safety
  controller is not reacting timely).  For the encoding of
  $\rfct[HC]$, \textsc{Yap}\xspace's input language supports the specification of the
  actions with the mishap $\mishp[HC]$ as a bad outcome if
  $\activ[HC]$ is undetected or not mitigated timely, the probability
  of $\mishp[HC]$ under these conditions, and the severity of the
  expected consequences from $\mishp[HC]$.
  Furthermore, we model the human error that, with a 10\% chance, the
  operator enters the cell, knowing that the \yapid{robotArm} and the
  \yapid{spotWelder} are active.
  Finally, we specify as a sensor failure that the range finder as the
  detector of $\acend{HC}$ fails in 5\% of the cases when the operator
  enters the cell.
\end{example}

\begin{wrapfigure}[9]{r}{.5\columnwidth}
  \includegraphics[width=.5\columnwidth]{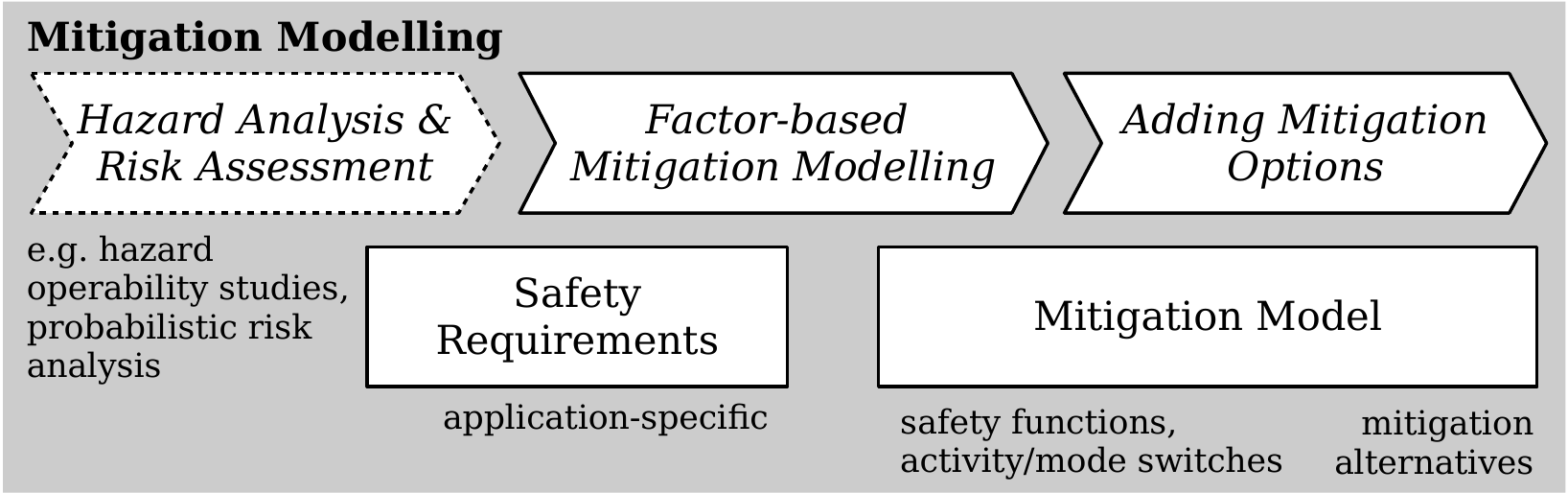}
  \caption{Overview of the work steps and artifacts of the mitigation modelling stage}
  \label{fig:procedure-mitig}
\end{wrapfigure}

The command patterns explained in this section can be combined,
offering many degrees of modelling freedom not further discussed
here. For practical examples, see also
\cite{Gleirscher2020-YAPToolSupport}.

\subsection{Modelling Mitigations}
\label{sec:mitigation-modelling}

For mitigation modelling (\Cref{fig:procedure-mitig}), we complete the
factor \acp{LTS} introduced in \Cref{sec:refin-fact-modell} with
mitigation actions.  The ability of stochastic models, such as
\acp{MDP}, to express nondeterminism supports the modelling of
alternative \emph{mitigation options}.  To extend the controller
design space, we can thus specify several such options for each
factor.  Differences in the quality of these options~(e.g.\xspace expected
nuisance and effort) can be quantified using reward structures.

The capabilities of actors in $\mathcal{P}$ determine the
controllability of critical events.  To restrict the controller design
space, we allow three kinds of actions: \emph{action filters}~(i.e.,\xspace
safety modes, cf.\xspace\Cref{sec:intro}), \emph{activity changes}~(e.g.\xspace
change from \yapid{welding} to \yapid{off}), and \emph{safety
  functions}~(e.g.\xspace interacting with the operator through warnings).
These mitigations are mirrored by corresponding resumptions.
We continue with our discussion of how mitigation and resumption
actions, according to the refined factor \ac{LTS} in
\Cref{fig:riskfactor}, are translated into \ac{pGCL}.

Let $\rfct\in F$ be the risk factor under consideration for the rest
of this section.  Given the ground truth predicate $\chi$ and the
corresponding detector predicate $\zeta$ for
$\rfct$~(\Cref{sec:refin-fact-modell}), we assume to have identified a
causal factor~$\kappa$ such that
$\neg\kappa\Rightarrow\neg\chi\land\neg\zeta$~(i.e.,\xspace an absent causal
factor eliminates the cause) and $\zeta \Rightarrow \kappa$~(i.e.,\xspace the
causal factor is detectable).  Factor dependencies, such as
\yapkeyw{requiresOcc} in \Cref{exa:riskfactor:yap}, can be used to
automatically derive part of $\zeta$~(cf.\xspace\textsc{Yap}\xspace,
\citealt{Gleirscher2020-YAPToolSupport}).

Let a \emph{mitigation option}
$(a_{t}, \mathit{sm}_{t}, \mathit{sf}) \in S_a\times S_{sm} \times SF$
for $\rfct$ with a target activity $a_t$, a target safety mode $sm_t$,
and $\mathit{sf}$ picked from a set $SF$ of safety functions.  For
each combination $(sm_{cur},a_{cur})\in S_{sm}\times S_{a}$ possible
in a state $s\in S$, the controller provides a \emph{safety mode
  switch}
\[
  [\acmit[sm]{}]\; \mathit{ok}_S \land \activ \land sm_{cur} \longrightarrow
  \mathit{sm}_{new} \mathbin{\&} \activ'
\]
and an \emph{activity switch} 
\[
  [\acmit[a]{}]\; \mathit{ok}_S \land \activ' \land a_{cur} \longrightarrow a_{new}
  \mathbin{\&} \activ''
\]
where $sm_{new}$ and $a_{new}$ are determined according to the scheme
\begin{align}
  \label{eq:catriskgrad-mit}
  x_{new} =
  \left\{
    \begin{array}{ll}
      x_{t}, & \text{if}\; \partial r /(x_{cur}\to x_{t}) \geq 0 \\
      x_{cur}, & \text{otherwise}
    \end{array}
  \right.\;.
\end{align}
We use the non-strict order $\geq$ because a switch to a desired
target within the same risk level should be allowed.  Then, the
controller activates a \emph{safety function} $sf$ through the
commands
\begin{align*}
  [\acmit[sf]{}]\;& \mathit{ok}_S \land \activ''
  \land \kappa \land \neg\mathit{sf} 
  \longrightarrow \mathit{sf} \mathbin{\&} t'=p
  \\
  [\acmit[sf]{}]\;& \mathit{ok}_S \land \activ''
  \land \kappa \land \mathit{sf}
  \longrightarrow t'=p \quad.
\end{align*}
Reaching phase $\activ''$ constitutes the first set of logical
controller actions.  The performance of these actions is followed by
an interaction with the process.  If this interaction results in the
elimination of $\kappa$, the controller finalises the mitigation stage
with the command
\[
  [\acmit{}]\; \mathit{ok}_S
  \land \activ'' %
  \land \neg\kappa
  \longrightarrow \mitig \;.
\]
The controller subsequently moves into the resumption stage,
continuing with the \emph{deactivation of the safety function} through
\[
  [\acres[sf]{}]\; \mathit{ok}_S \land \mitig
  \land \neg\kappa \land \mathit{sf} \longrightarrow \mathit{sf}^{-1}
\]
and the \emph{resumption from the current to a more progressive safety
  mode} from any risk state $rs\in R(F)$ by
\[
  [\acres[sm]{}]\; \mathit{ok}_S \land \mitig 
  \land \neg\zeta \land \neg\kappa
  \land rs 
  \land sm_{cur} \land \neg\mathit{sf}
  \longrightarrow sm_{new} \mathbin{\&} \mitig' \;.
\]
Beyond its basic enabling condition~($\mathit{ok}_S \land \mitig $),
the controller checks whether both the cause of the critical event and,
particularly, the causal factor subject of mitigation have been
removed~($\neg\zeta\land\neg\kappa$).

Analogously, the \emph{resumption of the current activity to a more
  productive activity} from any risk state $rs\in R(F)$ is
accomplished with
\[
  [\acres[a]{}]\; \mathit{ok}_S \land \mitig'
  \land \neg\zeta \land \neg\kappa
  \land rs 
  \land a_{cur} \land \neg\mathit{sf} 
  \longrightarrow a_{new} \mathbin{\&} \inact \mathbin{\&} t'=p\;.
\]
Given a \emph{resumption option} $(a_t,sm_t)\in S_a\times S_{sm}$
and the set $am(rs)\subseteq F$ of factors active or mitigated in
risk state $rs$, the reaction of the controller follows the scheme
\begin{align}
  \label{eq:catriskgrad-res}
  (a_{new},sm_{new}) = \arg_{(\rfct.a_t,\rfct.sm_t)}\max_{\rfct\in
    am(rs)}
  \big\{\nabla r\mid \nabla r < [a_t,sm_t]^T\big\}\;.
\end{align}
This scheme determines the most permissive yet allowed combination of
activity and safety mode to switch to among all activated or mitigated
factors in $rs$, that is, the combination with the maximum acceptable
risk as seen from $(a_{cur},sm_{cur})$ and $rs$ according to
$\nabla r$.
  
In order for the controller to be able to safely deal with a factor
set $F$, we assume that each of the controller actions is idempotent.
Note how phase indicators~(e.g.\xspace$\mitig$) ensure that the controller is
performing in the right context~(e.g.\xspace if
$rs\in\semRS{\mitig}\subset R(F)$).  The discussion of alleviations is
out of scope of this synthesis approach.

\begin{example}
  Continuing with \Cref{lst:moving}, we specify mitigation
  actions in our \textsc{Yap}\xspace model as shown below for the factor
  {$\rfct[HC]$} with three mitigation and two resumption options.
  \begin{lstlisting}[language={yapscript},
    label=lst:modespec] %
HC ...
  detectedBy (.HCdet)
  mitigatedBy (.HCHguidAud,.HCStOffVis,.HCSrmstIdleVis)
  resumedBy (.HCres,.HCres2) ...;
  
mode HCdet desc "light barrier"
  guard "hSM_PERM & hACT_WELDING & rngDet=close";
mode HCStOffVis desc "emergency stop / cobot+welder turned off / visual notif."
  cf "hST_HOinSGA" // causal factor
  update "(notif'=leaveArea)" // safety function
  target (act=off, safmod=stopped); // target activity/safety mode
mode HCSrmstIdleVis desc "safety-rated mon. stop / cobot+welder idle / visual notif."
  cf "hST_HOinSGA" update "(notif'=leaveArea)" target (act=idle, safmod=srmst);
mode HCStOffAud desc "emergency stop / cobot+welder turned off / audio-vis. notif."
  cf "hST_HOinSGA" update "(notif'=leaveArea)" target (act=off, safmod=stopped);
mode HCHguidAud desc "hand-guided welding / moderate vis. notif."
  cf "hST_HOinSGA" update "(notif'=leaveArea)" target (safmod=hguid);
mode HCres desc "exchange workpiece and start over"
  cf "hST_HOinSGA"
  update "(notif'=ok)" // deactivate safety function
  target (act=exchWrkp, safmod=normal);
mode HCres2 desc "continue welding if workpiece still unfinished"
  cf "hST_HOinSGA" update "(notif'=ok)" target (act=welding, safmod=normal);
  \end{lstlisting}
  The directive \yapkeyw{detectedBy} defines the sensor predicate
  $\zeta$, stating that ``the human operator is in the safeguarded
  area''~(\yapid{hST\_HOinSGA}).  The factor attribute
  \yapkeyw{mitigatedBy} associates $\rfct[HC]$ with three mitigation
  options, and the attribute \yapkeyw{resumedBy} with two resumption
  options.  For example, in the action \yapid{HCStOffVis},
  \begin{inparaenum}[(i)]
  \item \yapkeyw{update} models a safety function, issuing a
    notification to the operator to leave the safeguarded area, and
  \item \yapkeyw{target} switches the manufacturing cell to the
    activity \yapid{off} and to the safety mode
    \yapid{stopped},\footnote{In a design variant discussed in
      \cite{Gleirscher2020-SafetyControllerSynthesis}, we allow
      mitigations to synchronise with the \yapid{robotArm} and
      \yapid{spotWelder} on an event \yapid{stop}.} all triggered by the
    range finder (\yapid{rngDet=close}).
  \end{inparaenum}
  The \yapkeyw{guard} and \yapkeyw{detectedBy} attributes are
  translated into a pair of predicates for $\mathcal{M}$,
  \yapid{RCE\_HC} ($\chi$) describing world states, and \yapid{CE\_HC}
  ($\zeta$) signifying states monitored by the range finder.
  \begin{lstlisting}[language={prism},
  emph={RCE\_HC, CE\_HC},label=lst:monitors]
// HC:monitor "(H)uman (C)lose to active spot welder and cobot working"
formula CE_HC	= (hSM_PERM & hACT_WELDING & rngDet=close) & (HSp=act | HSp=mit1 | HSp=mit2 | HSp=mit | HSp=res) & (HRWp!=act);
formula RCE_HC	= (hSM_PERM & hACT_WELDING & hloc=atWeldSpot) & (HRWp!=act);
  \end{lstlisting}
\end{example}

\subsection{Modelling Performance}
\label{sec:perf-modell}

\begin{wrapfigure}{r}{.4\columnwidth}
  \includegraphics[width=.4\columnwidth]{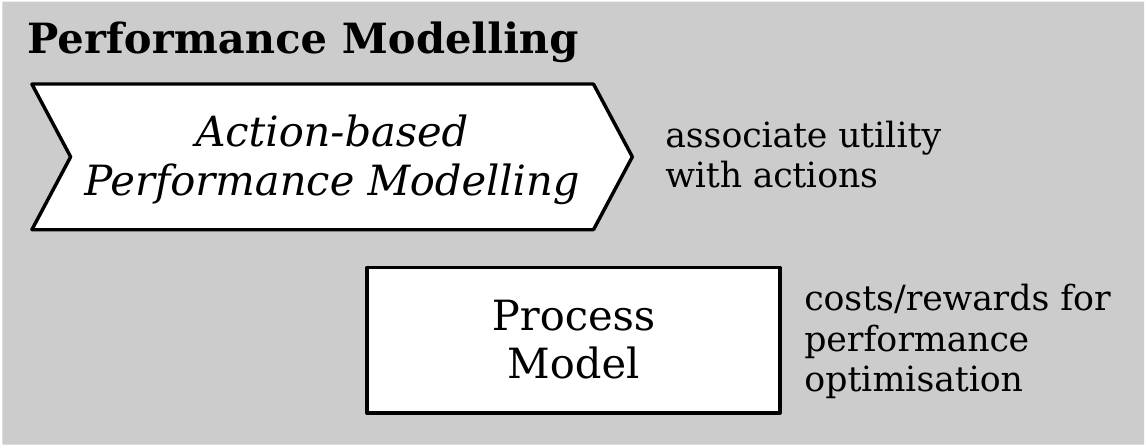}
  \caption{Work step and artifact for performance modelling}
  \label{fig:procedure-perf}
\end{wrapfigure}

In analogy to the rewards for mitigation actions, in this
stage~(\Cref{fig:procedure-perf}), we quantify performance (e.g.\xspace
effort, productivity) for all non-controller actions in the process.
As a result, optimal policy synthesis from $\mathcal{M}$ is based on
several reward structures quantifying the performance of both the
safety controller and the process.

For controller performance, we distinguish mitigation and resumption
options by manually estimated quantities such as \emph{disruption} of
the manufacturing process, \emph{nuisance} of the controller to the
operator, and resources (e.g.\xspace \emph{effort}, \emph{time}) consumed by
the controller.  We formalise these quantities as action rewards
$r^q_{action}$ with $q\in\{\textrm{disr,nuis,eff,time}\}$.  As shown
in \Cref{exa:perf-rewards}, rewards can depend on parameters other
than state variables.

Analogously, concerning process performance, we associate with each
process action a \emph{productivity} measure depending on the safety
mode, using an action reward structure $r_{action}^{prod}$.

\begin{example}
  \label{exa:perf-rewards}
  In a \textsc{Yap}\xspace model, mitigation and resumption options can be associated
  with action rewards:
  \begin{lstlisting}[language={yapscript},emph={nuisance,effort,disruption}]
mode HCStOffVis ...
  disruption=9 nuisance="alarmIntensity1 * 5" 
  effort=5.5; // wear/tear intensive, maintenance effort, energy
mode HCSrmstIdleVis ...
  disruption=7.5 nuisance="alarmIntensity1 * 6" effort=6.5;
mode HCStOffAud ...
  disruption=10 nuisance="alarmIntensity1 * 9" effort=5;
  \end{lstlisting}
  Note how nuisance depends on the parameter \yapid{alarmIntensity1}
  modelling the loudness or brightness of an alarm sound or lamp that
  can be varied in the search for an optimal controller.  Rewards for
  process actions can be specified in a concise manner in a \textsc{Yap}\xspace
  model.
  \begin{lstlisting}[language={yapscript},emph={guard_prod,prod}]
                        guard_prod        prod;
// actor: robotArm
r_moveToTable:          ""                "h";
r_grabLeftWorkpiece: 	""                "m";
r_placeWorkpieceRight:  ""                "h";
r_moveToWelder:         ""                "h";
  \end{lstlisting}
\end{example}

\section{Verified Synthesis of Safety Controllers}
\label{sec:optim-contr-synth}

\begin{wrapfigure}[11]{r}{.5\columnwidth}
  \includegraphics[width=.5\columnwidth]{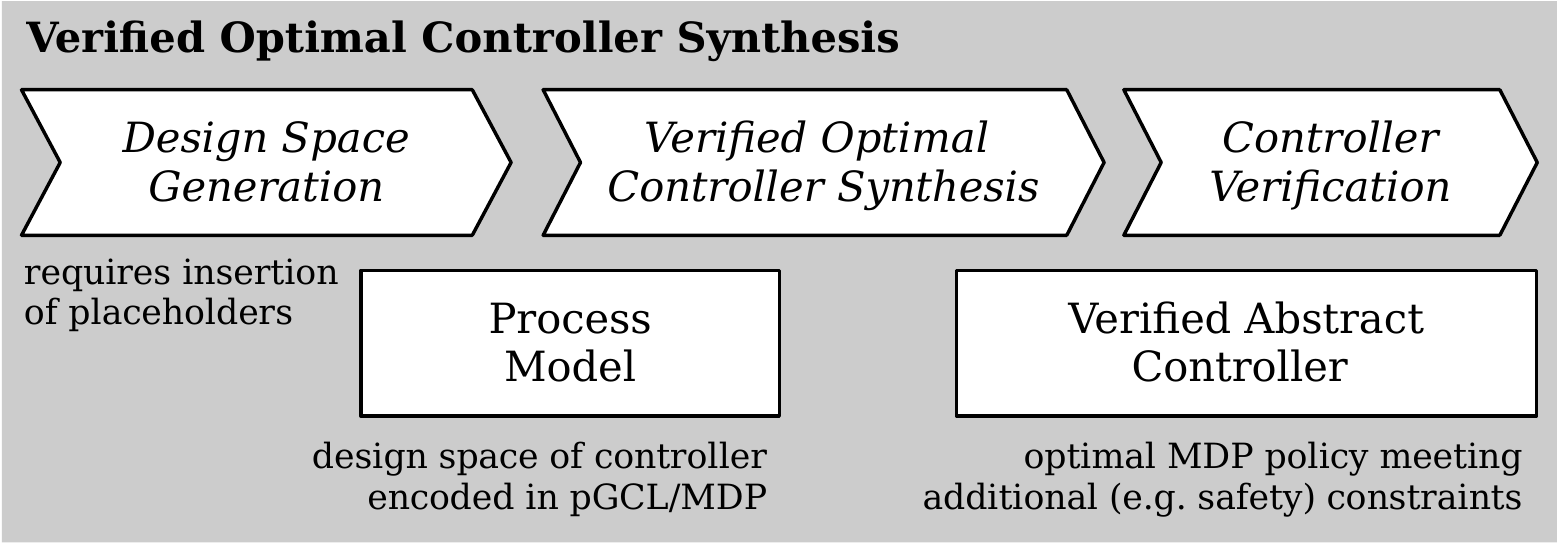}
  \caption{Overview of the work steps and artifacts of the controller synthesis stage}
  \label{fig:procedure-synth}
\end{wrapfigure}

In this stage, we integrate the risk and mitigation models with the
process model, resulting in a risk-informed reward-enhanced stochastic
model that includes the \emph{controller design space} in the process
decision space~(\Cref{fig:procedure-synth}).  The action sets for the
cobot, the operator, and the controller are now combined.
With this integrated model, we perform a constrained policy
  synthesis to select an optimal yet abstract safety controller from
the design space.  We use constraints to encode the \emph{safety
  requirements} and \emph{optimisation queries} to facilitate this
selection.
For this to work, we express safety requirements as \ac{PCTL}
properties and verify them using a stochastic model checker.  We
accomplish controller synthesis in \textbf{two settings}.

\paragraph{The \ac{MDP} Setting.}

We perform optimal policy synthesis from an \ac{MDP} (using \textsc{Prism}\xspace)
where the design space is encoded as non-deterministic choice (e.g.\xspace
among mitigation options).  This approach has already been discussed
in \cite{Gleirscher2020-SafetyControllerSynthesis}.

\paragraph{The \ac{pDTMC} Setting.}

We perform an evolutionary search~(using \textsc{evoChecker}\xspace) of a set of
\acp{DTMC}.  This set defines the design space by fixing the
parameters of a \ac{pDTMC}.  This \ac{pDTMC} can be obtained from the
original \ac{MDP} by replacing non-deterministic choice with random
choice and by introducing the corresponding parameters.  In both
cases, optimal policy synthesis produces a \ac{DTMC} containing an
abstract discrete-event safety controller.  The \ac{pDTMC}-based
approach avoids the split into two verification stages as previously
required in \cite{Gleirscher2020-SafetyControllerSynthesis}.

The synthesis follows a two-staged search through the controller
design space: The first stage focuses on the generation of the guarded
commands according to \Cref{sec:mitigation-modelling}.  The gradient
$\nabla r$~(\Cref{sec:situ-risk-modell}) resolves the calculation of
risk-minimal control updates for these commands.  Reward structures
are generated for risk and performance quantification
(\Cref{sec:perf-modell}) in the second stage.  Then, a stochastic
model checker performs verified policy synthesis for $\mathcal{M}$.
In \citet{Gleirscher2020-SafetyControllerSynthesis}, we use
$\mathcal{M}$ as an \ac{MDP} for policy synthesis with \textsc{Prism}\xspace
\citep{Kwiatkowska2011-PRISM4Verification} and extract a controller
from the resulting \ac{DTMC} representing the policy.

Here, we enhance this approach.  The flexible CSP-style concurrency of
the \ac{pGCL} modules is replaced by a more restrictive
alternation~(\Cref{sec:execution-semantics}) of the process
$\mathcal{P}$ and the safety controller, thereby resembling a
closed-loop control scheme.  The avoidance of interleaving and
synchronous events for modelling real-time phenomena~(e.g.\xspace sensor
faults) results in an \ac{MDP} that does not contain states where
process actors and the controller compete in non-deterministic
choices.  The only choices left are actor-internal choices including
the choice to idle and pass the token.  As a consequence, we obtain
\emph{fairness} for the controller and a simplification of
$\mathcal{M}$.
Additionally, we interpret $\mathcal{M}$ as a \ac{pDTMC} rather than
an \ac{MDP}.  Choice in the safety controller among action options is
explicitly controlled through decision parameters.  

\begin{wrapfigure}{r}{0.5\textwidth}
  \begin{minipage}{0.5\textwidth}
    \begin{algorithm}[H]
      \small
      \caption{Controller design space generation}
      \label{alg:gcgen-msm}
      \begin{algorithmic}[1]
        \Function{GenSafetyModeMitigations}{$F$, $S_{\mathit{sm}})$}
        \ForAll{$\mathit{sm}\in S_{\mathit{sm}}, \rfct\in F, ms\in \mathit{AS}_{\rfct}$}
        \State $m_{\mathit{sm}} \gets$ \Call{CompCmd}{}($\activ\land \mathit{sm}$,
        \Call{GradUpdM}{$\mathit{sm}$,$ms$})
        \Comment{create mit.}
        \State $A_{\mathcal{P}} \gets A_{\mathcal{P}} \cup\{ m_{\mathit{sm}} \}$ 
        \Comment{add mitigation command}
        \EndFor
        \EndFunction
        \State $ $
        \Function{GenActivityMitigations}{$F$, $\mathcal{T}_{S_{a}}$}
        \label{alg:gcgen-ma}
        \ForAll{$a\in \mathcal{T}_{S_{a}}, \rfct\in F, ms\in \mathit{AS}_{\rfct}$}
        \State $m_{a} \gets$ \Call{CompCmd}{}($\activ'\land a$, \Call{GradUpdM}{$a$,$ms$})
        \Comment{create command}
        \State $A_{\mathcal{P}} \gets A_{\mathcal{P}} \cup\{ m_{a} \}$ 
        \Comment{add mitigation command}
        \EndFor
        \EndFunction
        \State \ldots
        \State $ $
        \Function{GenModeResumptions}{$F$, $S_{\mathit{sm}}$}
        \label{alg:gcgen-ra}
        \ForAll{$\mathit{rs}\in R(F), \mathit{sm}\in S_{\mathit{sm}}, \rfct\in am(\mathit{rs}), ms\in
          \mathit{AS}_{\rfct}$}
        \State $(\zeta, \kappa, \mathit{sf}) \gets$ obtained from factor model for $\rfct$
        \State $r_{a} \gets$ \Call{CompCmd}{}($\mitig \land\neg\zeta
        \land\neg\kappa \land \mathit{rs} \land \mathit{sm} \land \neg
        \mathit{sf}$, \Call{GradUpdR}{$\mathit{rs}$,$\rfct$,$ms$})        
        \Comment{create safety mode resumption}
        \State $A_{\mathcal{P}} \gets A_{\mathcal{P}} \cup\{ r_{a} \}$ 
        \Comment{add command}
        \EndFor
        \EndFunction
      \end{algorithmic}
    \end{algorithm}
  \end{minipage}
\end{wrapfigure}
This scheme
results in the removal of non-deterministic choice from the controller
and a randomisation of residual choice in $\mathcal{P}$.  
We further
improve the way how accidents can happen in $\mathcal{M}$.  For 
fine-granular risk and performance quantification, we allow additional
design space parameters~(e.g.\xspace alarm intensity) to be used.  We use \textsc{evoChecker}\xspace~\citep{Gerasimou2018-Synthesisprobabilisticmodels}
for the search of optimal controllers in the space of \acp{DTMC}
defined by these parameters.  The modelling, analysis, and
pre-processing required for the approach in
\cite{Gleirscher2020-SafetyControllerSynthesis} and its enhancement
described here are supported by the \textsc{Yap}\xspace
tool~\citep{Gleirscher2020-YAPToolSupport}.

\subsection{Design Space and Reward Structure Generation}
\label{sec:design-space-gener}

The design space is created by instantiating the command patterns
of the generic factor \ac{LTS} in \Cref{fig:riskfactor}.  These
patterns are used by \Cref{alg:gcgen-msm} and implemented in
\textsc{Yap}\xspace.  The function \textsc{CompCmd} composes guard and update expressions
and integrates these into controller commands compliant with the
specifications in \Cref{sec:mitigation-modelling}.  The functions
\textsc{GradUpdM} and \textsc{GradUpdR} implement
\Cref{eq:catriskgrad-mit} respectively \Cref{eq:catriskgrad-res} for
controlling the updates of safety modes and activities.  $\mathit{AS}_{\rfct}$
refers to the set of mitigation options for factor $\rfct$.  $\mathcal{T}_{S_a}$
refers to the set of activity tuples, that is, combinations of
activities the actors can be involved at a particular point in time.
The other command patterns in \Cref{sec:mitigation-modelling} are
generated analogously.  We omit the corresponding generation functions
here.  The listing in \Cref{exa:design-space-hc} shows a fragment
of the design space.

\begin{example}
  \label{exa:design-space-hc}
  \ac{pGCL} fragment generated for the factor {$\rfct[HC]$}:
\begin{lstlisting}[language={prism},
  emph={RCE\_HC, CE\_HC, HC},label=lst:mitres]
// Endangerments (monitor, extension point: sensor and monitoring errors)
[si_HCact] OK_S & !(HCp=act | HCp=mit1 | HCp=mit2 | HCp=mis) & wact=idle & ract=exchWrkp & CE_HC -> (HCp'=act); ...
// Change of safety modes
[si_HCSrmstIdleVissafmod] OK_S & HCp=act & dpHCmit=HCHCSrmstIdleVis & safmod=normal -> (safmod'=srmst)&(HCp'=mit1); ...
// Frame switches
[s_HChalt] OK_S & HCp=mit1 & dpHCmit=HCHCSrmstIdleVis & wact=welding & ract=exchWrkp -> (wact'=idle)&(HCp'=mit2); ...
// Execution of safety functions
[si_HCSrmstIdleVisfun] OK_S & HCp=mit2 & dpHCmit=HCHCSrmstIdleVis & hST_HOinSGA & !(notif=leaveArea) 
-> (notif'=leaveArea)&(token'=mod(token+1,ag))&(turn'=token+1); ...
// For entering the mitigated phase
[si_HCmit] OK_S & HCp=mit2 & dpHCmit=HCHCSrmstIdleVis & !(hST_HOinSGA) -> (HCp'=mit); ...
// Switching off safety functions
[si_HCres2fun] OK_S & HCp=mit & dpHCres=HCHCres2 & !(hST_HOinSGA) & (notif=leaveArea | notif=leaveArea | notif=leaveArea) -> (notif'=ok); ...
// Meta-policy for resuming to a less restrictive safety mode
[si_HCres2safmod] OK_S & HCp=mit & dpHCres=HCHCres2 & !CE_HC & !hST_HOinSGA & (HSp=mit|HSp=res) & (HRWp=mit|HRWp=res) & safmod=normal & notif=ok -> (safmod'=pflim)&(HCp'=res); ...
// Resuming actor's activities
[s_HCresume2] OK_S & HCp=res & dpHCres=HCHCres2 & !CE_HC & !(hST_HOinSGA) & (HSp=mit|HSp=res) & (HRWp=mit|HRWp=res) & wact=welding & ract=exchWrkp -> (wact'=welding) & (ract'=welding) & (token'=mod(token+1,ag)) & (turn'=token+1) & (HCp'=inact); ...
\end{lstlisting}
\end{example}

\Cref{lst:rewards} shows a fragment of the reward structures generated
from the \textsc{Yap}\xspace model described in the
\Cref{sec:activity-based-risk,sec:action-based-risk,sec:perf-modell}.

\begin{example}
  \label{exa:prism-rewards}
  The listing below shows two reward structure fragments, one for risk
  from an active {$\rfct[HC]$} and one for nuisance.
\vspace{-2em}
\begin{lstlisting}[language={prism}, emph={RCE_HC,CE_HC}, label=lst:rewards,multicols=2,aboveskip=-2em]
// Risk of HC-mishap when performing nominal action ...
rewards "risk_HC"
[rw_leaveWelder] !CYCLEEND & (RCE_HC | CE_HC) & safmod=pflim	: 0.6 * 5 * 9.0;
[h_exitPlant] !CYCLEEND & (RCE_HC | CE_HC) & safmod=hguid	: 0.4 * 2 * 9.0;
[r_moveToTable] !CYCLEEND & (RCE_HC | CE_HC) & safmod=hguid	: 0.4 * 5 * 9.0;
...
endrewards

...
// Nuisance (e.g. to the human operator; per mitigation option)
rewards "nuisance"
[si_HCStOffVisfun] REWGUARD_HC	: alarmIntensity1 * 5 / 1;
[si_HRWmit2fun] REWGUARD_HRW	: alarmIntensity2 * 4 / 1;
[si_pflimfun] REWGUARD_HS	: alarmIntensity1 * 8 / 1;
...
endrewards
\end{lstlisting}%
\end{example}

\subsection{Verified Optimal Synthesis}
\label{sec:contr-verif}

\begin{wrapfigure}[33]{r}{.5\columnwidth}
  \begin{tikzpicture}
    \draw[draw] (-4.3,-4.7) rectangle (-.2,4.7);
    \draw[draw] (0,-4.7) rectangle (4.1,4.7);
    \node at (0,0) {\includegraphics[width=.5\columnwidth]{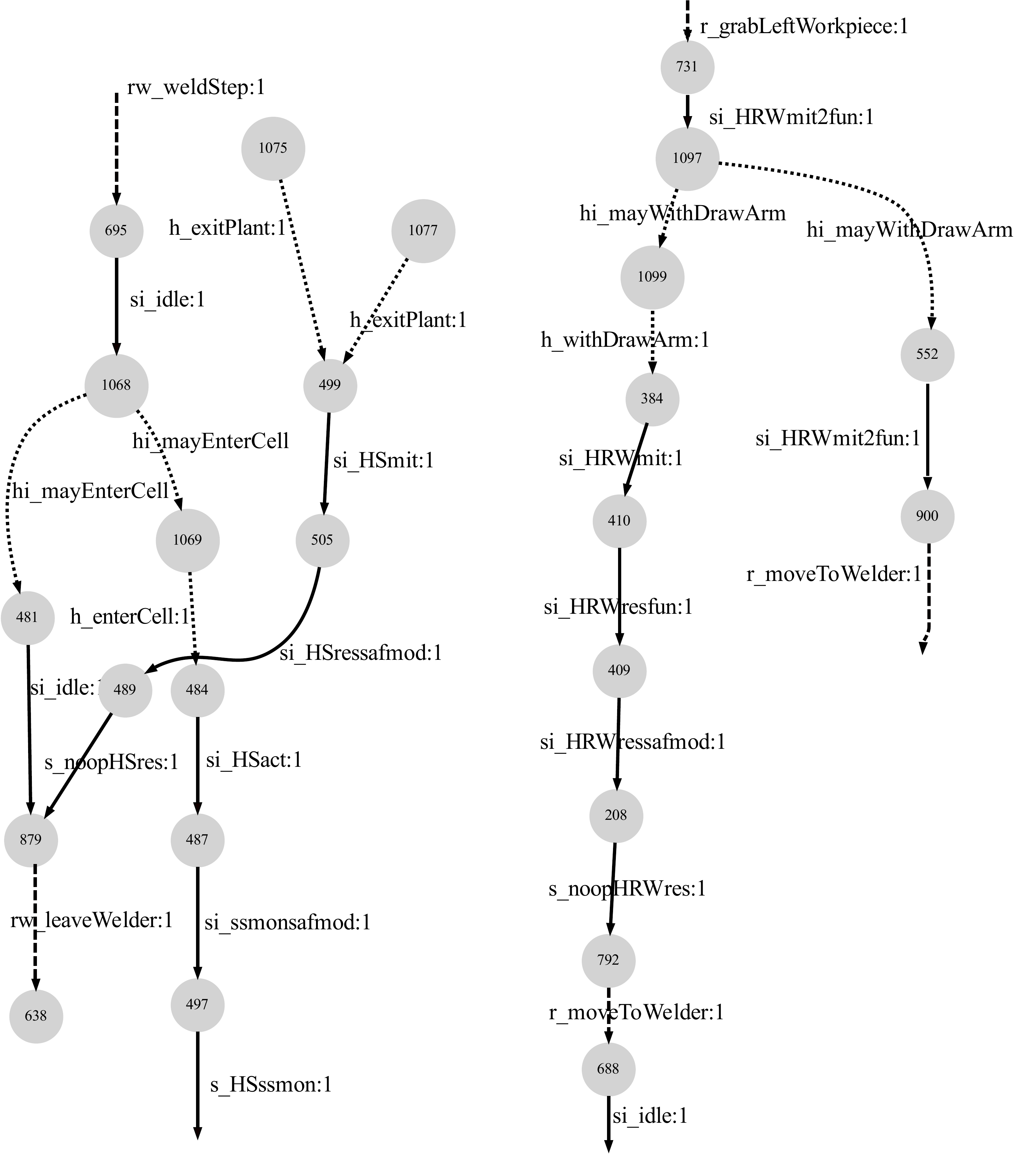}};
    \node[draw,fill=white,anchor=south east] at (4.1,-4.7)
    {\includegraphics[width=2cm]{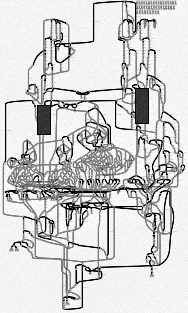}};
    \draw[very thick,-latex] (3.5,-2.2) edge ++(-.5,1.5) (2.4,-2.5) -> ++(-3,2);
  \end{tikzpicture}
  \caption{Visualisation of an optimal
    policy $\pi^{\star}$ (lower right box) refining the risk graph for
    our case study in \Cref{fig:riskgraph}.  The controller decisions
    (solid arcs), the spot welder and cobot actions (dashed arcs), 
    probabilistic choices of the operator (dotted arcs). The two zoomed fragments
    highlight that mitigation of $\rfct[HS]$ or $\rfct[HRW]$ can deal
    with a variety of environments, for example, adversarial human
    decisions.}
  \label{exa:process-policy} 
\end{wrapfigure}

The \ac{pGCL} action system consists of the process (i.e.,\xspace cobot,
welding machine, and operator) and the safety controller generated
according to \Cref{sec:design-space-gener}.  The policy space
$\Pi_{\mathcal{M}}$ includes the controller design space.  This action
system is expanded into an \ac{MDP} by a probabilistic model checker
such as \textsc{Prism}\xspace or it is used as a \ac{pDTMC} by \textsc{evoChecker}\xspace relying on
\ac{DTMC} model checking.  Choice in $\Pi_{\mathcal{M}}$ stems from
commands~(e.g.\xspace mitigations, resumptions) simultaneously enabled in a
state $s\in S$, yielding multiple policies for $s$ and from commands
enabled in multiple states, giving rise to a policy for each ordering
in which these commands can be chosen.

Controller solutions selected from the design space are subjected to
two kinds of requirements.  The first are optimisation
  objectives, such as minimal energy consumption.  The second are
probabilistic and reward-based constraints for safety~(i.e.,\xspace
unlikely reachability of bad states, e.g.\xspace accidents below probability
threshold; accumulated risk below reward threshold) and response~(e.g.\xspace
timely controller response exceeds probability threshold).  Both kinds
of requirements are expressed in reward-enhanced, quantitative
\ac{PCTL}~(\Cref{sec:probmc}).

\paragraph{Optimisation Objectives.}

An optimal policy $\pi^{\star}\in\Pi_{\mathcal{M}}$, including the
controller decisions, can be selected based on, for example, minimum
nuisance or maximum productivity.  For that, $\mathcal{M}$ includes
action rewards to quantify controller and
process performance~(\Cref{sec:perf-modell}), risk reduction
\emph{pot}ential (in the \textbf{MDP setting}), and risk based on factors, modes, and activities as explained in \Cref{sec:risk-modelling}.

\paragraph{Constraints.}
\label{sec:verif-mdp}

Constraints are of the form
$\pi^{\star} \models \phi_{\mathit{wf}} \land \phi_c$.
$\phi_{\mathit{wf}}$ captures
well-formedness,\footnote{Well-formedness refers to the class of
  properties (see, e.g.\xspace\Cref{tab:properties}) to be checked to
  establish basic model validity prior to more interesting correctness
  properties related to the application under consideration.}
including properties for the verification of, for example, hazard
occurrence and freedom from pre-$\mathit{final}$ deadlocks, and
properties for the falsification of, for example, that final states
must not be initial states.  Hazard occurrence fosters our focus on
adversarial environments.  $\phi_{\mathit{wf}}$ can help one to
simplify model debugging, decrease model size, remove deadlocking
states, and reduce vacuity of verification results.

$\phi_c$ specifies safety-carrying correctness and can include
progress, safety, liveness, and reliability properties. For example,
we want to verify \emph{reach-avoid} properties of type
$\mathop{\textbf{A}}\mathop{\textbf{G}}\mathop{\textbf{F}}\gamma \land
\mathop{\textbf{A}}\mathop{\textbf{G}}\neg\chi$; or that the probability of failure
on demand of the controller or the probability of a mishap from any
hazard is below a threshold.
With
$\mathop{\textbf{A}}\mathop{\textbf{G}}(\activ\rightarrow\mathop{\textbf{P}}_{>p}[\mathop{\textbf{F}}\mitig])
\land
\mathop{\textbf{A}}\mathop{\textbf{G}}(\mitig\rightarrow\mathop{\textbf{P}}_{>p}[\mathop{\textbf{F}}\inact])$,
we constrain the search for solutions in the design space to
controllers whose mitigation paths from critical events are
complete with at least probability $p$.
In the \textbf{\ac{MDP} setting}~(as explained in
\Cref{sec:optim-contr-synth}), our model allows the evaluation of
freedom from accidents in $\mathcal{M}$ with
\begin{align}
  \label{eq:accfreedom}
  {\mathop{\textbf{P}}}_{\neg A}\equiv
  f_{s\in\Xi}\; {\mathop{\textbf{P}}}^s_{\min=?}
  [\neg \mishp[F] \;\mathop{\textbf{W}}\; \inact[F]]
\end{align}
where $f\in\{\min, \mathrm{mean}, \max\}$.  For the non-accident
$F$-unsafe region~$\Xi$ (\Cref{sec:yap}), \Cref{eq:accfreedom}
requires the controller to minimise the probability of mishaps until
the $F$-safe region~(i.e.,\xspace $S\setminus(\Xi\cup\mishp[F])$) is reached.
$\mathop{\textbf{P}}_{\neg A}$ aggregates $\min$, the arithmetic
$\mathrm{mean}\;\mu$, and $\max$ probabilities over $\Xi$.  In the
\textbf{\ac{pDTMC} setting}, we use the plain quantification operator
$\mathop{\textbf{P}}$.  \Cref{tab:properties} contains further examples of
properties in $\phi_{\mathit{wf}}$ and $\phi_c$ to be verified or
falsified of $\mathcal{M}$.

\paragraph{Verified Synthesis.}

Based on the action rewards from the
\Cref{sec:activity-based-risk,sec:action-based-risk,sec:situ-risk-modell},
the model checker investigates all choice resolutions and parameter
valuations for $\mathcal{M}$ that fulfil well-formedness and further
\ac{PCTL} constraints.  The checker then identifies policies that
fulfil the given constraints and are (Pareto\nobreakdash-)optimal with
respect to the optimisation objectives.  The existence of an optimal
strategy depends on the existence of a strategy in $\Pi_{\mathcal{M}}$
that fulfils the \ac{PCTL} constraints.  Note that, by
\Cref{def:policy}, all policies considered for $\mathcal{M}$ are of
the same size but may vary in their distribution of choice among the
involved actors.
The overall result of this step is a verified and optimal abstract
controller extracted from the selected policy $\pi^\star$.  An example
of such a policy is visualised in \Cref{exa:process-policy}.

\begin{table}[t]
  \caption{Objectives to be queried over $\Pi_{\mathcal{M}}$ and
    properties to be checked of every $\pi\in\Pi_{\mathcal{M}}$}
  \label{tab:properties}
  \begin{tabularx}{\columnwidth}{%
    >{\hsize=.33\hsize}X
    >{\hsize=.67\hsize}X}
    \toprule
    \textbf{Property}$^\dagger$
    & \textbf{Description}
    \\\specialrule{\lightrulewidth}{\aboverulesep}{0em}
    \multicolumn{2}{l}{\cellcolor{lightgray}{\large$\phantom{\mid}$\hspace{-.25em}}\emph{Optimisation objectives}}
    \\\specialrule{\lightrulewidth}{0em}{\belowrulesep}
    $\mathop{\textbf{R}}^{\mathrm{eff}}_{\max=?} [ \mathop{\textbf{C}} ]$ 
    & Assuming an adversarial environment, select
      $\pi$ that maximally utilises the safety controller.
    \\
    $\mathop{\textbf{R}}^{\mathrm{nuis}}_{\min=?} [ \mathop{\textbf{C}}^{<T} ]$ 
    & Select $\pi$ that minimises nuisance up to time $T$.
    \\\specialrule{\lightrulewidth}{\aboverulesep}{0em}
    \multicolumn{2}{l}{\cellcolor{lightgray}{\large$\phantom{\mid}$\hspace{-.25em}}\emph{Objectives with reward-based constraints}}
    \\\specialrule{\lightrulewidth}{0em}{\belowrulesep}
    $\mathop{\textbf{R}}^{\mathrm{prod}}_{\max=?} [ \mathop{\textbf{C}} ] \land \mathop{\textbf{R}}^{\mathrm{sev}}_{\leq s}
    [ \mathop{\textbf{C}} ] \land \mathop{\textbf{R}}^{\mathrm{risk}}_{\leq r} [ \mathop{\textbf{C}} ]$
    & Select controller that maximises productivity 
      constrained by risk level $r$ and expected severity $s$.
    \\
    $\mathop{\textbf{R}}^{\mathrm{prod}}_{\max=?} [ \mathop{\textbf{C}} ] \land \mathop{\textbf{R}}^{\mathrm{sev}}_{\leq
      s} [ \mathop{\textbf{C}} ]$ 
    & Select controller that maximises productivity constrained by
      exposure $p$ to severe injuries. 
    \\\specialrule{\lightrulewidth}{\aboverulesep}{0em}
    \multicolumn{2}{l}{\cellcolor{lightgray}{\large$\phantom{\mid}$\hspace{-.25em}}\emph{Well-formedness constraints
    in $\phi_{\mathit{wf}}$}}
    \\\specialrule{\lightrulewidth}{0em}{\belowrulesep}
    $\mathop{\textbf{E}}\mathop{\textbf{F}} \mathit{final}$ 
    & $\mathcal{P}$ can finish the production cycle.
    \\
    $\mathop{\textbf{E}}\mathop{\textbf{F}}(\activ\land\neg \mathit{final})$
    & $\rfct$ can occur during a production cycle.
    \\
    $\mathop{\textbf{E}}\mathop{\textbf{F}}(\mathit{deadlock} \land\neg \mathit{final})$
    & $\mathcal{P}$ can deadlock early. (f)
    \\
    $\mathop{\textbf{A}}\mathop{\textbf{F}}\activ$
    & $\rfct$ is inevitable. (f)
    \\
    $\forall s\in S\colon \neg\mathit{final} \lor \neg\mathit{init}$
    & Some initial states are also final states. (f)
    \\\specialrule{\lightrulewidth}{\aboverulesep}{0em}
    \multicolumn{2}{l}{\cellcolor{lightgray}{\large$\phantom{\mid}$\hspace{-.25em}}\emph{Correctness constraints
       in $\phi_c$}}
    \\\specialrule{\lightrulewidth}{0em}{\belowrulesep}
    $\mathop{\textbf{A}}\mathop{\textbf{G}} (\zeta \rightarrow \mathop{\textbf{A}}\mathop{\textbf{F}}^{<t} \activ)$
    & The controller detects
      $\chi$ for $\rfct$ within $t$ steps.
    \\
    $\mathop{\textbf{A}}\mathop{\textbf{G}} (\zeta\land\inact \rightarrow
    \mathop{\textbf{A}}(\zeta\mathop{\textbf{U}}\activ))$%
    \label{eq:ctl-detected}%
    & The controller timely responds to
      the belief of $\chi$.$^\ddagger$%
    \\
    $\mathop{\textbf{A}}\mathop{\textbf{G}}(\activ\rightarrow\mathop{\textbf{P}}_{>p}[\mathop{\textbf{F}}\mitig])
    \land
    \mathop{\textbf{A}}\mathop{\textbf{G}}(\mitig\rightarrow\mathop{\textbf{P}}_{>p}[\mathop{\textbf{F}}\inact])$%
    \label{eq:ctl-handled}%
    & The controller lively handles hazard $\rfct$.
    \\
    $\mathop{\textbf{A}}\mathop{\textbf{G}}(\activ\rightarrow\mathop{\textbf{P}}_{>p}[\mathop{\textbf{F}} \mathit{final}])$
    & The controller resumes $\mathcal{P}$ so it can finish its
    cycle after $\rfct$ has occurred.
    \\
    $\mathop{\textbf{P}}_{>p}[\mathop{\textbf{G}} \neg\mishp[F]]$
    & Mishap freedom is more likely than $p$.
    \\
    $\mathop{\textbf{S}}_{< p} \mishp[F]$
    & The steady-state probability of any
    $\mishp$ is below $p$.
    \\\bottomrule
    
    \multicolumn{2}{p{.95\columnwidth}}{
    $^{\dagger}$
    $\mathit{deadlock}$: state with no commands enabled,
    $\mathit{final}$: end of manufacturing cycle,
    $\mathit{init}$: initial state of a manufacturing cycle,
    $\mishp[F]$: mishap state,
    $p$: probability bound,
    f: to be falsified,
    $\mathrm{prod}$: productivity,
    $\mathrm{eff}$: controller effectiveness,
    $\mathrm{sev}$: severity,
    $\mathrm{risk}$: risk level.
    $^{\ddagger}$We adopt the universality pattern after
    \citet{Dwyer1999-Patternspropertyspecifications} 
    using $\mathop{\textbf{U}}$ instead of $\mathbf{\mathsf{W}}$ because of the
    required response; note that we 
    have to use the sensor predicate $\zeta$ rather than the ground
    truth predicate $\chi$.}
  \end{tabularx}
\end{table}

\section{Controller Deployment and Validation}
\label{sec:ctr-depl}

\begin{wrapfigure}[11]{r}{.5\columnwidth}
  \includegraphics[width=.5\columnwidth]{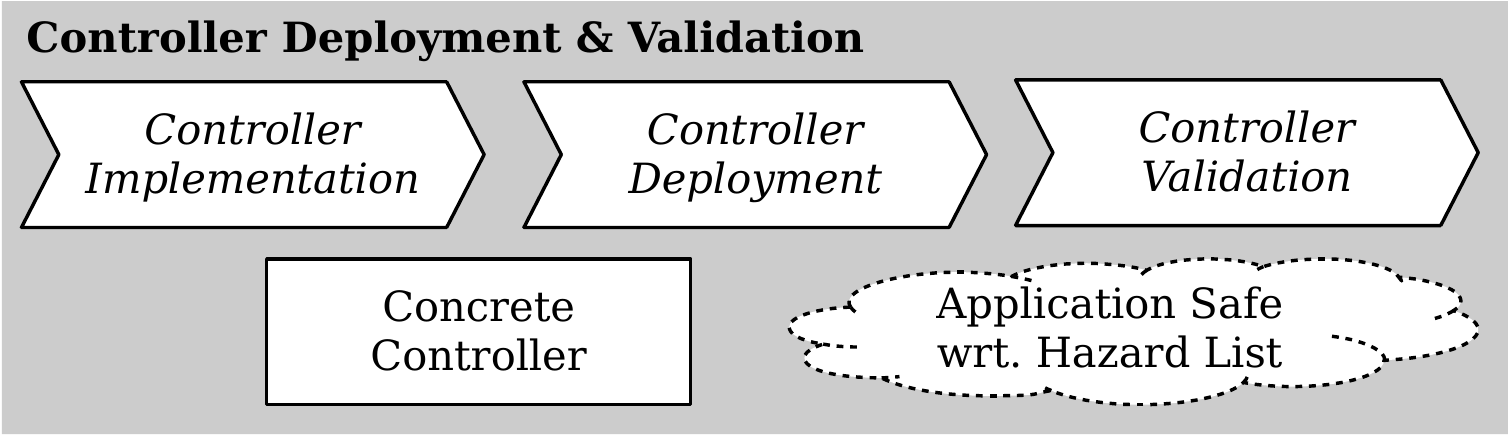}
  \caption{Overview of the work steps and artifacts of the controller
    deployment and validation stage}
  \label{fig:procedure-depl}
\end{wrapfigure}

Given the execution semantics of a target platform~(e.g.\xspace the
\acl{CDT}), the selected controller can now be translated into a
concrete executable form~(\Cref{fig:procedure-depl}) to be deployed,
validated, demonstrated, and eventually used on this platform.

\subsection{Controller Implementation}
\label{sec:contr-impl}

The abstract controller, $c$, is part of the calculated policy
$\pi^{\star}$, a \ac{DTMC} with state space $S_c\subseteq S$.  $S_c$
is a result of combining the risk state space, generated by \textsc{Yap}\xspace from
the factor set $F$, with the process state space.  According to
\Cref{def:policy}, the transition relation of $\pi^{\star}$ is a list
of $(\mathit{state, action, probability, state})$-tuples.  The
controller as a deterministic part of $\pi^{\star}$ is a set of
transitions $\delta_c\subseteq S_c\times S_c$ that, at the concrete
level, again comprises guarded commands of the form
\begin{align*}
  [\overbrace{\mbox{controller action}}^{\text{event}}]\;
  & \overbrace{\mbox{process state} \land \mbox{risk state}}^{\text{guard}} 
    \rightarrow
    \overbrace{\text{mode \& activity switch, safety function}}^{\text{update}}\;.
\end{align*}

\begin{wrapfigure}{r}{.4\columnwidth}
  \includegraphics[width=.4\columnwidth]{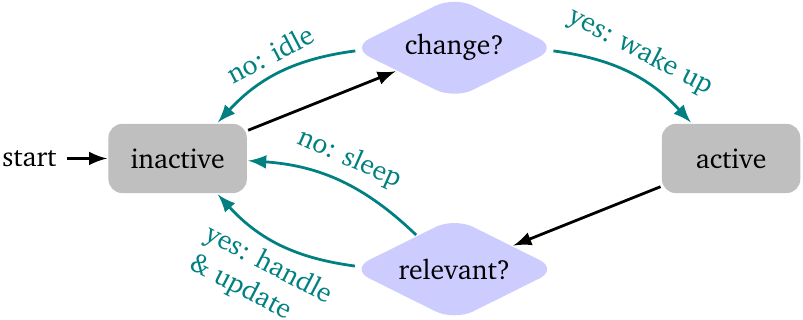}
  \caption{Execution of the controller in the \acl{CDT}}
  \label{fig:ctr-impl}
\end{wrapfigure}

We use the transition matrix obtained from the model checker
(e.g.\xspace\textsc{Prism}\xspace) to translate controller actions in $\pi^{\star}$ into
concrete actions.  Following the four-relation structure of controller
models in \cite{Parnas1995-FunctionalDocumentationComputer}, this step
involves
\begin{inparaenum}[(i)]
\item the translation of the abstract states into guard conditions
  based on a mapping of concrete process states into abstract states,
  and
\item the translation of the abstract updates into low-level
  procedures generating control inputs to the process.
\end{inparaenum}
\Cref{fig:ctr-impl} shows the activation scheme of the concrete
controller when deployed on an execution platform.
\Cref{alg:concrete-controller} describes the corresponding
discrete-event \textsc{SafetyController}.  According to
\Cref{fig:int-sem}, through \Cref{alg:wake-up}, the controller is
aware of each atomic update of any of the monitored variables.

\begin{wrapfigure}[12]{r}{0.5\textwidth}
  \begin{minipage}{0.5\textwidth}
    \begin{algorithm}[H]
      \small
      \caption{Execution of the safety controller} 
      \label{alg:concrete-controller}
      \begin{algorithmic}[1]
        \Procedure{SafetyController}{in \textsf{State} $m$, out
          \textsf{State} $c$}
        \State $(p,r,r') \gets$ \Call{init}{}
        \Comment{initialise controller and risk state}
        \While{true}
        \If{$(p,r) \neq (m,r')$} %
        \label{alg:wake-up}
        \Comment{wake up on event/change} 
        \State $(p,r) \gets (m,r')$
        \Comment{capture monitored process state}
        \If{$p\in S_c$} %
        \label{alg:check-event}
        \Comment{is the event relevant?}
        \State $c \gets$ \Call{HandleEvent}{$p, r$}
        \label{alg:handle-event}
        \Comment{issue control command}
        \State $r' \gets$ \Call{UpdateRiskState}{$p, r$}
        \Comment{stay active if changed}
        \EndIf
        \EndIf
        \EndWhile
        \EndProcedure
      \end{algorithmic}
    \end{algorithm}
  \end{minipage}
\end{wrapfigure}

\begin{example}
  In our case study, the check of whether $p\in S_c$ in
  \Cref{alg:check-event} of \Cref{alg:concrete-controller} is
  implemented as a switch statement iterating over all relevant
  combinations of events known from $\mathcal{M}$.  The function
  \textsc{HandleEvent} is responsible for issuing control inputs, such
  as switching into a power \& force limitation mode and notifying the
  operator to leave the work cell if $\rfct[HC]$ is activated.  The
  function \textsc{UpdateRiskState} is responsible for managing and
  remembering the risk state internal to the controller~(e.g.\xspace state in
  the mitigation of $\rfct[HC]$).  In the supplemental
  material\footnote{See
    \url{https://github.com/douthwja01/CSI-artifacts/}.}
  for this work, we provide modelling and code examples\footnote{See
    the \texttt{hrc2} example of the \textsc{Yap}\xspace package 
    (\url{https://github.com/ytzemih/yap}).} and a video\footnote{See
    \url{https://youtu.be/cm-XkZ_aitQ}.}  of the controller in action.
\end{example}

\paragraph{Expected Overhead.}
\label{sec:expoverhead}

The detection and handling overhead is the time elapsed in every
cycle of the \textbf{while} loop in \Cref{alg:concrete-controller}.
Let $d\colon \alpha_{\mathcal{P}}\to\mathbb{R}$ be the processing time
required for an action, for example, the calculation of the
detection of $\rfct[HC]$ in $\acend{\rfct[HC]}$.  If implemented as
part of a sequential cell controller, \Cref{alg:check-event} requires
a time slot of length $\Sigma_{\rfct\in F} d(\acend{\rfct})$ in each
control cycle.  If \Cref{alg:check-event} is monitored simultaneously
in dedicated safety controller hardware, the slowest detection rate
for $F$ is $1 / \max_{\rfct\in F} d(\acend{\rfct})$.  The overhead for
handling $\rfct$ in \Cref{alg:handle-event} can be estimated from
\Cref{fig:riskfactor} and may range from
$  d_{\min}^{\rfct} = d(\acmit[sm]{\rfct}) + d(\acmit[a]{\rfct})
  + d(\acmit{\rfct})
  + d(\acres[sm]{\rfct}) + d(\acres[a]{\rfct}) $
to $d_{\max}^{\rfct} = d_{\min}^{\rfct}
+ x \cdot d(\acmit[sf]{\rfct}) + d(\acres[sf]{\rfct})$
with a repetition factor $x\in\mathbb{N}$.
The overhead of the implementation in \Cref{alg:concrete-controller}
can be obtained by recording timed event traces from an execution
platform (e.g.\xspace the \ac{CDT}).  In order for the controller to interact
with such a platform, the \textsc{Yap}\xspace model is extended by an interface
specification in addition to the model fragments discussed in the
previous sections.  This interface is discussed in more detail in the
following section.

\subsection{Controller Deployment}
\label{subsec:controller-integration}

\begin{wrapfigure}[22]{r}{.3\columnwidth}
  \includegraphics[height=.3\columnwidth,angle=90]{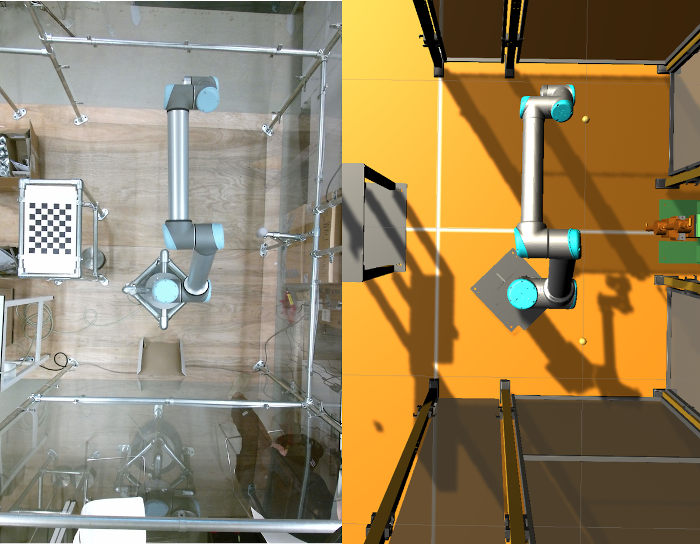}
  \caption{The \ac{CDT} applied to our case study. Requests issued to the
    robot within the \ac{CDT} are enacted by the physical twin in the
    real-world system. The robot, LIDAR, and light barrier provide
    feedback to the environment.}
  \label{fig:digital-twin-top-down}
\end{wrapfigure}

\Cref{subsec:what-is-a-digital-twin} introduces the notion of actors within a collaborative manufacturing setting involving a cobot and an operator (see \Cref{fig:digital-twin-top-down}).  We are able to reconstruct this setting in the form of a digital twin using our \acl{CDT}.  The \ac{CDT} is a toolchain developed in C\# and visualised in Unity3D that provides the kinematic, communication, and data infrastructure necessary to deploy digital twins on real world systems~\footnote{See \url{https://github.com/douthwja01/CSI-artifacts/JSS/}.}.  Using the \ac{CDT} APIs for the MATLAB\textsuperscript{\textregistered} robotics toolbox and the Robotic
Operating System framework\footnote{See \url{https://www.ros.org}.}, the safety controller's actions are exchanged with the physical platform in real-time where a response is demonstrated. 

We represent the scenario with as an aggregation of actions and
decisions made by each actor.  Actors may then be identified as
\begin{inparaenum}[(i)]
\item \emph{digital twins}---actors with a distinct collection of
  models, state-machines and behaviours that emulate the capabilities
  of the physical system---and 
\item \textit{abstract} actors without physical embodiment that may
  provide a service, communicate with or control other actors. 
\end{inparaenum}
The term \emph{environment} will be used to describe this complete set
of actors.

\paragraph{Process Representation.}
\label{subsubsec:process-representation}

Let $\mathcal{A}$ be the set of all actors.  Examining the process
from a network perspective allows us to model an actor
$n\in\mathcal{A}$ as a communication node, similar to the robotic
operating system framework.
Following terminology in \cite{Broy2010-LogicalBasisComponent}, the
actor $n$ is able to communicate with other nodes through an interface
$I_n = (S_n^j,P_n^k)$.  Here, $S_n = [S_n^1,S_n^2,...,S_n^j]$ and
$P_n = [P_n^1,P_n^2,...,P_n^k]$ denote the subscription (or input) and
publication (or output) channels of actor $n$ respectively.  This
decentralised structure allows us to represent a \emph{process
  controller} $p\in\mathcal{A}$ as abstract actor with known feedback
and command channels $S_{p}$ and $P_{p}$ respectively. The process
controller is modelled as a state machine that responds to feedback
from actor $n$ and issues a requested action on $P_{p}$.

Each sensor present in the process is similarly introduced as a
digital twin actor $s\in\mathcal{A}$. Here, data originating from the
sensing capabilities of $s$ are broadcast to assigned channels $P_s$
in order to inform the network of changes to the physical environment.
Manipulators and machinery are modelled as digital twins of the
physical equipment, with behaviours informed by the actuation
constraints of the physical system.  In response to commands issued by
$p$, the robot digital twin is able to interact with the work piece,
operator or tendered machine.  This communication is then forwarded
and expressed by the physical twin as seen in
\Cref{fig:digital-twin-top-down}.

\paragraph{Safety Controller.}
\label{subsubsec:safety-controller-integration}

\begin{wrapfigure}[33]{r}{.5\columnwidth}
  \includegraphics[width=.5\columnwidth]{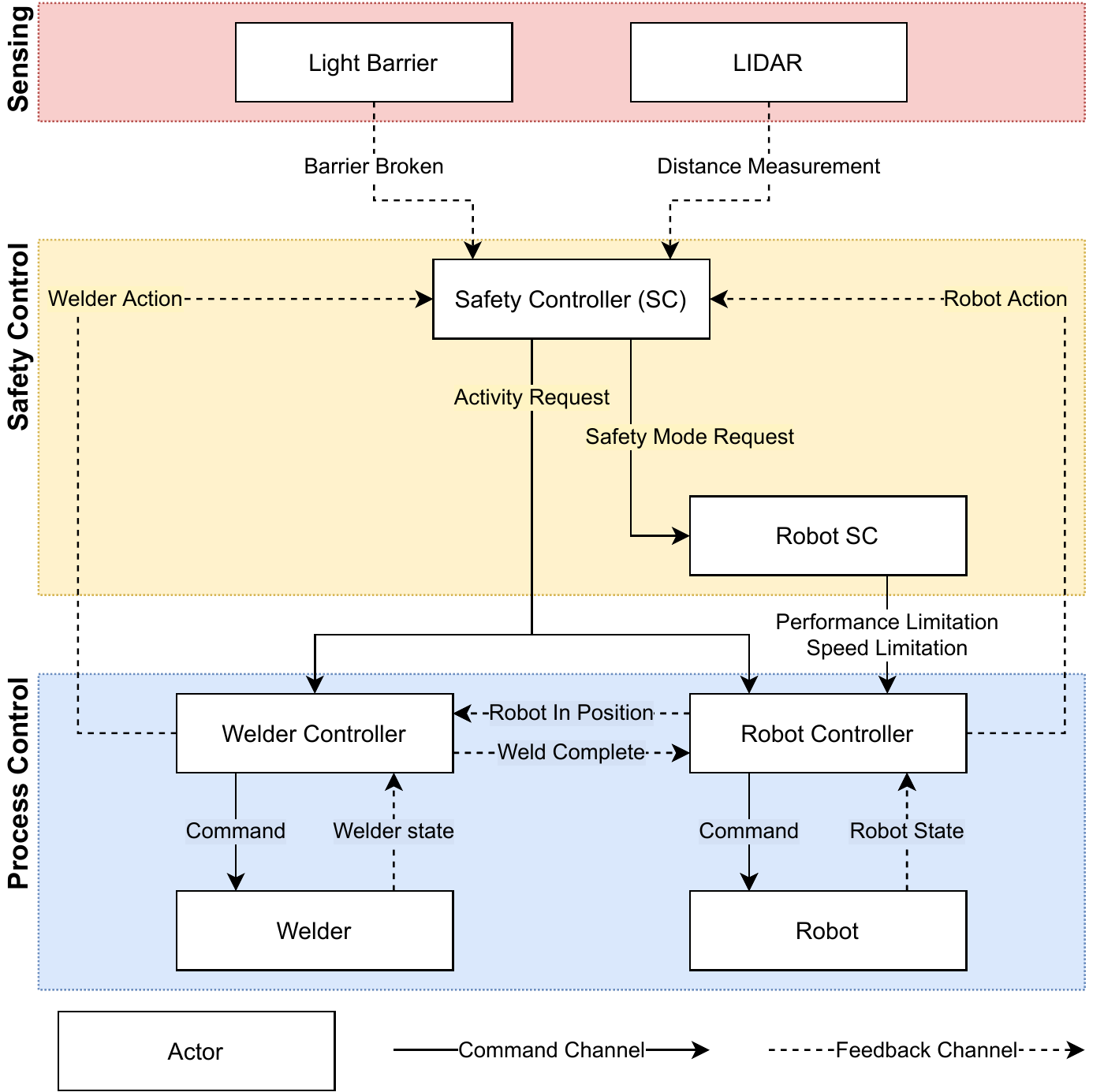}
  \caption{Communication structure in the \ac{CDT}.  In accordance
    with our case study, a safety mode is requested by the safety controller (middle block) to
    apply to the welder and robot digital twins (bottom block). In
    response, the nominal process for both systems is overridden. For
    example, in response to a detected hazard the controller may issue the
    safety action ``safety stop'', preventing the process continuing.}
  \label{fig:mapping-communications}
\end{wrapfigure}

A safety controller $c\in\mathcal{A}$ is introduced to the \ac{CDT} as
a singleton node declared as an abstract actor.  $c$
communicates on fixed channels
$P_{c} = [P_{c}^1,P_{c}^2,...,P_{c}^l]$ with a family of process actors
$\{p_i\}_{i\in 1..k}\subset\mathcal{A}$.  The nominal procedure of $p$
is governed by a local hierarchical state machine responding to
process updates on $S_p$.  To allow the safety controller to
intervene in this nominal procedure, the interpreter behaviour $S_c$
is implemented as a parent state machine as seen in
\Cref{fig:mapping-communications}.  This secondary state machine
allows then a safety controller to enact changes to $n$'s safety
mode(s) in response to requests made over channels in $P_c$.

\begin{example}
  As part of the case study presented in \Cref{sec:running-example},
  an environment has been developed to evaluate the synthesised safety
  controller.  This environment, shown in
  \Cref{fig:digital-twin-top-down}, presents us with a distributed
  system composed of multiple digital twins, with each physical actor
  (e.g.\xspace robot, spot welder, operator) and component (e.g.\xspace sensors) in
  the real-world process assigned its own individual twin.

  The process is a work-piece exchange and welding task, overseen by a
  process controller that issues and responds to tasks assigned to
  each actor. The safety controller observes feedback from the process
  controller, a workbench light barrier and a LIDAR positioned within
  the cell. \Cref{fig:mapping-communications} then describes the
  communication of safety mode and activity change requests to all
  actors within the environment on occurrence of a critical event,
  following the event handling logic in \Cref{fig:ctr-impl}.
\end{example}

\subsection{Controller Validation by Testing}
\label{sec:contr-valid-test}

The validation of a controller implementation $c$ in the \ac{CDT} can
be done by use-case-based testing of scenarios generated from the
process model $\mathcal{M}$.
Given a use case $U$, a misuse case $\mathit{MU}$, an initial state
$s_0\in S$, a set $\mathcal{T}_{\mathcal{V}}(U)$ of \emph{traces in vicinity
  $\mathcal{V}$ of $U$}, we say that $c$, as deployed in
the \ac{CDT}, \emph{conforms} with $\mathcal{M}$ with respect to $U$,
written $c \approx_U R'$, if and only if
$\mathcal{M}\models R \land \forall t\in \mathcal{T}_{\mathcal{V}}(U)\colon
t\models R'$.  Moreover, we say that $c$ complies with $\mathcal{M}$
with respect to $\mathit{MU}$ if and only if
$\mathcal{M}\not\models R \land \neg\exists t\in
\mathcal{T}_{\mathcal{V}}(\mathit{MU})\colon t \models R'$.  Finally, we say that the
controller implementation is $(U, \mathit{\mathit{MU}}, \mathcal{V},R)$-\emph{valid} if
and only if $c\approx_U R'$ and the closest $\mathit{MU}$ we can find, with
$c\not\approx_{\mathit{MU}} R'$ and no overlap with $U$ regarding
$\mathcal{V}$, suggests implausible inputs to the \ac{CDT}.  

For controller testing, we translate requirements $R$ given as
\ac{PCTL} properties~(\Cref{tab:properties}) into corresponding
linear and metric temporal logic properties in $R'$.  Particularly,
\begin{align}
  \label{exa:ltl-ctr-det}%
  \textstyle
  \mathop{\textbf{G}} (\zeta\land\inact \rightarrow
  \zeta\mathop{\textbf{U}}_{[0,d]}\activ)
\end{align}
requires $c$ to detect the critical event, that is, to get active
whenever the sensor predicate $\zeta$ holds true within $d$ time
units.  Furthermore,
\begin{align}
  \label{exa:ltl-ctr-liv}%
  \big((\mathop{\textbf{F}} \mathit{final})
  \rightarrow
  \big(
  \mathop{\textbf{G}}(\activ\rightarrow\mathop{\textbf{F}}\mitig)
  \land
  \mathop{\textbf{G}}(\mitig\rightarrow\mathop{\textbf{F}}\inact)
  \big)\big)
  \land
  \mathop{\textbf{G}}\neg\mishp
\end{align}
states for a completed process ($\mathop{\textbf{F}} \mathit{final}$) that
whenever the controller detects $\rfct$~(\Cref{tab:riskana}) it moves
the \ac{CDT} to a state where $\rfct$ is mitigated
($\activ\rightarrow\mathop{\textbf{F}}\mitig$) and from there, given the
environment~(e.g.\xspace operator) eventually reacts, it returns the process
to a state where $\rfct$ is inactive and operation is resumed as far
as possible ($\mitig\rightarrow\mathop{\textbf{F}}\inact$).  Even if the
environment does not react, an accident never occurs
($\mathop{\textbf{G}}\neg\mishp$).

We describe $U$ and $\mathit{MU}$ as (generated) sequences of inputs
to the \ac{CDT} and the vicinity $\mathcal{V}$ as a set of (randomly
generated) configurations of the \ac{CDT} (e.g.\xspace considering the
operator being far, near, and close to the spot welder, and entering
the cell at different times).  We statistically explore
$\mathcal{T}_{\mathcal{V}}(U)$ by playing in a sample to the \ac{CDT},
record the event trace $t$, and verify $t\models R'$ using the metric
temporal logic run-time checker \textsc{pyMTL}\xspace~\citep{pyMTL2019}.

\paragraph{Data Extraction.}
\label{subsubsec:digital-twin-data-extraction}

To allow communications between actors to be recorded and analysed, an
additional abstract actor is introduced as a network \emph{snooper}.
This actor subscribes to updates from a family of actors
$\{p_i\}_{i\in 1..n}\subset\mathcal{A}$ and exposes their
communications externally as a series of discrete event streams. As
the scenario is executed, the generated streams are marked with a
reference time and origin of each data packet.  Communications between
the safety controller, process controller and other process members
are then recovered for forensic analysis and evaluation of the safety
controller.

The safety controller may be exported as a C\# extension to our
\ac{CDT} and deployed within a digital twin of the case study in
\Cref{sec:running-example}. At the point of analysis, each data stream
is composed of a series of data packets with an assigned timestamp and
entity identifier.  Each event stream from time $t=0:n$ is exported
from the database following the scenario execution.  This allows the
data to be presented as a ledger of all monitored channels, indicating
where inter-actor communication occurred, actor states have changed
and instances where the controller guards have been triggered.

 \section{Evaluation}
\label{sec:eval}

In this section, we pose three research questions about safety,
utility, and scalability, describe the evaluation methodology for each
of these questions, and discuss the results of our evaluation.

\subsection{Research Questions and Methodology}
\label{sec:methodology}

Based on the questions raised in \Cref{sec:safety-req}, we investigate
key aspects of our approach by asking three research questions~(RQs).
\begin{description}
\item[RQ1 (Safety)] What is the likelihood of accident-free
  operation under the control of a synthesised safety controller?  
\item[RQ2 (Utility)] Does the safety controller reduce the number of hard stops
  of the robot due to hazards, compared to a basic controller that
  switches off the system whenever the operator intrudes the work
  cell?
\item[RQ3 (Scalability)] How well can the proposed approach deal with
  multiple hazards and mitigation and resumption options?
\end{description}

\paragraph{Methodology for RQ1 (Safety).}

We answer \textbf{RQ1} in two stages, first based on our modelling
approach (\textbf{RQ1a}) and, second, supported by our deployment and
validation approach (\textbf{RQ1b}).

\paragraph{Methodology for RQ1a.}

We evaluate freedom from accidents according to \Cref{eq:accfreedom},
leading to probability triples comprising $\min$, the arithmetic
$\mathrm{mean}\;\mu$, and $\max$.

In the \textbf{\ac{MDP} setting}, we use \textsc{Prism}\xspace to synthesise policies
from $\mathcal{M}$ according to the three optimisation queries
\begin{align}
  \label{eq:opt_a}
  {\mathop{\textbf{R}}}^{\mathrm{pot}}_{\max=?} [ \mathop{\textbf{C}} ]
  &\land {\mathop{\textbf{P}}}_{\max=?} [ \mathop{\textbf{F}} \mathit{final}_{\mathsf{t}}],
    \tag{a}
  \\
  \label{eq:opt_b}
  {\mathop{\textbf{R}}}^{\mathrm{prod}}_{\max=?} [ \mathop{\textbf{C}} ]
  &\land {\mathop{\textbf{P}}}_{\max=?} [ \mathop{\textbf{F}}
    \mathit{final}_{\mathsf{t}}], \;\text{and}
    \tag{b}
  \\
  \label{eq:opt_c}
  {\mathop{\textbf{R}}}^{\mathrm{eff}}_{\max=?} [ \mathop{\textbf{C}} ]
  &\land {\mathop{\textbf{R}}}^{\mathrm{nuis}}_{\max=?} [ \mathop{\textbf{C}} ].
    \tag{c}
\end{align}
where $\mathit{final}_{\mathsf{t}} = \{s\in S\mid s\in\mathit{final} \land
\mbox{all tasks finished} \land s\not\in\mishp[F]\}$.
In the spirit of negative testing, \Cref{eq:opt_a} aims at maximising
the use of the safety controller~(i.e.,\xspace approximating worst-case behaviour of
the operator and other actors) while maximising the probability of
finishing two tasks, that is, finishing a workpiece and carrying
through cell maintenance.  This query does not take into account
further opmitisation parameters defined for mitigations and
resumptions.
As opposed to that, \Cref{eq:opt_b} fosters the maximisation of
\textit{prod}uctivity, any combination of decisions allowing the
finalisation of tasks is preferred, hence, transitions leading to
accidents or the use of the controller are equally neglected.
While \Cref{eq:opt_c} also forces the environment to trigger the
controller, these policies represent the best controller usage in
terms of \textit{nuis}ance and \textit{eff}ort.  Because of
restrictions in the use of $\mathop{\textbf{R}}_{\min}$ for \acp{MDP}, we
maximise costs interpreting positive values as negative~(e.g.\xspace the
higher the nuisance the better).
We then investigate the Pareto front of optimal policies synthesised from
the \Cref{eq:opt_a,eq:opt_b,eq:opt_c}.  For policies with less than
1000 states, we inspect the corresponding policy graphs~(e.g.\xspace whether
there is a path from $\mathit{initial}$ to $\mathit{final}$ or whether
paths from unsafe states reachable from $\mathit{initial}$ avoid
deadlocks).
Finally, we evaluate accident freedom according to
\Cref{eq:accfreedom}, except that we use $\mathop{\textbf{P}}_{=?}$ for \acp{DTMC}
instead of $\mathop{\textbf{P}}_{\min=?}$.\footnote{To keep manual workload under
  control, if the model checker (here, \textsc{Prism}\xspace) lists several
  adversaries, we apply the experiment only to the first listed.}

In the \textbf{\ac{pDTMC} setting}, we use \textsc{evoChecker}\xspace to synthesise
policies for $\mathcal{M}$ according to the objective
\begin{align}
  & \textstyle
    \mathop{\textbf{P}}_{\leq 0} [ \mathop{\textbf{F}} (\mathit{deadlock} \land \neg \mathit{final}) ]
    \label{eq:pdtmc-obj}
  \\ \land
  & \textstyle 
    \mathop{\textbf{R}}^{\mathrm{prod}}_{\max} [\mathop{\textbf{C}}^{\leq T}] /
    \big(\mathop{\textbf{R}}^{\mathrm{disr}}_{\max} [\mathop{\textbf{C}}^{\leq T}] +
    \mathop{\textbf{R}}^{\mathrm{eff}}_{\max} [\mathop{\textbf{C}}^{\leq T}]\big)
    \tag{Productivity}
    \label{eq:pdtmc-prod}
  \\ \land
  & \textstyle
    \mathop{\textbf{R}}^{\mathrm{nuis}}_{\min} [\mathop{\textbf{C}}^{\leq T}]
    \tag{Nuisance}
    \label{eq:pdtmc-nuis}
  \\ \land
  & \textstyle
    \Sigma_{\rfct\in F}\big( s_{\rfct} \cdot
    \mathop{\textbf{R}}^{\mathrm{risk_{\rfct}}}_{\min} [\mathop{\textbf{C}}^{\leq T}]\big)
    \tag{Risk}
    \label{eq:pdtmc-risk}
\end{align}
over a time period $[0,T]$ and with factor-specific scaling factors
$s_{\rfct}$.  This objective contains a probabilistic constraint
ruling out early deadlocks, and three optimisation queries for
maximising productivity, minimising nuisance, and minimising overall
risk from a factor set~$F$.  We then assess the resulting Pareto
front, extract a solution from this front, and use \textsc{Yap}\xspace to refine
that solution into a concrete controller.

\paragraph{Methodology for RQ1b.}

In \Cref{sec:contr-verif}, we described the synthesis of a correct
abstract controller and, in \Cref{sec:contr-impl}, its translation
into a concrete controller $c$ interfacing with the \ac{CDT} explained
in \Cref{subsec:controller-integration}.  Then, how do we assure the
transfer of the results on freedom from accidents of the abstract
controller~(\textbf{RQ1a}) to the concrete one in the \ac{CDT}?  For
this, we follow the framework in
\Cref{subsubsec:digital-twin-data-extraction} and deploy and test $c$
in the \ac{CDT} for compliance with $\mathcal{M}$ in certain
use cases and assess its behaviour in misuse cases.

The \ac{CDT} is used in a simulation capacity to validate the proposed
safety controller.  The simulation provides for
\begin{inparaenum}[(i)]
\item automated testing and
\item a safe environment for evaluation,
\end{inparaenum}
whilst representing a faithful one-to-one construction of the work
cell.  Physical, digital, or mixed environments expose the same
interface for integration with the \ac{CDT}; the same controller can
be used in either context without modification.  As such, integration
with the physical work cell is not evaluated, in part due to limited
access to the lab replica.\footnote{At the time of writing, COVID-19
  prevents access to the Sheffield Robotics lab and physical
  components of the case study.}

We follow one use case~($U$) and one misuse case~($\mathit{MU}$) to
exercise the risk factors $\rfct[HC]$, $\rfct[HS]$, and $\rfct[HRW]$
described in \Cref{tab:riskana}:
\begin{itemize}
\item[$U$:] During operation, the operator reaches across the workbench
  and walks to the spot welder.
\item[$\mathit{MU}$:] During operation, an operator reaches across the
  workbench while another operator walks to the spot welder.
\end{itemize}
  
We perform tests for the use case as outlined in \Cref{tab:uc1}.  The
operator first heads to the workbench, breaking the light barrier
($\activ[HRW]$), before heading inside the cell ($\activ[HS]$) close
to the spot welder ($\activ[HC]$). The observed stream of events during
each test can be split for the independent validation of each risk
factor. Considering the stream as a whole provides for the validation
of hazard mitigation and the absence of impact on further actions.

The operator waits at given positions while the cobot and spot welder
proceed through their scheduled activities. Variations in the time
spent by the operator in different states result in different
interleavings of the operator and robot, and such variations in turn
might be ground for the activation of hazards in the system.  The
configuration for each test is a vector of 4 values corresponding to 4
wait operations of the operator: entering the workbench, at the
workbench, entering the work cell, and at the spot welder. To bound
the time taken by each test, the values are picked such that the
operator completes its actions in less than 20s. This is ample time
for the robot to perform its own actions, and the operator is able to
disrupt the different tasks in the process either by walking to the
spot welder or reaching across the workbench.  We rely on the
Dirichlet-Rescale algorithm \citep{Griffin2020-DRSSamplingAlgo} for
generating vectors such that the values of the vector sum to a given
total, and the distribution of vectors in the constrained space is
uniform.

We rely on situation-based coverage criteria, proposed
by~\cite{Alexander2015-SituationCoverage}, to assess the performance
of our test campaign, and whether we have achieved a satisfactory
level of testing. We define our situations in terms of either the
actions of the spot welder, the actions of the robot, or the position of
the arm, when the operator is reaching at the workbench or entering
the cell in accordance with the use case. Full coverage is
achieved when all such actions or positions have been observed under
both interference factors. We also ensure all valid states for all
risk factors have been encountered~(as defined in
Figure~\ref{fig:riskfactor}). This ensures causal factors have been
encountered, mitigated, and the system could resume operation.

\paragraph{Methodology for RQ2 (Utility).}

We argue that the synthesised safety controller is better than a
state-of-the-art controller that only has a stop mode and performs no
automatic resumption.  For this argument, we compare the whole range
of controllers from the design space with those controllers that
always perform a safety stop when detecting a critical event.  Based on
that, we informally assess the increase in productivity and fluency of
collaboration, and the decrease of mean-time to finishing a process
cycle.  This argument underpins our contribution to the problem
statement in \Cref{sec:safety-req}.

\paragraph{Methodology for RQ3 (Scalability).}

We prepare and analyse multiple increments of the risk model, each
adding one critical event, mitigation options, and constraints to the
model.  We record the resulting model sizes and analysis times.

\subsection{Results}
\label{sec:results}

In the following, we present the results of our evaluation separately
for each research question.

\subsubsection{Results for RQ1a: Safety in the Model}
\label{sec:results-rq1a}

We consider as inputs a risk model and a process model of the work
cell, with a single initial state of these models where all actors are
in the activity \yapid{off} and no critical event has occurred.  The
risk model is given in \textsc{Yap}\xspace's input language and the behavioural model
is given in \textsc{Prism}\xspace's flavour of \ac{pGCL}, instrumented with \textsc{Yap}\xspace
template placeholders.  We consider the two settings explained in
\Cref{sec:optim-contr-synth}.
In the \textbf{\ac{MDP} setting}, we use \textsc{Yap}\xspace~0.5.1 for generating the
design space~(\Cref{sec:design-space-gener}) embedded into a process
model without alternating execution semantics and with controller
synthesis directly from an \ac{MDP} using \textsc{Prism}\xspace~4.5.
In the improved \textbf{\ac{pDTMC} setting}, we use \textsc{Yap}\xspace~0.7.1 for
generating the design space embedded into a process model with the
alternating execution semantics described in
\Cref{sec:execution-semantics}, an improved accident model, and with
controller synthesis from a \ac{pDTMC} using \textsc{evoChecker}\xspace and
\textsc{Prism}\xspace~4.5.
For \textbf{RQ1a}, we used GNU/\-Linux 5.4.19 and
5.8.0~(x86, 64bit), and an Intel\textregistered~Core i7-8665U CPU with up
to 8~Threads of up to 4.8~GHz, and 16~GiB RAM.

\begin{table*}%
  \caption{Results of the experiment for \textbf{RQ1a}~(accident-free
    operation) and \textbf{RQ3}~(scalability) in the \textbf{\ac{MDP} setting}}
  \label{tab:exp:results} 
  \setlength{\tabcolsep}{5.5pt}
  \begin{tabularx}{\textwidth}{Xcrr|crc|crr|crr|crr}
    \toprule
    \multicolumn{4}{c|}{\textbf{Risk Model}$^\dagger$} 
    & \multicolumn{3}{c|}{\textbf{\ac{MDP}}$^\dagger$}
    & \multicolumn{3}{c}{\textbf{(a) max-ASC}$^\dagger$}
    & \multicolumn{3}{c}{\textbf{(b) max-prod}}
    & \multicolumn{3}{c}{\textbf{(c) opt-ASC}}
    \\\rowcolor{lightgray}{\large$\phantom{\mid}$\hspace{-.25em}}
    $F$ & $mr/c$ & $\vert R(F)\vert$ & $t_Y$
    & $\mathop{\textbf{P}}_{\neg A}$  & $\Xi$ & $sta/tra$
    & $\mathop{\textbf{P}}_{\neg A}$ & $\Xi$ & $t_P$
    & $\mathop{\textbf{P}}_{\neg A}$ & $\Xi$ & $t_P$
    & $\mathop{\textbf{P}}_{\neg A}$ & $\Xi$ & $t_P$
    \\\rowcolor{lightgray}{\large$\phantom{\mid}$\hspace{-.25em}}
    & & & [ms] & $\lfloor\mu\rceil$ & &
    & $\lfloor\mu\rceil$ & & [s]
    & $\lfloor\mu\rceil$ & & [s]
    & $\lfloor\mu\rceil$ & & [s]
    \\\specialrule{\lightrulewidth}{0em}{\belowrulesep} %
    $\rfct[HC]$ & 5/0 & 3 & 40 & [.9,.9,.9] & 14 & 322/1031
    & [1,1,1] & 3 & .02
    & [1,1,1] & 1 & .02
    & [1,1,1] & 6 & .15
    \\ %
    $+\rfct[HS]$ & 9/2 & 5 & 52 & [.92,.96,.98] & 256 & 930/3483
    & [.07,.66,1] & 11 & .77
    & [0,.88,1] & 8 & .82
    & [.95,.98,1] & 18 & .9 %
    \\ %
    $+\rfct[WS]$ & 11/3 & 8 & 44 & [.93,.97,1] & 288 & 1088/3865 
    & [0,.29,1] & 17 & 2.1
    & [0,.8,1] & 5 & 2 
    & [1,1,1] & 24 & 1.5
    \\ %
    $+\rfct[HRW]$ & 13/7 & 16 & 65 & [.93,.97,1] & 981 & 7675/33322
    & [1,1,1] & 17 & 9.7
    & [1,1,1] & 11 & 9.4 
    & [1,1,1] & 15 & 13.3 
    \\ %
    $+\rfct[HW]$ & 15/8 & 36 & 76 & [.93,.97,1] & 2296 & 21281/98694
    & [1,1,1] & 15 & 42.9
    & [0,.71,1] & 7 & 41.4 
    & [1,1,1] & 15 & 46.6
    \\ %
    $+\rfct[RT]$ & 15/9 & 50 & 87 & [.93,.97,1] & 2864 & 21965/100133
    & [1,1,1] & 13 & 48.2
    & [1,1,1] & 9 & 46.4
    & [1,1,1] & 15 & 53.8
    \\ %
    $+\rfct[RC]$ & 15/15 & 122 & 162 & [.93,.99,1] & 12079 & 21670/102263
    & [0,.94,1] & 35 & 38
    & [0,.72,1] & 22 & 36.6
    & [1,1,1] & 36 & 51.1
    \\\bottomrule
    \multicolumn{16}{p{.98\textwidth}}{%
      $^{\dagger}$
      $F$\dots critical event set;
      mr/c\dots number of mitigations+resumptions/constraints;
      $\vert R(F)\vert$\dots cardinality of the relevant subset of
      $R(F)$ defined in \Cref{sec:yap};
      $t_Y$\dots \textsc{Yap}\xspace's processing time;
      $\mathop{\textbf{P}}_{\neg A}$\dots probability of conditional freedom from accidents;
      $\Xi$\dots set of $F$-unsafe states;
      $\mathit{sta/tra}$\dots number of states/transitions of the \ac{MDP} ($\mathit{sta}$ equals
      the size of the policies);
      \Cref{eq:opt_a,eq:opt_b,eq:opt_c}\dots optimisation queries;
      $t_P$\dots \textsc{Prism}\xspace's processing time
    }
  \end{tabularx}
\end{table*}

The results for the original \textbf{\ac{MDP} setting} are displayed
in \Cref{tab:exp:results}, which shows the data collected based on
seven increasingly more complex risk models.  The
$(\min, \mathrm{mean}, \max)$ probability triples are denoted by
$\lfloor\mu\rceil$.  The result $\lfloor\mu\rceil = [1,1,1]$ for a
policy denotes 100\% conditional freedom from accidents.  This desirable
result is most often achieved with \Cref{eq:opt_c} due to the fact
that simultaneity of decisions of the environment and the safety controller in
the same state is avoided by focusing on rewards only specified for
controller actions.  Such rewards model the fact that a controller is
usually much faster than an operator.
\Cref{eq:opt_a,eq:opt_b} show poorer freedom from accidents because
\textit{prod}uctivity rewards given to the environment compete with
rewards given to the safety controller to exploit its risk reduction
\textit{pot}ential.

\begin{wraptable}{r}{.4\columnwidth}
  \caption{Comparison of $\min$imum, average ($\mu$), and $\max$imum 
    freedom from accidents according to~\Cref{eq:accfreedom} of a process with a
    controller %
    and a process without one.} %
  \label{fig:exa-avg-acc-freed}
  \setlength{\tabcolsep}{5.5pt}
  \begin{tabularx}{.4\columnwidth}{Xccc}
    \specialrule{\heavyrulewidth}{\aboverulesep}{0em}\rowcolor{lightgray}{\large$\phantom{\mid}$\hspace{-.25em}}
    \textbf{Process} \dots
    & $\min$ & $\mu$ & $\max$ 
    \\\specialrule{\lightrulewidth}{0em}{\belowrulesep}
    \dots with a safety controller
    & 63\% & 97\% & 100\%\\
    \dots without a safety controller
    & 63\% & 87\% & 100\%
    \\\bottomrule
  \end{tabularx}
\end{wraptable}

In the improved \textbf{\ac{pDTMC} setting}, we focus on a single risk
model with the three factors $\rfct[HS]$, $\rfct[HC]$, and
$\rfct[HRW]$.  We calculate $\lfloor\mu\rceil$ for
\Cref{eq:accfreedom} for a process without a safety controller and one
with a controller.  \Cref{fig:exa-avg-acc-freed} shows an increase in
average freedom from accidents from 87\% to around 97\% when using a safety
controller.  Starting from any state in the process shows that there
are very safe states with a probability of accidents of down to
0\%~($\max$ column) and rather dangerous states with a probability of
up to 37\%~($\min$ column).  Overall, going from 13\% down 3\% average
accident probability~(across the three risk factors) means that the
controller in this particular process leads to a reduction of
accidents by 77\%.

\begin{wrapfigure}{r}{.5\columnwidth}
  \subfloat[Projection: risk and productivity]{\label{fig:exa-pareto-risk-vs-prod}
  \includegraphics[width=.25\columnwidth]{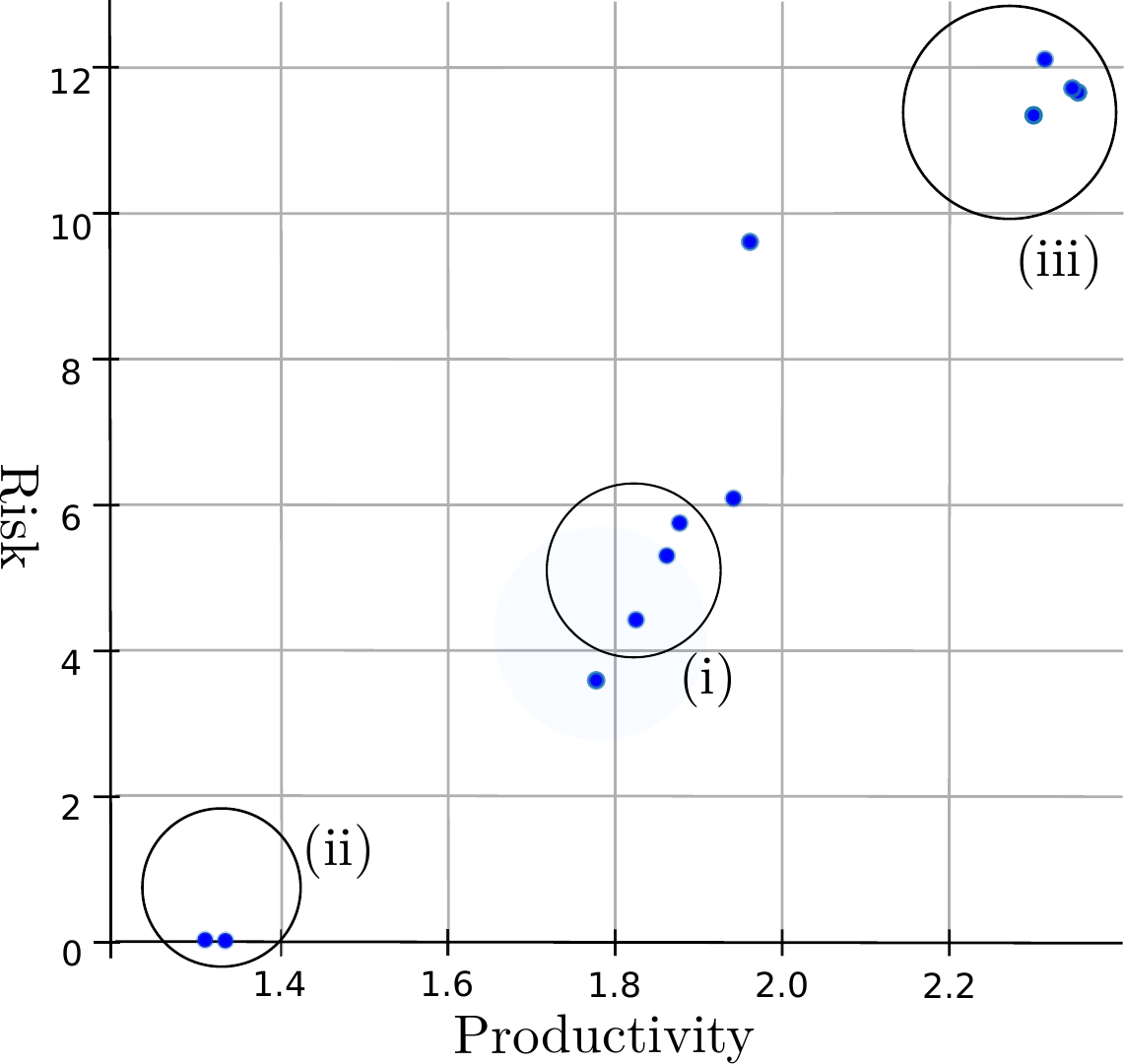}}
  \subfloat[Projection: nuisance and productivity]{\label{fig:exa-pareto-nuis-vs-prod}
    \includegraphics[width=.25\columnwidth]{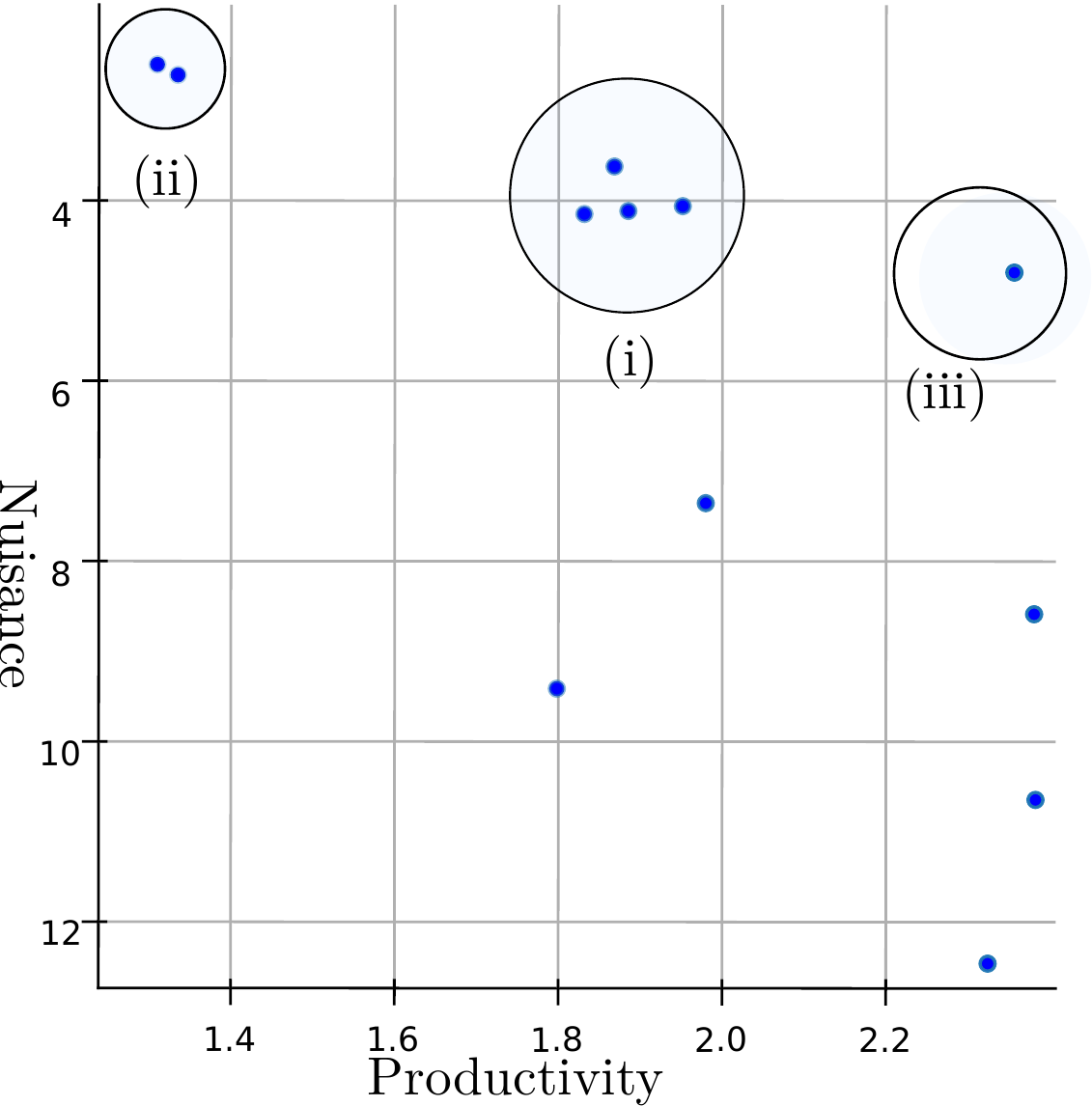}}
  \caption{Two projections of a 3D Pareto front for the optimisation objective in
    Formula~(\ref{eq:pdtmc-obj}) for the time period up to $T=30$
    \label{fig:pareto}}
\end{wrapfigure}

The 3D Pareto front in \Cref{fig:pareto} indicates that the lower the
risk of a controller~(Conjunct~\ref{eq:pdtmc-risk} in
Formula~\ref{eq:pdtmc-obj}) the lower the
productivity~(Conjunct~\ref{eq:pdtmc-prod}) due to the interventions
of the controller (cf.\xspace\Cref{fig:exa-pareto-risk-vs-prod}).
\Cref{fig:exa-pareto-risk-vs-prod,fig:exa-pareto-nuis-vs-prod} show
\begin{inparaenum}[(i)]
\item four controllers in the design space with low
  nuisance~(Conjunct~\ref{eq:pdtmc-nuis}) and medium productivity and
  risk, and
\item two minimal-risk controllers with low nuisance but low
  productivity.
\item Controllers that minimise nuisance and maximise productivity
  tend to do this at the cost of high risk.
\end{inparaenum}
If the minimal-risk controllers under (ii) are not satisfactory, a
more detailed trade-off regarding the controllers in (i) or a
repetition of the analysis with different design parameters or an
altered process and controller model will have to be made.

\subsubsection{Results for RQ1b: Safety in the Digital Twin}
\label{sec:results-rq1b}

\begin{figure}
  \parbox{.48\columnwidth}{
    \subfloat[System configuration for
    testing, with the operator's path highlighted by numbered
    waypoints.]{\label{fig:twin-setup:config}
      \includegraphics[width=0.14\columnwidth]{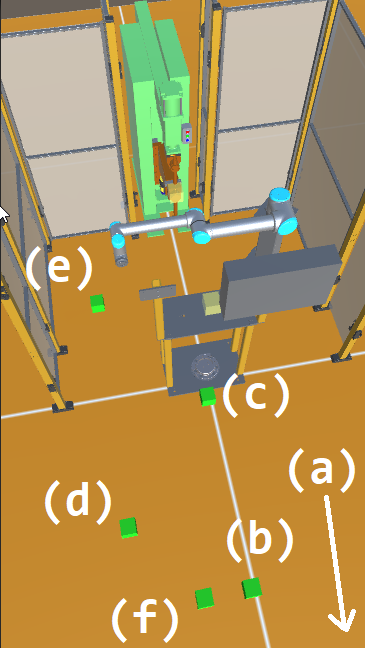}}
    \hfill
    \subfloat[Initial state. The operator waits in front the
    workbench while the cobot moves to retrieve a work piece.]
    {\label{fig:twin-setup:initial}
      \includegraphics[width=0.14\columnwidth]{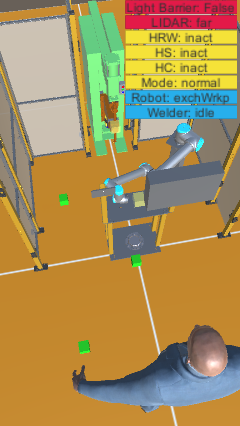}}
    \hfill
    \subfloat[The controller issues mitigations as the operator
    and cobot simultaneously reach for the workbench
    ({$\activ[HRW]$}).]{\label{fig:twin-setup:at table}
      \includegraphics[width=0.14\columnwidth]{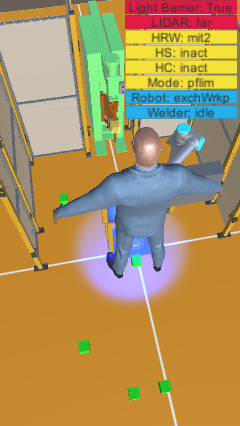}}
    \hfill
    \subfloat[The operator leaves the workbench while the cobot
    moves to the spot welder. The risk factor has been removed.]
    {\label{fig:twin-setup:front plant}
      \includegraphics[width=0.14\columnwidth]{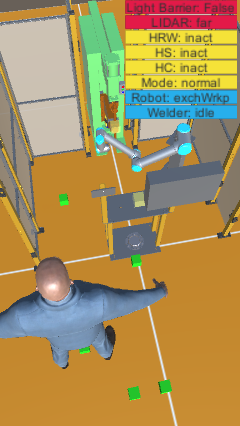}}
    \hfill
    \subfloat[The controller issues a stop in response to the
    operator approaching the spot welder during operation
    ({$\activ[HS]$ and $\activ[HC]$}).]  {\label{fig:twin-setup:at plant}
      \includegraphics[width=0.14\columnwidth]{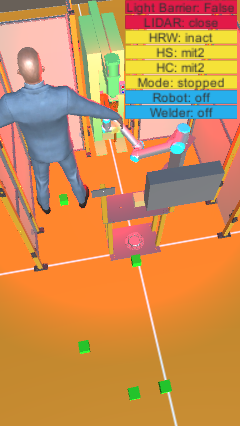}}
    \hfill
    \subfloat[Welding resumes after the operator leaves the
    work cell. The notification to leave the cell is deactivated.]
    {\label{fig:twin-setup:final}
      \includegraphics[width=0.14\columnwidth]{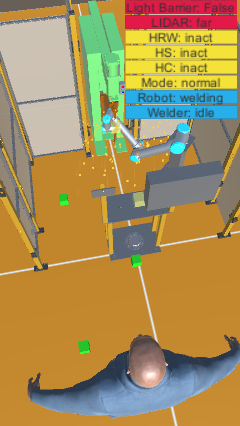}}
  }
  \hfill
  \parbox{.5\columnwidth}{
    \subfloat[{Use case describing operator, cobot and spot
    welder (inter)actions. 
    Each image on the left further highlights,
    in the top right corner, the state of the different actors of the
    system as the use case progresses, that is, the sensors (light
    barrier and LIDAR), the safety controller ($\activ[HRW]$,
    $\activ[HS]$, $\activ[HC]$, and the safety \emph{mode}), and the
    activities (of the cobot and the spot welder).  A visual
    notification instructing the operator to leave the handover table or
    welding area are issued in the form of blue and red lights
    respectively.}]{
      \label{tab:uc1}\includegraphics[width=.5\columnwidth]{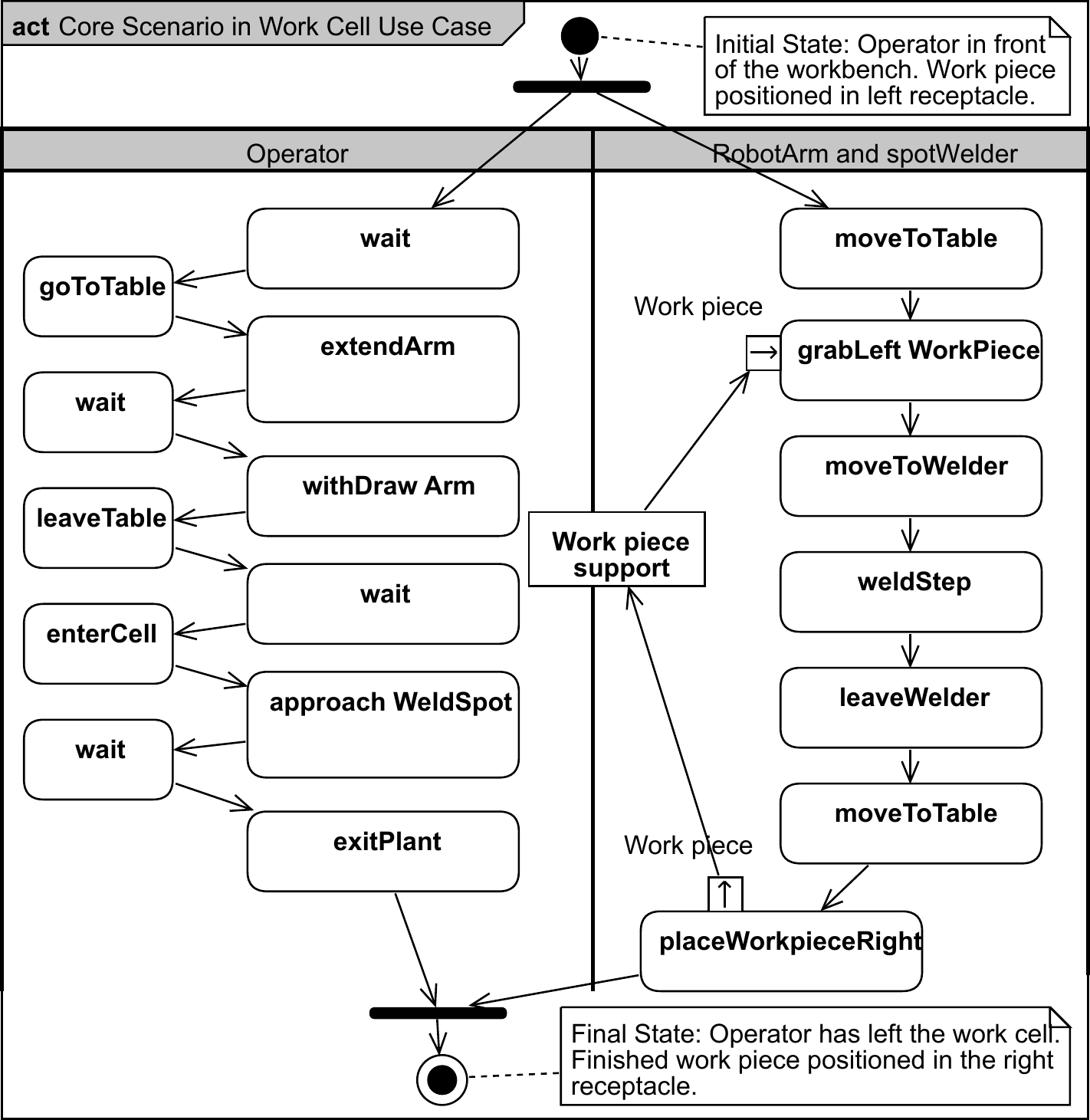}}}
  \caption{Illustration of different states
    (\ref{fig:twin-setup:config})-(\ref{fig:twin-setup:final}) of the
    use case described in (\ref{tab:uc1}).
  \label{fig:twin-setup}}
\end{figure}

We performed 100 tests for the use case shown in \Cref{tab:uc1}.  This
proved sufficient to achieve full coverage of
\begin{inparaenum}[(i)]
\item the mitigation states for all considered risk factors, and
\item operator interference types across spot welder and cobot states.
\end{inparaenum}
All recorded traces $t$ verify $t\models R'$, that is, all traces
satisfy the selected properties translated from
\Cref{tab:properties}. Risk factors were correctly detected~(with
$d=0.25ms$ in Property~\ref{exa:ltl-ctr-det}) and mitigated by the
synthesised safety controller in all observed situations in the
\ac{CDT}.  All results for \textbf{RQ1b} were produced under the
Windows 10 Home Edition operating system, build 19042.985, on an
Intel\textregistered\ Core i5-8250U 4-core CPU at 1.8~GHz, with 8~GiB
RAM.  The whole test suite, including validation of the traces, took
less than two hours to complete.  No time compression was used, the
\ac{CDT} ran simulations in real-time.

\begin{wrapfigure}{r}{.35\columnwidth} %
  \begin{tikzpicture}
    [label/.style={text width=1.5cm,align=right}]
    \node at (0,0)
    {\includegraphics[width=.23\columnwidth]{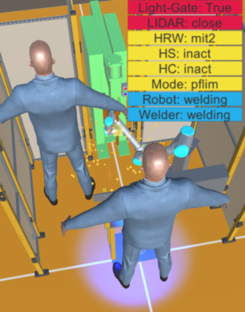}};
    \node[label] (sw) at (-3,2.25) {Spot welder};
    \node[label] (op) at (-3,.75) {Operator};
    \node[label] (cb) at (-3,0.1) {Cobot};
    \node[label] (wb) at (-3,-1.5) {Workbench};
    \draw[->,white,line width=2pt]
    (sw) edge ++(2.5,0)
    (op) edge ++(1.5,0)
    (cb) edge ++(3.9,0)
    (wb) edge ++(2.8,0)
    ;
  \end{tikzpicture}
  \caption{Result for the misuse case ($\mathit{MU}$): the second operator is
    not notified to leave the work cell, and the spot welder remains active
    while he is in close proximity.
    \label{fig:twin-mu}}
\end{wrapfigure}

\Cref{fig:twin-setup} illustrates the output of the \ac{CDT} for the
use case~(\Cref{tab:uc1}). Each image captures the state of the system
as the operator begins a wait operation. The use case highlights the
activation of all monitored risk factors~(see \Cref{tab:riskana}) with
the operator and cobot making concurrent use of the workbench, and the
operator in the vicinity of the spot welder as an operation is about
to start.

$\mathit{MU}$ relies on two operators simultaneously following the
behaviours outlined by $U$.  The tested model assumes a single
operator interacts with the system, such that an operator must leave
the workbench before entering the work cell.  As such, the safety
controller correctly mitigates the risk due to both cobot and operator
reaching for the workbench.  However, as the second operator enters
the cell the related risk factors are neither identified as active nor
mitigated.  \Cref{fig:twin-mu} highlights the situation, with the
second operator in close proximity to an active spot welder.

The work cell is safe considering a single operator following the use
case, that is entering the work cell or the workbench during
operation.  However, the risk and process models would need to be
revised to account for multiple operators interacting with the
system~(e.g.\xspace by removing factor dependencies between $\rfct[HRW]$ and
$\rfct[HS]$, see \Cref{sec:refin-fact-modell}).  The case with two
operators would normally be picked up in an extended risk assessment
following \Cref{tab:riskana}, but it was not relevant to this
particular process.  This case is also not within scope for the
purposes of keeping this example simpler.

\subsubsection{Results for RQ2: Utility}
\label{sec:results-rq2}

In the \textbf{\ac{MDP} approach}, \Cref{fig:exa-ctr-productivity}
highlights several aspects.  First, the fact that none of the controller
configurations achieves the productivity level~(64.19) of the process
without a controller.  Such a process reaches a theoretical
productivity maximum assuming no safety requirements, which is for
obvious reasons not desirable.  Perhaps unsurprisingly, safety comes
at the cost of productivity.
However, more importantly, among all
configurations, the one switching to ``stopped'', whenever
$\acend{HS}$, $\acend{HC}$, or $\acend{HRW}$ occurs, is the one
performing the poorest (37.06, highlighted in gray).  As expected,
controllers implementing the \yapkeyw{pflim} and \yapkeyw{ssmon}
safety modes as part of their policy are among the highest performing
variants~(51.42).  With the best safety controller, the process
achieves about 40\% more productivity than with the worst controller
and is only 25\% less productive than the maximum productivity
achievable in the process.

Moreover, the most risky configurations (68, highlighted in gray) are
around 77\% less risky than not using a safety controller.  However,
the most difficult trade-off is to be made between very low risk
controllers ($\leq 9$) and moderate risk controllers ($\geq 49$) where
the gain in productivity is comparatively little.  Finally, with this
model instance, the addition of the factor $\rfct[HC]$ does neither
influence risk nor productivity.

Regarding the expected overhead (\Cref{sec:expoverhead}), we used 
the C\# profiling tool integrated in the development environment for 
the \ac{CDT}.  We measured the execution time of control cycles in 
the \ac{CDT} across 1000 invocations of the controller.  Our 
observations cover both nominal activity, and the occurrence of risk 
factors.  The longest observed execution time  
$d^{\rfct[HS,HC,HRW]}_{\max}$, its high watermark, is around $40ms$. 
The high watermark occurs upon activation of a risk factor, and
accounts for mode switch, and the required mitigation actions.  In 
the absence of a mitigation requirement, the average execution time  
$d^{\rfct[HS,HC,HRW]}$ is less than $1ms$.  Execution times
are expected to be lower in the physical twin.  Profiling was performed
on the same environment as for \textbf{RQ1b}, that is a 1.8~GHz i5 CPU
with 8~GiB of RAM running Windows 10.

\subsubsection{Results for RQ3: Scalability}
\label{sec:results-rq4}

We answer this question using six increments applied to a basic risk
model~(\Cref{tab:exp:results}).  We only consider the \textbf{\ac{MDP}
  setting} because applying model increments in the \textbf{\ac{pDTMC}
  setting} will not provide additional insights into the scalability
aspect of our method.

For demonstration of \textsc{Yap}\xspace's capabilities, the incident $\rfct[RT]$ and
the accident $\rfct[RC]$ are included in the risk model without
handler commands.  However, these factors add further constraints on
$R(F)$ to be dealt with by the safety controller.  Hence, $mr$ stays at 15
actions and $c$ rises to $15$ constraints.  In model 7~(last line of
\Cref{tab:exp:results}), $R(F)$~(122 risk states) and the
$\Xi$-region of $S$~(12079 states) differ by two orders of
magnitude.
Risk states offer a higher level of abstraction for risk assessment.
The derivation of properties that focus on relevant regions of the
\ac{MDP} state space from a risk structure can ease the state
explosion problem in explicit model checking.  For example, the
constraints $\rfct[HRW]$ \yapkeyw{prevents} $\rfct[HC]$ and
$\rfct[HC]$ \yapkeyw{prevents} $\rfct[HRW]$ express that the combined
occurrence of $\rfct[HC]$ and $\rfct[HRW]$ is considered infeasible or
irrelevant by the safety engineer.  Hence, checking properties of the
corresponding region of the \ac{MDP} state space can be abandoned.

 \subsection{Threats to Validity}
\label{sec:threats}

\paragraph{Internal Validity.}

\emph{Too strong environmental assumptions reduce the scope of the
  safety guarantee.}  To limit the need for game{\Hyphdash}theoretic
reasoning about $\mathcal{M}$, we reduce non{\Hyphdash}deterministic
choice for the environment~(i.e.,\xspace operator, cobot, spot welder).  The
more deterministic such choice, the closer is the gap between the
policy space $\Pi_{\mathcal{M}}$ and the controller design space.  Any
decisions left to the environment will make a verified policy
$\pi^{\star}$ safe relative to $\pi^{\star}$'s environmental
decisions.  These decisions form the assumption of the controller's
safety guarantee.

\begin{wraptable}{r}{.5\columnwidth}
  \caption{Comparison of a process without a safety controller with processes
    including controllers using different safety mode and task
    switching policies.  Safety functions such as warnings are not
    taken into account in this comparison.}
  \label{fig:exa-ctr-productivity}
  \setlength{\tabcolsep}{5.5pt}
  \begin{tabularx}{.5\columnwidth}{Xlllrr}
    \specialrule{\heavyrulewidth}{\aboverulesep}{0em}\rowcolor{lightgray}{\large$\phantom{\mid}$\hspace{-.25em}}
    \textbf{Process}
    & \multicolumn{3}{c}{\textbf{Configuration}$^{\dagger}$}
    & \textbf{Productivity}$^{\ddagger}$
    & \textbf{Risk}$^{\ddagger}$
    \\\rowcolor{lightgray}{\large$\phantom{\mid}$\hspace{-.25em}}
    & $\rfct[HC]$ & $\rfct[HRW]$ & $\rfct[HS]$
                  & [units] & [units]
    \\\specialrule{\lightrulewidth}{0em}{\belowrulesep}
    & smrst / idle & stop & stop %
                  & 37.06 & 0 \\
    \rowcolor{lightgray}
    & stop / off & stop & stop & 37.06 & 0 \\
    & hguid / -- & stop & stop & 37.06 & 0 \\
    & smrst / idle %
    & pflim %
    & stop %
    & 44.75 & 9 \\
    & stop / off %
    & pflim %
    & stop %
    & 44.75 & 9 \\
    \rowcolor{lightgray}
    & hguid / -- %
    & pflim %
    & stop %
    & 44.75 & 9 \\
    \rowcolor{lightgray}
    & smrst / idle %
    & stop %
    & pflim %
    & 46.26 & 49 \\
    & stop / off %
    & stop %
    & pflim %
    & 46.26 & 49 \\
    Using controller
    & hguid / -- & stop & pflim & 46.26 & 49 \\
    & smrst / idle & stop & ssmon %
    & 48.15 & 59 \\
    & stop / off & stop & ssmon & 48.15 & 59 \\
    & hguid / -- & stop & ssmon & 48.15 & 59 \\
    & smrst / idle %
    & pflim %
    & pflim %
    & 49.90 & 58 \\
    & stop / off & pflim & pflim & 49.90 & 58 \\
    & hguid / -- & pflim & pflim & 49.90 & 58 \\
    \rowcolor{lightgray}
    & smrst / idle & pflim & ssmon & 51.42 & 68 \\
    \rowcolor{lightgray}
    & stop / off & pflim & ssmon & 51.42 & 68 \\
    \rowcolor{lightgray}
    & hguid / -- & pflim & ssmon & 51.42 & 68
    \\\specialrule{\lightrulewidth}{0em}{\belowrulesep}
    No controller
    & -- & -- & -- & 64.19 & 291
    \\\bottomrule
    \multicolumn{6}{p{.48\columnwidth}}{%
    $^{\dagger}$Configuration of mitigation options (safety mode and
    activity switches),
    \emph{smrst}: safety-rated monitored stop,
    \emph{stop}: power shutdown with user-initiated work cell reset,
    \emph{hguid}: hand-guided operation,
    \emph{pflim}: power \& force limitation,
    \emph{ssmon}: speed \& separation monitoring,
    \emph{idle}: currently not performing work steps,
    \emph{off}: work cell not in operation.\newline
    $^{\ddagger}$According to $\mathop{\textbf{R}}^{prod}_{\max=?}[\mathop{\textbf{C}}^{\leq
    T}]$ and Property~\eqref{eq:pdtmc-risk} with $T=50$.}
  \end{tabularx}
\end{wraptable}
Occupational health and safety assumes trained operators not to act
maliciously, suggesting ``friendly environments'' with realistic human
errors.  To increase the priority of the controller in the
\textbf{MDP-based approach}, we express realistic worst-case
assumptions, for example, by minimising factor- and activity-based
risk and maximising the risk reduction potential.  The
\textbf{\ac{pDTMC}-based approach} weakens such assumptions by the
fact that all outcomes of environmental decisions remain in the policy
in form of probabilistic choices.  Hence, the controller extracted
from $\pi^{\star}$ will contain an optimal choice for all states
reachable from environmental decisions.
Obtaining appropriate trade-offs between productivity and safety is
challenging.  Hence, in the future we may switch to game-theoretic
approaches which may allow for a more nuanced view of the environment
and hence allow for even greater productivity.

\emph{Incorrect application of tools.}  \textsc{evoChecker}\xspace relaxes
constraints if the constrained solution space is empty, leading to
solution under relaxed constraints.  Hence, the design space needs to
be redefined in order for \textsc{evoChecker}\xspace to select from a non-empty set
of solutions obeying safety-related hard constraints.

\emph{Insufficient validation.}  Confidence in the testing-based
validation depends on the range and complexity of (mis)use cases and
the parameters used for randomisation.  The use case for \textbf{RQ1b}
is generic enough to cover a wide range of work cell scenarios and the
misuse case clarifies the environmental assumptions of the use
case.

\paragraph{External Validity.}

\emph{Reality gap through inadequate quantification.}  The lab replica
of the work cell matches closely to the one in the company.  This
allowed us to develop an accurate digital twin and an \ac{MDP}
compliant with the actual work cell that incorporates all relevant
parameters (particularly, probabilities).  We crafted the \ac{MDP}
such that assessment and optimisation based on probabilities and
rewards rely more on qualitative relations between the parameters
rather than accurate numbers.  It is more important and easier to
conservatively assess whether an activity or safety mode is more or
less dangerous than another and it is less important and more
difficult to provide accurate numbers.  The resulting gap between
optimality in the model and optimality in the digital twin should
hence not alter the demonstrated confidence in the controller.  More
accurate numbers can be obtained from observations, and after a model
update, a repeated synthesis can improve productivity of the safety
controller.

\emph{Incorrect sensing assumptions.} 
\label{sec:requ-sens-syst}
In our case study, the safety controller relies on the detection of an
operator~(e.g.\xspace extremities, body) and a robot~(e.g.\xspace arm, effector)
entering a location, the cell state~(e.g.\xspace grabber occupied, workbench
support filled), and the work piece location~(e.g.\xspace in grabber, in
support).  For $\mathcal{M}$, we assume the tracking system~(i.e.,\xspace
range finder and light barrier in the industrial setting,
MS~Kinect\footnote{See \url{https://en.wikipedia.org/wiki/Kinect}.}
in the lab replica) to map the location of the operator and robot to
the areas ``in front of the workbench'', ``on the workbench'', ``in
the workcell'', and ``close to the spot welder'' rather than to a
fine-grained occupancy grid.  In \Cref{fig:setting-actual1}, the range
finder signals ``at welding spot'' if the closest detected object is
nearer than the close range, and ``in cell'' if the closest object is
nearer than the wide range.
Tracking extensions, not discussed here, could include object
silhouettes and minimum distances, operator intent, or joint
velocities and forces.

\emph{Incorrect real-time behaviour.} 
\ac{pGCL}, as we used it, requires care with the modelling of
real-time behaviour, particularly, when actions from several
concurrent modules are enabled.  To model real-time controller
behaviour, we synchronise operator actions with sensor events and, in
the original \textbf{\ac{MDP} setting}, force the priority of
controller reactions in $\pi^{\star}$ by maximising the risk reduction
potential~(cf.\xspace$\mathit{pot}$ in \Cref{tab:properties}).  While
synchronisation restricts global variable use, increasing
$\mathcal{M}$'s state space, we found it to be the best solution in
the multi-module \textbf{\ac{MDP} setting}.  The alternating execution
scheme used in \textbf{\ac{pDTMC} setting}, however, avoids the use of
rewards to implement priorities and prevent actors from competing in
an unnatural way.

\subsection{Discussion}
\label{sec:disc}

\paragraph{Tool Restrictions and Model Debugging.}

State rewards allow a natural modelling of, e.g.\xspace risk exposure.
However, in \textsc{Prism}\xspace~4.5, one needs to use action rewards for
multi-objective queries of \acp{MDP}.
Risk gradients help to overcome a minor restriction in \textsc{Prism}\xspace's
definition of action rewards.\footnote{Currently, rewards cannot be
  associated with particular updates, that is, with incoming
  transitions rather than only states.}  Alternatively, we could have
introduced extra states at the cost of increasing $\mathcal{M}$'s
state space, undesirable for synthesis.
Rewards require the elimination of non-zero end components~(i.e.,\xspace
deadlocks or components with cycles that allow infinite paths
and, hence, infinite reward accumulation).  \textsc{Prism}\xspace provides useful
facilities to identify such components, however, their elimination is
non-trivial and laborious in larger models and can require intricate
model revisions.
We strongly discretise model parameters such as location to further
reduce the state space and keep the model small.  We use probabilistic
choice in synchronous updates only in one of the participating
commands to simplify debugging.  We avoid global variables to support
synchronisation with complex updates

\paragraph{Misuse Cases for Controller Testing.}

Misuse cases help in exploiting deficiencies of a control concept
through counterexamples.  The latter can be generated from the process
model and exercised as negative test cases to aid in debugging the
controller.  We explored this idea in
\cite{Gleirscher2011-HazardbasedSelection}, applying a
\textsc{Prolog}-based \textsc{Golog} interpreter that constructs
counterexamples from backward depth- and breadth-first search from a
given accident state.  \textsc{Prism}\xspace uses forward breadth-first search from
an initial state and stops when reaching an accident state.  In both
cases, explicit state exploration is used with the usual problem of
state space explosion.  In any case, an environment model is needed to
express accident states.  In this work, we prefer a model checker
because the encoding of the stochastic process in \ac{pGCL} appeared
to be easier than in a stochastic situation calculus and because of
more flexibility of expressing properties to be verified in \ac{PCTL}.

 \section{Related Work}
\label{sec:related-work}

Our work builds upon a set of well known theoretic principles on which
research on controller design and
synthesis~\citep{kress-gazit_synthesis_2018} for collaborative robots
has been carried out.  In the following, we compare our results with
other results.

\paragraph{Risk-informed Controller Design and Verification.}

\citet{Askarpour2016-SAFERHRCSafety} discuss controller assurance of a
cobot work cell %
based on a discrete-event formalisation in the linear-time temporal
language TRIO.  Actions are specified as
\emph{pre/\-inv/\-post}-triples for contract-based reasoning with the
SAT solver Zot.  Violations of the safety invariant \emph{inv} lead to
pausing the cell.  Fine-grained severity quantification
\citep{vicentini_safety_2020} allows a controller to trigger safety
measures on exceeding of certain risk thresholds.
Additionally,
\cite{askarpour_formal_2019} present a model of erroneous or
non{\Hyphdash}deterministic operator behaviour to enable designers to
refine controller models until erroneous behaviours are mitigated.
This approach aims at risk-informed prototyping and
exhaustive exploitation of all possible executions to identify
constraint violations and remove hazardous situations.  
Risk thresholds correspond to a refined variant of
Property~\eqref{exa:ltl-ctr-det}.  Whilst not the focus of this work,
our approach also allows for the quantification of severity in
detector predicates ($\zeta$).  Their severity model
\citep{askarpour_formal_2019,vicentini_safety_2020} inspires future
risk factor models.  Instead of a priority parameter, which reduces
state variables, we use guards to implement action orderings.  Our
approach is more flexible because it can deal with multiple mitigation
options offering more variety in safety responses.  Beyond model
consistency checks and the search of counterexamples for model repair,
our approach yields an executable policy.  Our use of \ac{pGCL} and
\ac{PCTL}~\citep{Kwiatkowska2007-StochasticModelChecking} results in a
separation of action modelling and property specification.  Although
this separation is a non-essential difference, it syntactically
supports a more independent working on two typical abstraction levels,
required process properties and process implementation.

\paragraph{Verified Controller Synthesis for Cobots.}

A number of authors have suggested synthesis approaches for controllers in
human-robot collaboration.  Key features of these approaches are
verified optimal synthesis for collaborative plan execution,
quantitative verification of plans, code generation for deployment on
robot platforms.

For generic robots, \cite{Orlandini2013-ControllerSynthesisSafety} and
\cite{bersani_pursue_2020} employ synthesis by game solving~(i.e.,\xspace
finding winning strategies) over timed game automata~(TIGA) supported
by the model checker UPPAAL-TIGA.  Correctness properties can be
formulated as reach-avoid problems~(i.e.,\xspace reach the goal state and
avoid unsafe states) specified as
$\mathop{\textbf{A}}[\mathit{safe} \mathop{\textbf{U}} \mathit{goal}]$ in timed
computation tree logic.
From a timeline-based plan
description~\citep{Cesta2008-timelinerepresentationframework},
\cite{Orlandini2013-ControllerSynthesisSafety} generate a TIGA with
clock constraints encoding temporal degrees of freedom for performing
control actions.  From the TIGA, a winning strategy (i.e.,\xspace a robust
plan execution minimising violations of clock constraints) is then
synthesised.  The TIGA-based stage is continuously performed during
operation.  The distinction of controllable~(i.e.,\xspace duration known) from
uncontrollable actions~(i.e.,\xspace duration unknown) allows one to react to
temporally uncertain environmental events by obtaining strategies that
can schedule robot actions in the presence of worst-case timing of the
environment.
In \cite{Cesta2016-Towardsplanningbased}, an extension of this
approach is applied to controller synthesis for safe human-robot
collaboration.
For task coordination between
humans and robots, \cite{Cesta2016-Towardsplanningbased} and other works utilise the distinction of uncontrollable and controllable actions.
Our reactive execution scheme~(\Cref{fig:int-sem}) accommodates this basic feature required
to separate the capabilities of the cobot from the capabilities of its physical environment.

\cite{kshirsagar_specifying_2019} demonstrate on-line controller
synthesis for human-robot handovers obeying timing constraints
specified in signal temporal logic.
Cobot kinetics need to be given in terms of ordinary differential
equations.  Being suitable for low-level synthesis in homogeneous
action systems, this approach could be integrated into our framework,
for example, for controlling a speed and separation monitoring mode
switched on during a handover.

\ac{MDP} policy synthesis has also been used to generate optimal motion plans for mobile robots~\citep{lahijanian_temporal_2012}. The \acp{MDP} used in this solution model the physical layout of the space within which a mobile robot can navigate. As such, the problem addressed in this work is complementary to our use of probabilistic models and \ac{MDP} policy synthesis, as our approach tackles the generation of optimal hazard mitigation actions.

\paragraph{Modelling of Controllers for Cobots.}

\emph{Domain{\Hyphdash}specific languages} for controller design can
support control engineers in encoding their control schemes.
\cite{Cesta2016-Towardsplanningbased} employ domain and problem
definition languages~(DDL \& PDL) for encoding the state spaces and
action domains a controller can select from.
\cite{bersani_pursue_2020} provide a lean language for robotic
controller design.  Models in that language typically include a
description of the uncontrollable environment.
Safety properties can be difficult to formulate, as shown in
\cite{bersani_pursue_2020} and, particularly, when dealing with
multiple hazards or risk factors.  We provide \textsc{Yap}\xspace's input language for safety controller design informed
by risk models, streamlining the specification of safety properties from multiple hazards.

Approaches such as \cite{lahijanian_temporal_2012} and
\cite{kshirsagar_specifying_2019} focus on \emph{homogeneous action
  systems}, that is, systems with few and similar types of actions
manipulating type-wise simple state spaces~(e.g.\xspace movement in a
Euclidean plane or in a 2D grid).  Such systems make it easier to
provide complete and tractable synthesis algorithms for realistic
models.  In contrast, and as suggested
by~\cite{kress-gazit_synthesis_2018}, our approach focuses on
\emph{heterogeneous action systems}, such as human-robot
collaboration, which can be characterised by a wide variety of actions
over a heterogeneously typed state space.  Such actions can differ in
their complexity and discrete or continuous nature~(e.g.\xspace grabbing a
work piece, moving a robot arm, performing a welding action, switching
a mode, turning off an alarm sound).

Robust controllers are able to \emph{deal with unstructured
  environments}, to react to a wide variety of uncontrollable adverse
events~(e.g.\xspace human error).  Overall, they guarantee correct behaviour
under weak assumptions.  Game-based approaches, such as policy
synthesis for TIGAs \citep{raskin_guided_2007,
  Orlandini2013-ControllerSynthesisSafety, bersani_pursue_2020},
inherently support such environments.
Whilst our \ac{MDP}-based approach benefits from more efficient
algorithms and provides fine-grained methodological guidance, an
extension to utilise the greater flexibility of game-based approaches
should be part of our next steps.

A distinctive feature of safety controllers is their ability to
\emph{control the resumption of normal operation} (i.e.,\xspace the recovery
from a conservative or degraded safe state) in addition to mitigation.
While resumption is implicit to many solutions for
homogeneous{\Hyphdash}action reach{\Hyphdash}avoid
problems~\citep{Orlandini2013-ControllerSynthesisSafety,
  bersani_pursue_2020}, for heterogeneous action systems, resumptions
often need to be modelled separately (e.g.\xspace switching off a safety mode
or function, resuming/restarting a suspended/cancelled task).  Risk
factors at the core of our approach accommodate primitives for
specifying resumptions.

Overall, an advantage of timeline-based approaches
\citep{Cesta2016-Towardsplanningbased} is their ability to encode
physical domains qualitatively.  In \ac{pGCL}, corresponding domain
constraints need to be distributed over action guards, which can be
cumbersome.  Moreover, an advantage of TIGAs over \acp{MDP} is that
time is continuous and managed via clocks, leading to more concise
models.
However, in summary, our use of \ac{pGCL} enables the concise encoding
of the action domain of human-robot collaboration with multiple actors
operating over a discrete state space.  \acp{MDP} represent a natural
model of interaction of agents with an uncontrollable and uncertain
environment.  Reward-enhanced \ac{PCTL} provides a flexible and
expressive language for specifying multiple objectives (e.g.\xspace minimum
risk, maximum performance) and reward constraints (e.g.\xspace accepted risk
thresholds).  We could enhance our approach to probabilistic timed
automata in order to further increase model fidelity.

\paragraph{Controller Deployment.}

An important step in many synthesis approaches is the deployment of
the resulting controllers in an execution environment.
Although for another domain (i.e.,\xspace climate control in a pig stable),
\cite{raskin_guided_2007} show how controllers synthesised using
UPPAAL-TIGA can be embedded as components of a Simulink model to
perform validation by simulation.
As part of their evaluation, \cite{bersani_pursue_2020} demonstrate
the deployment of their controllers on the low-cost platform Turtlebot
used as a cobot.
\cite{Orlandini2013-ControllerSynthesisSafety} deploy their approach
as a module on a G${^{en}}{_{o}}$M-based robotic software platform
with a mapping into the real-time framework BIP.  Their flexible
approach could be a potential route for future extensions of our
approach.
Currently, \textsc{Yap}\xspace provides a generator for C\# modules for the
\ac{CDT}~(\Cref{subsec:what-is-a-digital-twin,subsec:controller-integration}).

Overall, to the best of our knowledge, our method is the first
end-to-end approach to synthesising and deploying safety controllers
for handling multiple risks from collaborative robots in manufacturing
processes.  The works discussed above either address parts of the
synthesis challenge or implement alternative solutions for cobot
(safety) controllers.

 \section{Conclusion}
\label{sec:conclusion}

We introduced a tool-supported software engineering approach for the verified synthesis of
optimal safety controllers from \aclp{MDP}, focusing on human-robot
collaboration.  These software controllers implement regulatory safety goals
for such applications.  We describe steps for streamlined application
modelling and risk-informed controller design and demonstrate our
method using a tool chain consisting of
\textsc{Yap}\xspace~\citep{Gleirscher-YapManual} for structured risk modelling and
\ac{pGCL} program generation,
\textsc{evoChecker}\xspace~\citep{Gerasimou2018-Synthesisprobabilisticmodels} for
search-based policy synthesis from \acp{pDTMC}, and
\textsc{Prism}\xspace~\citep{Kwiatkowska2011-PRISM4Verification} for probabilistic
model checking and \ac{MDP} policy synthesis.
We show that our approach can be used to incrementally build up
multi-hazard models including alternative mitigation and resumption
strategies.  We also discuss how our approach can simplify explicit
model checking when dealing with large state spaces.
Our approach improves the state of the art of controller synthesis for
collaborative robots, particularly when dealing with multiple risks,
mitigation options, activities and safety modes.  That way, we
contribute to the alignment of lower-level requirements posed by cobot
safety standards~(e.g.\xspace ISO~15066) with higher-level cobot hazard
analysis and risk
assessment~\citep{Chemweno2020-Orientingsafetyassurance} through a
structured risk-informed design approach.
Furthermore, we translate the verified controllers into executable
code and deploy them on the \acl{CDT} and, thus, accomplish a smooth
transition between formal controller verification and testing-based
controller validation in a realistic environment.  Using the \ac{CDT}, 
we demonstrate the controller's correct and timely 
response in a representative use case. 
In summary, the verification and validation results generated by our
approach can contribute evidence to a controller assurance
case~\citep{Gleirscher2019-EvolutionFormalModel,
  Foster2020-TowardsDeductiveVerification,entrust2018}.

\paragraph{Future Work.}

For optimal synthesis, the proposed method uses parameters such as
upper risk and severity bounds as constraints.  We plan to introduce
parameters for probabilities, such as sensor failure and human error,
into the \ac{MDP} and to use parametric risk gradients by extending
\textsc{Yap}\xspace.

It is important to model all relevant behaviours of the environment,
or more generally the uncontrollable actions, the safety controller
needs to respond to in order to increase freedom from accidents.
Hence, in future work, we plan to use stochastic games to more
accurately (and less conservatively) model the behaviour of the
environment in which this controller operates.
We also plan to explore online policy synthesis~\citep{synthesis2017}
to allow more variety in environmental decisions~(e.g.\xspace mitigating
hazards due to malicious operators).  This corresponds to weakening
the assumptions under which the controller can guarantee safety.

Accident data for the considered industrial application was generally
not available.  We were also unable\footnote{Due to restricted lab
  access during the COVID-19 pandemic.} to collect data from the lab
replica.  Thus, we had to make best guesses of
probabilities~(cf.\xspace\Cref{sec:stochastic-modelling}).  However, the
frequency of undesired intrusion of operators into the safeguarded
area and accident likelihood can be transferred into our case study.

The \ac{CDT} is a full-fledged digital twin, from which the
interaction between the safety controller and the real system can be
demonstrated.  Limited access to the physical cell has required
validation in the digital twin to take precedence. Further validation
integrating the physical work cell will therefore be the focus of
future work.

The case study can be extended by randomised control decisions with
fixed probabilities~(e.g.\xspace workload), by adding uncertain action
outcomes~(e.g.\xspace welding errors), and by time-dependent randomised
choice of mitigation options.  To use time in guarded commands, we
want to explore clock-based models rather than only using reward
structures, as far as the synthesis capabilities allow this.

For motion planning over finite partitions of geometric spaces,
\cite{lahijanian_temporal_2012} describe algorithms for the synthesis
of \ac{MDP} policies that maximise the probability of a given
arbitrary \ac{PCTL} formula.  By assuming that state estimation is
precise, the authors avoid the use of partially observable \acp{MDP}
in their application.  To improve our approach regarding the synthesis
over homogeneous action systems, their algorithms could be integrated
into our \ac{MDP} synthesis tool chain, for example, in addition to
\textsc{Prism}\xspace or underpinning the search-based synthesis in \textsc{evoChecker}\xspace.

\bibliography{}
\end{document}